\PassOptionsToPackage{dvipsnames,svgnames,table}{xcolor}
\documentclass[10pt,journal,compsoc]{IEEEtran}

\ifCLASSOPTIONcompsoc
  \usepackage[nocompress]{cite}
\else
  \usepackage{cite}
\fi

\ifCLASSINFOpdf
\else
\fi

\usepackage[utf8]{inputenc} 
\usepackage[T1]{fontenc}    
\usepackage{hyperref}       
\usepackage{url}            
\usepackage{booktabs}       
\usepackage{amsfonts}       
\usepackage{nicefrac}       
\usepackage{microtype}      
\usepackage{xcolor}         
\usepackage{graphicx}
\usepackage{subfigure}
\usepackage{amsmath,amssymb} 
\usepackage{caption} 

\usepackage{wrapfig}
\usepackage[american]{babel}
\usepackage[edges]{forest}
\usepackage{tikz}
\usepackage{hyperref}
\usepackage{pifont}

\usepackage{graphicx} 
\usepackage{algorithm}
\usepackage{algorithmic}
\usepackage{times}
\usepackage{soul}
\usepackage{url}
\usepackage[utf8]{inputenc}
\usepackage{amsmath}
\usepackage{amsthm}
\usepackage{booktabs}
\usepackage{algorithm}
\usepackage{algorithmic}
\usepackage{subfigure}
\usepackage{amssymb}
\usepackage{booktabs}
\usepackage{multirow}
\usepackage{makecell}
\usepackage{paralist,algorithmic,algorithm}
\usepackage[american]{babel}
\usepackage{microtype}
\usepackage{lipsum}
\usepackage{bm}
\usepackage{overpic}
\usepackage[switch]{lineno}  %

\usepackage{tikz}
\usepackage[table]{xcolor}
\usepackage{booktabs}

\definecolor{monte_carlo}{RGB}{127,191,156}

\usepackage{xcolor}
\usepackage{tikz}

\newcommand{\heatrowhigh}[3]{%
    \begin{tikzpicture}[baseline]
        \pgfmathsetmacro{\intensity}{%
            (#3==#2 ? 50 : max(0,min(100,100*((#1)-(#2))/((#3)-(#2)))))
        }
        \fill[monte_carlo!\intensity!white, rounded corners=1pt]
             (-0.6em,-0.3em) rectangle (2.6em,1em);
        \node[inner sep=0pt] at (1em,0.7ex) {#1};
    \end{tikzpicture}%
}

\newcommand{\heatrowlow}[3]{%
    \begin{tikzpicture}[baseline]
        \pgfmathsetmacro{\intensity}{%
            (#3==#2 ? 50 : max(0,min(100,100*((#3)-(#1))/((#3)-(#2)))))
        }
        \fill[monte_carlo!\intensity!white, rounded corners=1pt]
             (-0.6em,-0.3em) rectangle (2.6em,1em);
        \node[inner sep=0pt] at (1em,0.7ex) {#1};
    \end{tikzpicture}%
}

\usepackage{bbm}
\usepackage{threeparttable}
\usepackage{amssymb}
\usepackage{pifont}
\usepackage{booktabs}
\usepackage{multirow}
\usepackage{array}
\usepackage{xcolor}
\usepackage{colortbl}   
\usepackage{pgf}        

\definecolor{lightcoral}{rgb}{0.94, 0.5, 0.5}
\definecolor{lightgreen}{rgb}{0.56, 0.93, 0.56}
\definecolor{harvestgold}{rgb}{0.85, 0.57, 0.0}
\definecolor{brightlavender}{rgb}{0.75, 0.58, 0.89}
\definecolor{capri}{rgb}{0.0, 0.75, 1.0}
\definecolor{carminepink}{rgb}{0.92, 0.3, 0.26}
\definecolor{celadon}{rgb}{0.67, 0.88, 0.69}
\definecolor{darkpastelgreen}{rgb}{0.01, 0.75, 0.24}

\definecolor{deepred}{rgb}{0.698,0.133,0.133}
\definecolor{blue}{rgb}{0,0,1}

\usepackage{csquotes}
\begin{document}

%
\title{A Systematic Evaluation of Black-Box Uncertainty Estimation Methods for Large Language Models}
%
%
%
%

\author{Jiayi Wang, Xu-Yao Zhang,~\IEEEmembership{Senior Member,~IEEE}
\thanks{
Jiayi Wang and Xu-Yao Zhang are with the School of Advanced Interdisciplinary Sciences, University of Chinese Academy of Sciences, Beijing 100049, China, and the State Key Laboratory of Multimodal Artificial Intelligence Systems, Institute of Automation, Chinese Academy of Sciences, Beijing 100190, China. Email: wangjiayi2025@ia.ac.cn, xyz@nlpr.ia.ac.cn.}
}

\IEEEtitleabstractindextext{%
\begin{abstract}

Although large language models (LLMs) have shown strong capabilities across a wide range of tasks, their outputs often remain unreliable and may contain hallucinations, making uncertainty estimation (UE) essential for building trustworthy LLMs. In practice, many mainstream LLMs are only accessible through restricted APIs, where internal signals such as logits and hidden states are unavailable, making black-box UE especially important. However, existing work on black-box UE for LLMs remains fragmented in methodology and lacks a unified empirical comparison. To address this gap, we present a systematic review of black-box UE methods and organize them into five categories: verbalization-based, sampling-based, explanation-based, multi-agent, and hybrid methods. We further build a unified evaluation framework and benchmark 24 representative methods across 4 models and 4 dataset settings. Our results show that no single method consistently dominates across all settings. Nevertheless, methods that reason over and compare candidates in the answer space are generally effective, and hybrid methods that combine multiple uncertainty signals perform well under most conditions. By releasing the benchmark data and a unified evaluation framework, we aim to facilitate reproducible comparisons and support future research, while our empirical findings provide practical guidance for developing future black-box UE methods for LLMs.

\end{abstract}
}

\maketitle

\IEEEdisplaynontitleabstractindextext

%
\IEEEpeerreviewmaketitle

\section{Introduction}

Large language models (LLMs) have achieved impressive progress across a wide range of tasks, including question answering, complex reasoning, summarization, and code generation~\cite{achiam2023gpt,liu2024deepseek}. As their capabilities continue to improve, they are being increasingly deployed in real-world applications such as conversational assistants, autonomous agents, and intelligent search systems~\cite{wang2024survey,guo2024large}. Despite these advances, LLMs still suffer from a fundamental reliability problem, as they may produce responses that are fluent and seemingly plausible, yet factually incorrect or insufficiently supported~\cite{huang2025survey,ji2023survey}. This phenomenon, commonly referred to as \emph{hallucination}, may arise from biases in training data, ambiguity in user inputs, or the model's tendency to generate text that appears coherent without being grounded in evidence~\cite{alansari2025large}. In high-stakes domains such as medicine~\cite{goh2024large}, law~\cite{dahl2024large}, and finance~\cite{joshi2025comprehensive}, such failures may not only reduce usability and trust, but also lead to substantial real-world risks~\cite{liu2025scales,wang2024safety}. For this reason, a reliable LLM should not only provide an answer, but also indicate how much that answer should be trusted.

Against this background, uncertainty estimation (UE) has become an important component in building trustworthy LLM systems. Its goal is to estimate a score that reflects how likely a model's output is to be correct, so that unreliable generations can be identified and handled before they cause downstream harm. In practice, such uncertainty signals can support a range of decisions: the system may abstain from answering~\cite{wen2025know}, trigger external verification~\cite{huang2025confrag}, route the query to a stronger model~\cite{zhang2025leveraging}, or defer the case to human experts. UE is also useful beyond risk filtering. For instance, in reasoning-intensive tasks, uncertainty can be used to adaptively control reasoning depth, thereby balancing computational cost and answer reliability~\cite{fu2025deep}. In multi-agent systems, it can guide budget allocation, branch selection, and redundant-interaction reduction~\cite{zhu2026demystifying}. From the perspective of user interaction, explicitly exposing uncertainty or robustness information about model outputs can further improve system transparency and user trust.

\begin{figure*}
    \centering
    \includegraphics[width=0.99\linewidth]{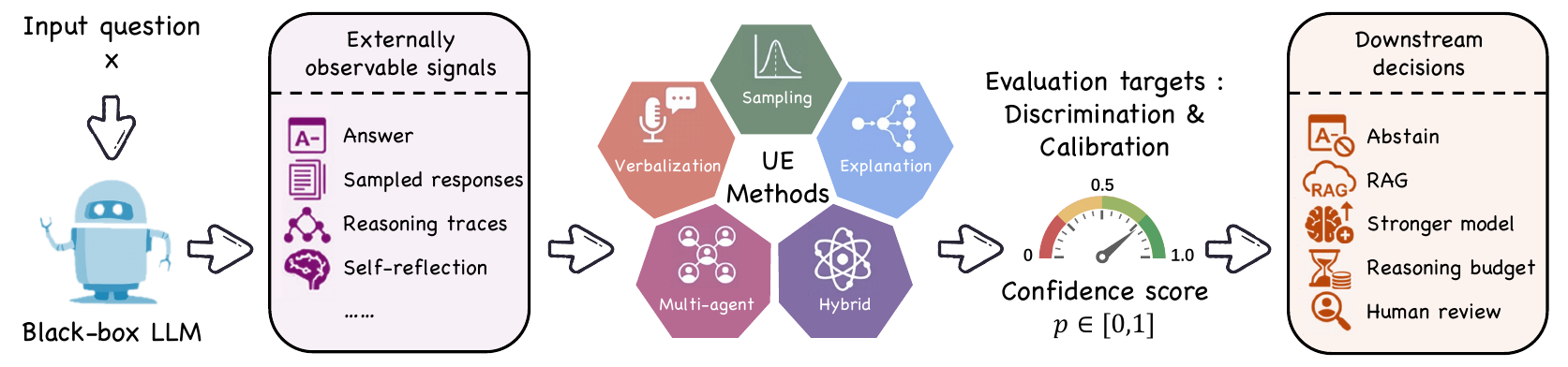}
    \caption{Overview of black-box UE for LLMs.}
    \label{fig:overview}
\end{figure*}

Existing UE methods for LLMs can generally be divided into \emph{white-box} and \emph{black-box} approaches, depending on whether internal model information is accessible. White-box methods typically rely on internal signals such as token-level probability distributions~\cite{duan2024shifting,lin2024contextualized,qiu2024semantic,wang2025genuine}, hidden-state representations~\cite{chen2024inside,zhang2025icr}. By contrast, black-box methods can only use externally observable outputs, such as verbalized confidence scores or sampled responses~\cite{farquhar2024detecting,nikitin2024kernel,xiong2306can,tian2023just}. This distinction is especially important in practice, because many mainstream commercial LLMs are accessible only through restricted APIs, where internal logits and hidden states are unavailable. Under such deployment constraints, black-box UE is often not merely a convenient alternative, but the only feasible option.

Although several recent surveys have discussed UE for LLMs from a broader perspective~\cite{xie2024survey,shorinwa2025survey,vashurin2025benchmarking,liu2025uncertainty,bakman2025reconsidering,bouchard2025uncertainty,xia2025survey}, systematic analysis specifically focused on the black-box setting remains relatively limited. Many existing surveys discuss white-box and black-box methods together, without providing sufficiently fine-grained categorization of black-box approaches or comprehensive coverage of evaluation settings and method types. To address these gaps, this paper focuses on UE for LLMs under black-box settings and aims to provide both a structured methodological review and a unified empirical comparison. Specifically, we study how black-box methods extract uncertainty signals from observable model outputs, how they differ in applicability and underlying modeling mechanisms, and how they perform across different task formats and model families. Figure~\ref{fig:overview} illustrates the overall landscape considered in this work.

Although this paper focuses on black-box UE for LLMs, its implications go beyond this specific setting. Since black-box UE infers uncertainty from externally observable model behavior~\cite{xiong2306can,tian2023just,farquhar2024detecting,nikitin2024kernel}, it provides a behavioral view of model reliability and can further complement white-box signals such as logits and hidden states, while also offering useful guidance for future white-box UE methods. On the other hand, in UE for multimodal large language models (MLLMs), uncertainty in LLM-based text generation remains a fundamental problem, as MLLMs typically rely on language-model components to produce final answers, explanations, and reasoning processes, and many existing MLLM UE methods are also primarily built around the text-generation component~\cite{ruiyang2025,lau2026uncertainty}. Therefore, the taxonomy and empirical findings presented in this paper can provide a valuable starting point for future research on MLLM UE. Building on this foundation, MLLM UE further needs to consider uncertainty sources that are specific to multimodal settings, such as visual perception ambiguity, and cross-modal alignment failures~\cite{bai2024hallucination, liu2024survey}.


Our main contributions are as follows:
\begin{enumerate}
    \item We present a systematic review of black-box UE methods for LLMs and organize them into five categories: verbalization-based, sampling-based, explanation-based, multi-agent, and hybrid methods. This taxonomy offers a clearer and more fine-grained view of the black-box UE landscape.

    \item We build a unified evaluation framework and conduct a systematic comparison of 24 representative black-box UE methods across 4 models and 4 dataset settings. Our evaluation covers both open-ended and closed-ended question answering (QA), and examines method performance from the perspectives of both discrimination and calibration.

    \item We derive several empirical insights into the behavior of black-box UE methods. In open-ended QA, methods based on answer-space reasoning, sampling-aggregated verbalization, and complementary uncertainty signals generally perform well, whereas in closed-ended QA, methods that explicitly exploit the known candidate space show clearer advantages.
    
    \item To support future research, we have released the data and evaluation framework used in this work and maintain a public project page for related papers and resources at \href{https://github.com/wangjy0116/Black-Box-UE-Hub}{Black-Box-UE-Hub}.
\end{enumerate}

\section{Problem Formulation}

This section presents a unified formulation of UE for black-box LLMs. Given an input $x$, a black-box LLM produces a predicted answer $y$ together with a set of externally observable signals, such as verbalized scores, sampled responses, reasoning traces, or multi-agent interaction results. Based on these observable outputs, a UE method assigns a score to quantify the reliability of the current answer.

Different UE methods do not always produce scores with the same interpretation. Some methods produce an \emph{uncertainty} score, denoted by $U(x)$, which characterizes the instability of the model's predictive behavior for the input $x$. In such cases, the final predicted answer $y$ is often obtained separately through an additional low-temperature decoding process. Other methods directly output a \emph{confidence} score, denoted by $C(x,y)$, which reflects the degree to which a specific predicted answer $y$ is believed to be correct, usually in the range $[0,1]$. Usually, a higher uncertainty score corresponds to a lower confidence score, and vice versa. When a method originally produces uncertainty rather than confidence, its output can be transformed monotonically into a confidence score for unified comparison. Therefore, in the following evaluation, we treat the outputs of all methods uniformly as confidence scores.

A useful confidence score should satisfy two key properties. First, it should have strong \emph{ranking}, or equivalently, \emph{discriminative} ability. That is, it should rank correct answers ahead of incorrect ones as much as possible. Ideally, for two samples $(x^{(i)}, y^{(i)})$ and $(x^{(j)}, y^{(j)})$, with corresponding ground-truth answers $\hat{y}^{(i)}$ and $\hat{y}^{(j)}$, we should have
\begin{multline}
C(x^{(i)}, y^{(i)}) \leq C(x^{(j)}, y^{(j)}) \iff \\
p\!\left(y^{(i)}=\hat{y}^{(i)} \mid x^{(i)}\right)
\leq
p\!\left(y^{(j)}=\hat{y}^{(j)} \mid x^{(j)}\right).
\end{multline}
In other words, samples assigned higher confidence should also have a higher probability of being correct. Second, beyond ranking ability, a good confidence score should also be well \emph{calibrated}, meaning that its numerical value should match the empirical correctness rate. Formally, an ideal confidence score should satisfy
\begin{equation}
p(y=\hat{y}\mid C(x,y)=q)=q.
\end{equation}
For example, if a method assigns confidence $0.8$ to a group of samples, then ideally about $80\%$ of the predictions for those samples should be correct.

Overall, the UE task for black-box LLMs can be formulated as follows: under the constraint that only externally observable outputs are available, the goal is to construct either an uncertainty score $U(x)$ or a confidence score $C(x,y)$ for the generated answer. A good UE method should support both discrimination and calibration, enabling more reliable risk control, answer selection, and downstream decision making~\cite{zhang2023survey}.
\section{Taxonomy of Black-box Uncertainty Estimation Methods}

\begin{table*}[!t]
    \centering
    \caption{Summary of black-box UE methods. \textbf{Category} denotes the taxonomy used in this survey; \textbf{Venue} indicates the publication venue and year; \textbf{Computation} indicates relative inference cost; \textbf{Calibration} indicates whether the method directly produces a confidence score in [0,1] without requiring further calibration or normalization; and \textbf{Level} indicates whether the method is applicable to open-ended QA and/or closed-ended QA.}
    \setlength{\tabcolsep}{5pt}
    {\fontsize{6.0pt}{8pt}\selectfont
    \resizebox{0.99\textwidth}{!}{

    \begin{tabular}{
>{\raggedright\arraybackslash}p{6.0cm}
>{\centering\arraybackslash}p{2.0cm}
>{\centering\arraybackslash}p{1.2cm}
>{\centering\arraybackslash}p{1.7cm}
>{\centering\arraybackslash}p{1.25cm}
>{\centering\arraybackslash}p{0.9cm}
>{\centering\arraybackslash}p{0.9cm}
>{\centering\arraybackslash}p{0.4cm}
}
    \toprule
       \multirow{2}{*}{\textbf{Method}} & 
       \multirow{2}{*}{\textbf{Venue}} &
       \multirow{2}{*}{\textbf{Category}} &  
       \multirow{2}{*}{\textbf{Computation}} & 
       \multirow{2}{*}{\textbf{Calibration}} &
       \multicolumn{2}{c}{\textbf{Level}} & 
       \multirow{2}{*}{\textbf{Code}} \\ 
       \cmidrule{6-7}
       & & & & & \makebox[0pt][c]{\textbf{Open-ended}} & \makebox[0pt][c]{\textbf{Closed-ended}} & \\ 
       \midrule

       TopK~\cite{tian2023just} 
       & EMNLP 2023 
       & \multirow{5}{*}{\makebox[0pt][c]{\shortstack[c]{Verbalization-\\based}}} 
       & Low / Medium & \ding{52} & \ding{52} & \ding{52} & - \\ 

       CoT~\cite{xiong2306can} 
       & ICLR 2024 
       &  
       & Low / Medium & \ding{52}  & \ding{52} & \ding{52} 
       & \href{https://github.com/MiaoXiong2320/llm-uncertainty}{Link} \\ 

       Verbalized Probability Distribution (VPD)~\cite{wang2025don} 
       & arXiv 2025
       &  
       & Low / Medium & \ding{52} & \ding{52} & \ding{52} & - \\ 

       Ling~\cite{tian2023just} 
       & EMNLP 2023 
       &  
       & Low / Medium & \ding{52}  & \ding{52} & \ding{52} & - \\ 

       Marker Confidence (MarConf)~\cite{liu2025revisiting} 
       & ACL 2025 
       &  
       & Low / Medium & \ding{52}  &  & \ding{52} & \href{https://github.com/HKUST-KnowComp/MarConf}{Link} \\ 

       \cmidrule{1-8}

       Semantic Entropy (SE)~\cite{farquhar2024detecting} 
       & Nature 2024 
       & \multirow{16}{*}{\makebox[0pt][c]{\shortstack[c]{Sampling-\\based}}} 
       & Medium &   & \ding{52} & \ding{52} 
       & \href{https://github.com/lorenzkuhn/semantic_uncertainty}{Link} \\ 

       SelfCheckGPT (SelfCheck)~\cite{manakul2023selfcheckgpt} 
       & EMNLP 2023 
       &  
       & Medium &  & \ding{52} &  
       & \href{https://github.com/potsawee/selfcheckgpt}{Link} \\ 

       Sum of Eigenvalues of the Graph Laplacian (EigV)~\cite{lin2023generating} 
       & TMLR 2024 
       &  
       & Medium &   & \ding{52} &  
       & \href{https://github.com/zlin7/UQ-NLG}{Link} \\

       The Degree Matrix (Deg)~\cite{lin2023generating} 
       & TMLR 2024 
       &  
       & Medium &   & \ding{52} &  
       & \href{https://github.com/zlin7/UQ-NLG}{Link} \\

       Eccentricity (Ecc)~\cite{lin2023generating} 
       & TMLR 2024 
       &  
       & Medium &   & \ding{52} &  
       & \href{https://github.com/zlin7/UQ-NLG}{Link} \\

       Kernel Language Entropy (KLE)~\cite{nikitin2024kernel} 
       & NeurIPS 2024 
       &  
       & Medium &   & \ding{52} &  
       & \href{https://github.com/AlexanderVNikitin/kernel-language-entropy/tree/main/kle}{Link} \\

       Long-text Uncertainty Quantification (LUQ)~\cite{zhang2024luq} 
       & EMNLP 2024 
       &  
       & Medium & \ding{52}  & \ding{52} &  
       & \href{https://github.com/caiqizh/LUQ}{Link} \\

       Jiang et al. (GU)~\cite{jiang2024graph} 
       & NeurIPS 2024 
       &  
       & Medium & \ding{52}  & \ding{52} &  
       & \href{https://github.com/Mingjianjiang-1/Graph-based-Uncertainty}{Link} \\

       Semantic Embedding Uncertainty (SEU)~\cite{grewal2024improving} 
       & arXiv 2024 
       &  
       & Medium &   & \ding{52} &  & - \\

       Multi-Dimensional Uncertainty Quantification (MDUQ)~\cite{chen2025uncertainty} 
       & arXiv 2025 
       &  
       & Medium &   & \ding{52} &  & - \\

       Convex Hull~\cite{catak2024uncertainty} 
       & DAI 2024 
       &  
       & High &   & \ding{52} &  & - \\

       Semantic INconsistency Index (SINdex)~\cite{abdaljalil2025sindex} 
       & arXiv 2025 
       &  
       & Medium &   & \ding{52} &  & - \\

       Semantic Nearest Neighbor Entropy (SNNE)~\cite{nguyen2025beyond} 
       & ACL Findings 2025 
       &  
       & Medium &   & \ding{52} &  
       & \href{https://github.com/BigML-CS-UCLA/SNNE}{Link} \\
       
       Semantic Structural Entropy (SeSE)~\cite{zhao2025sese} 
       & arXiv 2025 
       &  
       & Medium &   & \ding{52} &  
       & \href{https://github.com/SELGroup/SeSE}{Link} \\
        
       Sampling with Perturbation for Uncertainty Quantification (SPUQ)~\cite{gao2024spuq} 
       & EACL 2024 
       &  
       & Medium & \ding{52} & \ding{52} &  
       & \href{https://github.com/intuit-ai-research/SPUQ/tree/main}{Link} \\

       Inverse-Entropy (InvE)~\cite{song2025inv} 
       & NeurIPS 2025 
       &  
       & Medium &   & \ding{52} &  
       & \href{https://github.com/UMDataScienceLab/Uncertainty-Quantification-for-LLMs}{Link}  \\

       \cmidrule{1-8}

       CoT Explanations (COTA)~\cite{tanneru2024quantifying} 
       & AISTATS 2024 
       & \multirow{7}{*}{\makebox[0pt][c]{\shortstack[c]{Explanation-\\based}}} 
       & Medium & \ding{52}  & \ding{52} & \ding{52} & - \\

       Think twice before trusting (T3)~\cite{li2024think} 
       & EMNLP Findings 2024 
       &  
       & Medium & \ding{52}  &  & \ding{52} & - \\

       Topo-UQ~\cite{da2025understanding} 
       & COLM 2025 
       &  
       & Medium &   & \ding{52} & \ding{52}
       & \href{https://github.com/LongchaoDa/LLM-Topology}{Link} \\

       Introspective UQ (IUQ)~\cite{mei2025reasoning} 
       & arXiv 2025 
       &  
       & Low / Medium & \ding{52}  & \ding{52} & \ding{52} & - \\

       CenConf~\cite{zhang2025all} 
       & EMNLP 2025 
       &  
       & Medium & \ding{52}  & \ding{52} & \ding{52} & - \\

       PathConv~\cite{zhang2025all} 
       & EMNLP 2025 
       &  
       & Medium & \ding{52}  & \ding{52} & \ding{52} & - \\

       PathWeight~\cite{zhang2025all} 
       & EMNLP 2025 
       &  
       & Medium & \ding{52}  & \ding{52} & \ding{52} & - \\

       \cmidrule{1-8}

       CollabCalibration (Collab)~\cite{yang2024confidence} 
       & ICLR Workshop 2024 
       & \multirow{3}{*}{Multi-agent} 
       & High & \ding{52}  & \ding{52} & \ding{52} 
       & \href{https://github.com/minnesotanlp/collaborative-calibration}{Link} \\

       ArgLLMs~\cite{freedman2025argumentative} 
       & AAAI 2025 
       &  
       & Medium & \ding{52}  & \ding{52} & \ding{52} 
       & \href{https://github.com/CLArg-group/argumentative-llms}{Link} \\

       DiverseAgentEntropy (DAE)~\cite{feng2024rethinking} 
       & EMNLP Findings 2025
       &  
       & High &  & \ding{52} & \ding{52} 
       & \href{https://github.com/amazon-science/DiverseAgentEntropy}{Link} \\

       \cmidrule{1-8}

       BSDetector~\cite{chen2024quantifying} 
       & ACL 2024 
       & \multirow{4}{*}{Hybrid} 
       & Medium & \ding{52}  & \ding{52} &  & - \\

       UF Calibration (UF)~\cite{zhang2024calibrating} 
       & EMNLP 2024 
       &  
       & Medium & \ding{52}  &  & \ding{52} & - \\

       SteerConf~\cite{zhou2025steerconf} 
       & NeurIPS 2025 
       &  
       & Medium & \ding{52}  & \ding{52} & \ding{52} 
       & \href{https://github.com/scottjiao/SteerConf}{Link} \\

       Distractor-Normalized Coherence (DiNCo)~\cite{wang2025calibrating} 
       & ICLR 2026 
       &  
       & Medium & \ding{52}  & \ding{52} &  
       & \href{https://github.com/victorwang37/dinco}{Link} \\       

    \bottomrule
    \end{tabular}}}
    \label{tab:llm}
\end{table*}

We organize black-box UE methods according to their primary information sources and modeling strategies into five categories: verbalization-based, sampling-based, explanation-based, multi-agent, and hybrid methods, as summarized in Table~\ref{tab:llm}. It is worth emphasizing that sampling is a common operation in many black-box methods, and therefore the use of sampling itself is not the main criterion for classification in this paper. Instead, we reserve the term "sampling-based" for methods that estimate uncertainty by analyzing the variation or similarity across multiple responses obtained through repeated or perturbed generation.

\subsection{Verbalization-based}

\label{Sec:verb}

Verbalization-based methods estimate uncertainty by explicitly eliciting the model's own assessment of how likely its answer is to be correct. Their central idea is that uncertainty can be expressed directly in the model's output, either in numeric form or through natural-language cues. Accordingly, we divide this family into two subtypes: \emph{numeric verbalization}, which asks the model to report an explicit probability or score, and \emph{linguistic verbalization}, which infers confidence from hedging expressions or epistemic language.

\subsubsection{Numeric verbalization}

Numeric verbalization methods prompt the model to output a scalar, typically in $[0,1]$ or $[0,100]$, representing its subjective probability that the answer is correct. For example, after producing an answer, the model may continue with outputs such as "Confidence: $0.78$" or "Probability of correctness: $82\%$". Because these methods directly return a numeric score, they are naturally compatible with thresholding, ranking, and calibration analysis.

A representative approach is \textbf{CoT} prompting~\cite{xiong2306can}. This method introduces explicit reasoning into numeric verbalization by asking the model to first generate a reasoning chain together with a final answer and then report the corresponding confidence, following a "think first, then judge" process. The underlying intuition is that making the reasoning process explicit may help the model form a more informed self-assessment. This approach can also be combined with repeated sampling to improve UE stability.

\textbf{TopK}~\cite{tian2023just} extends numeric verbalization from scoring a single answer to scoring $K$ candidate answers. Specifically, it prompts the model to generate $K$ candidate answers together with their associated confidence scores in a single output, for example, "Answer 1: A ($0.70$), Answer 2: B ($0.30$), Answer 3: C ($0.20$)." This allows the model to distribute belief across multiple candidate answers rather than expressing confidence in only one answer, thereby providing a richer view of uncertainty and helping reduce overconfidence.

Recently, \textbf{VPD}~\cite{wang2025don} further extends this idea. Similar to TopK, it asks the model to output multiple candidate answers, but in open-ended QA, it additionally introduces a \emph{"None of the above"} option and constrains the confidence scores assigned to all candidates to sum to $1.0$. For example, the model may output "Paris: $0.62$, London: $0.18$, Berlin: $0.05$, None of the above: $0.15$". This extends numeric verbalization from a local confidence judgment to a full confidence distribution over the answer space.

\subsubsection{Linguistic verbalization}
Unlike numeric verbalization, linguistic verbalization methods do not require the model to output an explicit probability. Instead, they treat naturally occurring epistemic expressions, such as hedging, cautious phrasing, or explicit statements of uncertainty, as proxy signals of confidence. For example, a model may respond with expressions such as "I am not very sure" or "We believe". Although such expressions do not directly correspond to numeric probabilities, they still convey useful information about how certain the model is about its answer.

The main challenge in this line of work is that linguistic expressions are not directly comparable across samples. Therefore, an additional quantization or calibration step is usually required to map natural-language uncertainty expressions into comparable numeric scores. Common strategies include assigning human-annotated confidence values to specific expressions~\cite{tian2023just}, learning a mapping from expressions to scores on a small calibration set~\cite{tian2023just, liu2025revisiting}, or using an LLM-as-a-judge to reinterpret the expression in context and infer a corresponding score~\cite{yona2024can}. For example, an output such as "I am not very sure" may subsequently be mapped to a score such as $0.3$ using manual rules or an auxiliary model.

Overall, verbalization-based methods are attractive because they are simple, low-cost, and directly produce confidence scores that are easy to use in downstream systems. Their main weakness is that the reported confidence may reflect the model's learned communication style rather than its true epistemic state~\cite{yona2024can,liu2025revisiting}. As a result, these methods can be sensitive to prompt wording, instruction style, and the model's tendency toward overconfidence or excessive caution.

\subsection{Sampling-based}
\label{Sec:sampling}

Sampling-based methods estimate uncertainty by generating multiple responses for the same input, either directly or under controlled perturbations, and then analyzing their variation, similarity, or stability. The intuition is that if the model produces similar answers across repeated generations, its prediction is more likely to be reliable; in contrast, if the responses vary substantially, the model is likely to be more uncertain. According to whether the responses are generated from the original input or from perturbed variants of the input, these methods can be divided into two categories: \emph{direct sampling} and \emph{perturbation-based sampling}.

\subsubsection{Direct sampling}

Direct sampling methods estimate uncertainty by repeatedly
generating multiple responses for the same input $x$ and analyzing
their semantic relations. Let
$\mathcal{Y}(x)=\{y_1,\ldots,y_M\}$ denote the $M$ sampled
responses generated for input $x$. To make such comparisons
computable, a pairwise semantic relation function is applied to
the generated text outputs. Specifically, for a relation type $\psi$,
let $s_\psi(y_i,y_j)$ denote the semantic relation between two
sampled responses $y_i$ and $y_j$. The corresponding relation
matrix is defined as
\begin{equation}
S_{\mathcal{Y}}^{(\psi)} \in \mathbb{R}^{M\times M}, \qquad
\left(S_{\mathcal{Y}}^{(\psi)}\right)_{ij}=s_\psi(y_i,y_j).
\end{equation}
Here, $s_\psi(\cdot,\cdot)$ is computed by external auxiliary
models over the generated text, rather than by accessing the
internal states of the target black-box LLM. Therefore, this
formulation remains consistent with the black-box setting, as it
does not require logits, hidden states, or any other internal
information from the target model. The resulting relation matrix
provides structured pairwise semantic signals for measuring
response consistency, disagreement, or semantic diversity.

In this survey, we mainly consider two common instantiations of
$s_\psi(\cdot,\cdot)$. The first is the \emph{NLI-based semantic
relation}, which uses an auxiliary \emph{Natural Language
Inference} (NLI) model to determine whether the semantic content
of one response entails, is neutral to, or contradicts that of another
response. Specifically, $y_i$ is treated as the premise and $y_j$ as
the hypothesis, and the NLI model performs three-way classification
over \emph{entailment}, \emph{neutral}, and \emph{contradiction}.
The auxiliary NLI model first outputs a logit vector
\begin{equation}
\boldsymbol{\ell}_{\mathrm{NLI}}(y_i,y_j)
=
\bigl[
\ell_{\mathrm{ent}}(y_i,y_j),
\ell_{\mathrm{neu}}(y_i,y_j),
\ell_{\mathrm{con}}(y_i,y_j)
\bigr],
\end{equation}
where $\ell_{\mathrm{ent}}$, $\ell_{\mathrm{neu}}$, and
$\ell_{\mathrm{con}}$ denote the logits of the auxiliary NLI
classifier for the entailment, neutral, and contradiction classes,
respectively. They are converted into class probabilities by softmax:
\begin{equation}
p_{\mathrm{NLI}}(\zeta\mid y_i,y_j)
=
\frac{\exp(\ell_{\zeta}(y_i,y_j))}
{\sum_{\zeta'\in\{\mathrm{ent},\mathrm{neu},\mathrm{con}\}}
\exp(\ell_{\zeta'}(y_i,y_j))},
\end{equation}
where $\zeta\in\{\mathrm{ent},\mathrm{neu},\mathrm{con}\}$ denotes
an NLI label. The label with the largest probability is the predicted
NLI class. Since NLI relations are directional, for each NLI label
\(\zeta\in\{\mathrm{ent},\mathrm{neu},\mathrm{con}\}\), we compute
the relation in both directions and define a symmetric NLI relation
score as
\begin{equation}
s_{\zeta}(y_i,y_j)
=
\frac{1}{2}
\Bigl(
p_{\mathrm{NLI}}(\zeta\mid y_i,y_j)
+
p_{\mathrm{NLI}}(\zeta\mid y_j,y_i)
\Bigr).
\end{equation}
The corresponding NLI-based relation matrix for label \(\zeta\) is
then defined as
\begin{equation}
S_{\mathcal{Y}}^{(\zeta)}\in\mathbb{R}^{M\times M},
\qquad
\bigl(S_{\mathcal{Y}}^{(\zeta)}\bigr)_{ij}
=
s_{\zeta}(y_i,y_j).
\end{equation}

The second instantiation is the \emph{embedding-based cosine
similarity}, which measures continuous semantic similarity between
sampled responses. Each sampled response $y_i$ is fed into
Sentence-BERT to obtain a sentence embedding $e(y_i)$. The semantic similarity between two responses is
then computed by cosine similarity in the embedding space:
\begin{equation}
s_{\mathrm{cos}}(y_i,y_j)
=
\frac{e(y_i)^\top e(y_j)}
{\|e(y_i)\|_2\,\|e(y_j)\|_2}.
\end{equation}
The corresponding embedding-based cosine similarity matrix over
sampled responses is
\begin{equation}
S_{\mathcal{Y}}^{(\mathrm{cos})}\in\mathbb{R}^{M\times M},
\qquad
\bigl(S_{\mathcal{Y}}^{(\mathrm{cos})}\bigr)_{ij}
=
s_{\mathrm{cos}}(y_i,y_j).
\end{equation}

\textbf{SE}~\cite{farquhar2024detecting} (Semantic Entropy) groups the sampled responses $\mathcal{Y}(x)$ into semantic clusters $\mathcal{Z}=\{z_1,\dots,z_K\}$ based on bidirectional entailment. Specifically, two responses $y_i$ and $y_j$ are placed in the same cluster if the NLI model predicts entailment in both directions. Let $|z_k|$ denote the number of responses in the $k$-th cluster. Its empirical frequency is defined as
\begin{equation}
p_k=\frac{|z_k|}{M}.
\end{equation}
The SE uncertainty score is then computed as
\begin{equation}
U_{\mathrm{SE}}(x)=-\sum_{k=1}^{K}p_k\log p_k.
\end{equation}
Thus, uncertainty is low when sampled responses concentrate in a few semantic clusters, and high when they are distributed across many clusters.

\textbf{SNNE}~\cite{nguyen2025beyond} (Semantic Nearest Neighbor Entropy) avoids the hard clustering step in SE and instead computes uncertainty directly from pairwise entailment scores. Its uncertainty is defined as
\begin{equation}
\label{eq:snne}
U_{\mathrm{SNNE}}(x)
=
-\frac{1}{M}\sum_{i=1}^{M}
\log\sum_{j=1}^{M}\exp\!\left(\frac{s_{\mathrm{ent}}(y_i, y_j)}{\tau}\right),
\end{equation}
where \(\tau\) is a temperature scaling parameter. The inner summation in Eq.~\eqref{eq:snne} effectively accounts for both intra- and inter-cluster similarities. Compared with SE, SNNE can provide a smoother uncertainty signal when responses only partially overlap in meaning.

Since $S^{(\mathrm{ent})}_{\mathcal{Y}}$ can be viewed as a weighted adjacency matrix, the degree matrix can be defined as
\begin{equation}
(D^{(\mathrm{ent})}_{\mathcal{Y}})_{ii}=\sum_{j=1}^{M}(S^{(\mathrm{ent})}_{\mathcal{Y}})_{ij}.
\end{equation}
The normalized graph Laplacian is then defined as
\begin{equation}
L^{(\mathrm{ent})}_{\mathrm{norm}}
=
I-\bigl(D^{(\mathrm{ent})}_{\mathcal{Y}}\bigr)^{-1/2}S^{(\mathrm{ent})}_{\mathcal{Y}}\bigl(D^{(\mathrm{ent})}_{\mathcal{Y}}\bigr)^{-1/2}.
\end{equation}
where $I$ is the identity matrix.
\textbf{EigV}~\cite{lin2023generating} (Sum of Eigenvalues of the Graph Laplacian) defines uncertainty using the eigenvalues $\{\lambda_i\}_{i=1}^{M}$ of $L^{(\mathrm{ent})}_{\mathrm{norm}}$:
\begin{equation}
U_{\mathrm{Eig}}(x)=\sum_{i=1}^{M}\max(0,1-\lambda_i).
\end{equation}
\textbf{Deg}~\cite{lin2023generating} (The Degree Matrix) measures uncertainty from node connectivity and defines
\begin{equation}
U_{\mathrm{Deg}}(x)=\frac{\mathrm{Tr}(MI-D^{(\mathrm{ent})}_{\mathcal{Y}})}{M^2},
\end{equation}
where $\mathrm{Tr}(\cdot)$ denotes the trace operator.
\textbf{Ecc}~\cite{lin2023generating} (Eccentricity) takes the first $r$ eigenvectors $u_1,\dots,u_r$ of $L^{(\mathrm{ent})}_{\mathrm{norm}}$, and defines the spectral embedding of the $j$-th response as
\begin{equation}
v_j=[u_{1,j},\dots,u_{r,j}]^\top.
\end{equation}
Its centered representation is then defined as
\begin{equation}
\tilde v_j=v_j-\frac{1}{M}\sum_{j'=1}^{M}v_{j'},
\end{equation}
and the uncertainty is given by
\begin{equation}
U_{\mathrm{Ecc}}(x)=\frac{1}{M}\sum_{j=1}^{M}\|\tilde v_j\|_2.
\end{equation}
Taken together, these three methods transform semantic entailment relations among sampled responses into a graph and then derive uncertainty from different graph properties, namely spectral structure, node connectivity, or embedding geometry.

\textbf{KLE}~\cite{nikitin2024kernel} (Kernel Language Entropy) is built on the
semantic support matrix
\begin{equation}
S^{(\mathrm{sup})}_{\mathcal{Y}}
=
S^{(\mathrm{ent})}_{\mathcal{Y}}
+
\frac{1}{2}S^{(\mathrm{neu})}_{\mathcal{Y}},
\end{equation}
where entailment is treated as full semantic support and neutral as partial
semantic support. Based on this matrix, the degree matrix is defined as
\begin{equation}
(D^{(\mathrm{sup})}_{\mathcal{Y}})_{ii}
=
\sum_{j=1}^{M}
(S^{(\mathrm{sup})}_{\mathcal{Y}})_{ij}.
\end{equation}
KLE then uses the unnormalized graph Laplacian
\begin{equation}
L^{(\mathrm{sup})}_{\mathrm{KLE}}
=
D^{(\mathrm{sup})}_{\mathcal{Y}}
-
S^{(\mathrm{sup})}_{\mathcal{Y}},
\end{equation}
and constructs the heat kernel
\begin{equation}
K_t
=
\exp\bigl(-tL^{(\mathrm{sup})}_{\mathrm{KLE}}\bigr).
\end{equation}
To obtain a unit-trace positive semidefinite kernel, KLE further normalizes
$K_t$ entry-wise as
\begin{equation}
K_{\mathrm{norm}}(i,j)
=
\frac{
K_t(i,j)
}{
\sqrt{K_t(i,i)\,K_t(j,j)}\,M
}.
\end{equation}
The uncertainty is then defined as the von Neumann entropy of the normalized
kernel:
\begin{equation}
U_{\mathrm{KLE}}(x)
=
-\mathrm{Tr}
\left(
K_{\mathrm{norm}}\log K_{\mathrm{norm}}
\right).
\end{equation}
Compared with entropy over discrete clusters, KLE measures uncertainty through
the spectral entropy of a continuous kernel operator, and therefore provides a
finer-grained characterization of semantic diversity.

\textbf{SeSE}~\cite{zhao2025sese} (Semantic Structural Entropy) extends semantic UE from flat clustering to hierarchical semantic structure. Given sampled responses $\mathcal{Y}(x)$, it constructs a directed semantic graph $\mathcal{G}=(\mathcal{V},\mathcal{E})$ that preserves the direction of NLI relations. The edge weight from $y_i$ to $y_j$ is defined as
\begin{equation}
w_{ij}
=
p_{\mathrm{NLI}}(\mathrm{ent}\mid y_i,y_j)
+
\frac{1}{2}
p_{\mathrm{NLI}}(\mathrm{neu}\mid y_i,y_j).
\end{equation}
Let $P \in \mathbb{R}^{M \times M}$ be the row-normalized transition matrix,
\begin{equation}
P_{ij} = \frac{w_{ij}}{\sum_{k=1}^{M} w_{ik}},
\end{equation}
and let $\pi$ be the stationary distribution satisfying $\pi P = \pi$. Based on this random-walk view, SeSE searches for an optimal hierarchical encoding tree ${T}_{\mathcal{G}}$ over the semantic graph $\mathcal{G}$. For any non-root tree node $v \in {T}_{\mathcal{G}}$, let $v^{-}$ denote its parent, let $V_{v}$ denote the random-walk volume of the subtree rooted at $v$, and let $g_{v}$ denote the total incoming flow from outside that subtree. The structural entropy contribution of node $v$ is defined as
\begin{equation}
H_{{T}_{\mathcal{G}}}(\mathcal{G};v)
=
-\frac{g_{v}}{\mathrm{vol}(\mathcal{G})}
\log
\frac{V_{v}}{V_{v^{-}}},
\end{equation}
where $\mathrm{vol}(\mathcal{G})$ is the graph volume. The SeSE uncertainty score is then defined as the minimum total structural entropy over all encoding trees whose height does not exceed $d_{\max}$:
\begin{equation}
U_{\mathrm{SeSE}}(x)
=
\min_{\mathrm{height}({T}_{\mathcal{G}}) \le d_{\max}}
\sum_{v \in {T}_{\mathcal{G}},\, v \neq v_r}
H_{{T}_{\mathcal{G}}}(\mathcal{G};v),
\end{equation}
where $v_r$ is the root node and $d_{\max}$ is the maximum tree depth. Intuitively, if the sampled responses form a semantic space with clear hierarchical regularities, the graph can be compressed more efficiently and the resulting structural entropy is lower; if the responses are semantically scattered and lack coherent organization, the entropy becomes higher. Therefore, SeSE can be viewed as a hierarchical generalization of SE. While SE
measures uncertainty over one-shot semantic partitions, SeSE further captures
multi-level semantic organization within the sampled response set.

\textbf{MDUQ}~\cite{chen2025uncertainty} (Multi-Dimensional Uncertainty Quantification) assumes that responses may show different consistency patterns at the surface semantic level and at the factual knowledge level. To capture this, the method not only computes semantic relations on the original sampled responses $\mathcal{Y}(x)=\{y_1,\dots,y_M\}$, but also extracts the corresponding knowledge from each response and denotes the resulting set as $\mathcal{Y}'(x)=\{y'_1,\dots,y'_M\}$, where $y'_i$ denotes the knowledge or factual representation extracted from response $y_i$. MDUQ then constructs the semantic relation matrix
\begin{equation}
(S^{(\mathrm{ent})}_{\mathcal{Y}})_{ij}=s_{\mathrm{ent}}(y_i,y_j),
\end{equation}
and the knowledge relation matrix
\begin{equation}
(S^{(\mathrm{ent})}_{\mathcal{Y'}})_{ij}=s_{\mathrm{ent}}(y'_i,y'_j).
\end{equation}
These two relation matrices are then stacked along a new dimension to form a third-order tensor $\mathcal{S}\in\mathbb{R}^{M\times M\times 2}$. Based on this tensor, MDUQ applies both Canonical Polyadic (CP) decomposition and Tucker decomposition for low-rank modeling, and uses the normalized reconstruction error as the basis for uncertainty:
\begin{equation}
U_{\mathrm{MDUQ}}(x)
=
\phi\!\left(
\frac{\|\mathcal{S}-\hat{\mathcal{S}}^{\mathrm{CP}}_{R}\|_F}{\|\mathcal{S}\|_F},
\frac{\|\mathcal{S}-\hat{\mathcal{S}}^{\mathrm{Tucker}}_{R}\|_F}{\|\mathcal{S}\|_F}
\right)
\end{equation}
where $\phi(\cdot,\cdot)$ is an aggregation function, which can be $\min$ or $\mathrm{sum}$, and $\hat{\mathcal{S}}^{\mathrm{CP}}_{R}$ and $\hat{\mathcal{S}}^{\mathrm{Tucker}}_{R}$ denote the rank-$R$ reconstructions from CP and Tucker decomposition, respectively. A larger reconstruction error means that this set of responses is harder to explain with a shared low-rank structure, and therefore indicates higher uncertainty.

\textbf{SelfCheck}~\cite{manakul2023selfcheckgpt} (SelfCheckGPT) estimates the
factuality of each sentence in the predicted answer $y$ by comparing it with
multiple randomly sampled responses. Specifically, let $o_j$ denote the $j$-th
sentence in the predicted answer $y$, and let $y_i$ denote the $i$-th sampled
response. For the pair $(y_i,o_j)$, SelfCheck-NLI treats $y_i$ as the premise and
$o_j$ as the hypothesis. Different from the three-way NLI probability defined
above, SelfCheck-NLI computes a binary softmax over only the entailment and
contradiction logits:
\begin{equation}
\hat{p}_{\mathrm{NLI}}(\mathrm{con}\mid y_i, o_j)
=
\frac{
\exp(\ell_{\mathrm{con}}(y_i,o_j))
}{
\exp(\ell_{\mathrm{ent}}(y_i,o_j))
+
\exp(\ell_{\mathrm{con}}(y_i,o_j))
}.
\end{equation}
Based on this, SelfCheck defines the average contradiction between sentence
$o_j$ and all sampled responses as the sentence-level uncertainty:
\begin{equation}
U_{\mathrm{SelfCheck}}(x,o_j)
=
\frac{1}{M}
\sum_{i=1}^{M}
\hat{p}_{\mathrm{NLI}}(\mathrm{con}\mid y_i, o_j).
\end{equation}
A higher score means that the sentence is less consistent with the sampled
responses, and is therefore more likely to contain hallucinated content.

\textbf{LUQ}~\cite{zhang2024luq} (Long-text Uncertainty Quantification) is similar to the SelfCheck method. For a sentence in a sampled response, LUQ estimates its uncertainty by using the degree to which its content is entailed by other sampled responses. In addition, LUQ introduces a more fine-grained variant, \textbf{LUQ-ATOMIC}, which first breaks the response into atomic facts and then measures the consistency among different sampled responses at the atomic fact level. Compared with sentence-level modeling, this method can capture local factual inconsistencies in long text more precisely, and is therefore more suitable for factuality and uncertainty analysis in long-form text.

\textbf{GU}~\cite{jiang2024graph} is also a UE method for long-form generation. It first decomposes sampled responses into a set of factual claims, and then constructs a bipartite graph between response nodes and claim nodes based on whether a response supports a claim:
\begin{equation}
\mathcal{G}=((\mathcal{Y},\mathcal{F}),\mathcal{E}),
\end{equation}
where $\mathcal{Y}$ is the set of sampled responses, and $\mathcal{F}$ is the set of deduplicated factual claims. If a response $y_i\in\mathcal{Y}$ supports a claim $f_j\in\mathcal{F}$, then an edge $(y_i,f_j)\in\mathcal{E}$ is added. On this graph, the centrality of each claim is used as an uncertainty-related score. Among different centrality measures, the original paper finds that closeness centrality performs the best. Its basic form is
\begin{equation}
C_{\mathrm{GU}}(x,f)=
\frac{|\mathcal{V}|-1}{\sum_{u\in\mathcal{V}} dist(f,u)} \cdot \frac{|\mathcal{V}|}{|\mathcal{V}_f|},
\end{equation}
where $\mathcal{V}$ is the set of all nodes in the graph, $\mathcal{V}_f$ is the set of nodes connected to $f$, and $dist(f,u)$ is the shortest-path distance in the graph. Intuitively, true claims are usually closer to more responses and other claims, and therefore have higher centrality. In contrast, false claims are often more peripheral, which leads to lower centrality and higher uncertainty.

\textbf{SEU}~\cite{grewal2024improving} (Semantic Embedding Uncertainty) estimates uncertainty based on the semantic consistency among multiple sampled responses. Specifically, it first computes the pairwise cosine similarity between all sampled answers, and then uses the inverse of their average similarity as the uncertainty score:
\begin{equation}
U_{\mathrm{SEU}}(x)
=
1-\frac{2}{M(M-1)}\sum_{1\le i<j\le M}s_{\mathrm{cos}}(y_i, y_j).
\end{equation}

\textbf{Convex Hull}~\cite{catak2024uncertainty} further combines continuous semantic-space modeling with clustering, and measures uncertainty through the \emph{geometric spread} of sampled responses in the embedding space. Each sampled response $y_i$ is mapped to a semantic embedding $e(y_i)\in\mathbb{R}^d$, and the resulting embeddings are projected to a two-dimensional space by PCA for visualization and clustering. Let $\tilde{e}(y_i)\in\mathbb{R}^2$ denote the projected point of response $y_i$. The method next applies a density-based clustering algorithm such as DBSCAN~\cite{ester1996density} to the set of projected points $\{\tilde{e}(y_i)\}_{i=1}^{M}$, and obtains the set of non-noise clusters $\mathcal{Z}=\{z_1,\dots,z_K\}$. For each cluster $z_k$, let $E_{z_k}$ denote the corresponding set of two-dimensional points. The uncertainty is then defined as the sum of convex hull areas of all non-noise clusters:
\begin{equation}
U_{\mathrm{ConvexHull}}(x)
=
\sum_{k=1}^{K}
\operatorname{Area}\!\left(
\operatorname{ConvHull}\!\left(E_{z_k}\right)
\right).
\end{equation}
Here, $\operatorname{ConvHull}(E_{z_k})$ is the smallest convex polygon enclosing all points in cluster $z_k$, and its area measures the spatial extent of that cluster in the projected semantic space. If the sampled responses occupy a larger semantic region, or if they form multiple separated clusters with a larger total convex hull area, then the response set is more dispersed, which indicates higher uncertainty.

\textbf{SINdex}~\cite{abdaljalil2025sindex} (Semantic INconsistency Index) performs hierarchical clustering on the semantic vectors. Unlike SE, which computes entropy only from cluster frequencies, SINdex further uses the average within-cluster similarity to adjust the weight of each cluster:
\begin{equation}
p_k' = p_k \cdot \frac{2}{|z_k|(|z_k|-1)} \sum_{u,v \in z_k,\; u \ne v} s_{\mathrm{cos}}(u,v).
\end{equation}
Here, $p_k$ denotes the empirical frequency of the $k$-th semantic cluster. Based on this, the uncertainty is defined as
\begin{equation}
U_{\mathrm{SINdex}}(x)= -\sum_{k=1}^{K} p_k' \log p_k'.
\end{equation}
Compared with semantic entropy based only on cluster frequency, SINdex can reflect uncertainty from both \emph{between-cluster dispersion} and \emph{within-cluster looseness}.

\subsubsection{Perturbation-based sampling}

Perturbation-based sampling aims to model both aleatoric uncertainty and epistemic uncertainty. The former is mainly reflected in the diversity of the outputs themselves, while the latter reflects how sensitive the model is to small input changes. Direct sampling methods mainly estimate uncertainty through the differences among multiple outputs, and therefore focus more on aleatoric uncertainty. In contrast, perturbation-based sampling further studies prediction stability by perturbing prompts or temperature, and thus adds information about epistemic uncertainty.

\textbf{SPUQ}~\cite{gao2024spuq} (Sampling with Perturbation for Uncertainty Quantification) estimates the uncertainty by introducing both temperature perturbations and prompt perturbations. Temperature perturbation refers to using a temperature different from the original setting during generation, either keeping it fixed across all perturbed inputs or randomly sampling it from a predefined range. Prompt perturbation includes input rewriting, inserting virtual tokens into the prompt, and rewriting the system message. Given an original input $x$, let $x_0=x$, and let $\mathcal{X}(x)=\{x_1,\dots,x_M\}$ denote the set of perturbed inputs constructed from $x$. Let $\mathcal{T}(x)=\{t_1,\dots,t_M\}$ denote the corresponding set of temperatures. For each $x_i$, the model generates an output $y_i$ under temperature $t_i$, where $y_0$ is the output generated from the original input $x$ under temperature $t_0$. Based on these outputs, SPUQ evaluates the confidence of the original answer $y_0$ through inter-sample aggregation:
\begin{equation}
C_{\mathrm{SPUQ\text{-}inter}}(x,y_0)
=
\frac{\sum_{i=1}^{M} s_{\mathrm{cos}}(y_0,y_i)\,s_{\mathrm{cos}}(x_0,x_i)}
{\sum_{i=1}^{M}s_{\mathrm{cos}}(x_0,x_i)}.
\end{equation}
This score measures the consistency of the model's answer under different perturbations. If the model is highly sensitive to small input perturbations, so that its outputs change substantially across perturbed inputs, then its epistemic uncertainty is high. Besides inter-sample aggregation, SPUQ also supports intra-sample aggregation, where the confidence of each output is first estimated separately and then averaged:
\begin{equation}
C_{\mathrm{SPUQ\text{-}intra}}(x)
=
\frac{1}{M+1}\sum_{i=0}^{M} C_{\mathrm{VC}}(x_i,y_i).
\end{equation}
Here, $C_{\mathrm{VC}}(x_i,y_i)$ denotes the model's self-reported verbalized confidence in output $y_i$ under input $x_i$.

\textbf{InvE}~\cite{song2025inv} (Inverse-Entropy) estimates the uncertainty of LLMs from an inverse probability modeling perspective. This method considers both input perturbations and the semantic structure of the corresponding outputs. Given the original input $x=x_0$, let $y_0$ denote the corresponding output. Let $\mathcal{X}(x)=\{x_1,\dots,x_M\}$ denote the set of perturbed inputs constructed from $x$, and let $\mathcal{Y}(x)=\{y_1,\dots,y_M\}$ denote the corresponding set of outputs, where each $y_i$ is generated from $x_i$. InvE then computes pairwise semantic similarity in both the input space and the output space, and constructs two random-walk transition matrices:
\begin{equation}
\begin{aligned}
P^X_{ij}&=\frac{s_{\mathrm{cos}}(x_i,x_j)}{\sum_{k=0}^{M} s_{\mathrm{cos}}(x_i,x_k)},\\
P^Y_{ij}&=\frac{s_{\mathrm{cos}}(y_i,y_j)}{\sum_{k=0}^{M} s_{\mathrm{cos}}(y_i,y_k)}.
\end{aligned}
\end{equation}
Here, $P^X$ captures the local semantic structure among perturbed inputs, while $P^Y$ captures the semantic structure among the corresponding outputs. Based on these two structures, InvE further models the inverse conditional distribution $P(X\mid Y)$ and uses its conditional entropy to estimate uncertainty:
\begin{equation}
U_{\mathrm{InvE}}(x)
=
-\sum_{i=0}^{M} p(x_i\mid y_i)\log p(x_i\mid y_i).
\end{equation}
A larger value indicates that the structural correspondence between perturbed inputs and outputs is more mixed, suggesting that the model's prediction is less stable and thus more uncertain.

Overall, sampling-based methods examine whether the model remains stable under repeated or perturbed generation. They are often more robust than single-pass self-reported confidence, especially when semantic disagreement among samples is informative. However, they require multiple model calls and depend heavily on the quality of the relation function used to compare responses. In open-ended tasks, surface diversity may not always imply factual disagreement, while semantically similar responses may still differ in crucial details.

\subsection{Explanation-based}
\label{Sec:explanation}

Explanation-based methods estimate uncertainty from the model's reasoning process rather than from the final answer alone. Their core assumption is that uncertainty may be reflected in the completeness or internal consistency of the generated reasoning chain. Accordingly, these methods either analyze weaknesses within a single reasoning process or compare multiple sampled reasoning processes to quantify confidence.

\textbf{COTA}~\cite{tanneru2024quantifying} (CoT Explanations) first records the original reasoning chain $R$ used to generate the predicted answer, and then obtains an additional set of sampled reasoning chains $\{R^{(m)}\}_{m=1}^{M}$. Let
\begin{equation}
R^{(a)}=\{r^{(a)}_1,\dots,r^{(a)}_{L_a}\}, \qquad
R^{(b)}=\{r^{(b)}_1,\dots,r^{(b)}_{L_b}\},
\end{equation}
where $r^{(a)}_i$ denotes the $i$-th reasoning step in chain $R^{(a)}$. Using the NLI model to judge whether two reasoning steps semantically entail each other, COTA defines the consistency between two reasoning chains as
\begin{equation}
\begin{aligned}
\mathrm{COTA}\!\left(R^{(a)},R^{(b)}\right)
&=
\frac{1}{L_a+L_b}
\Biggl(
\sum_{i=1}^{L_a}\max_{j} I_e\!\left(r^{(a)}_i,r^{(b)}_j\right) \\
&\qquad\qquad +
\sum_{j=1}^{L_b}\max_{i} I_e\!\left(r^{(a)}_i,r^{(b)}_j\right)
\Biggr),
\end{aligned}
\end{equation}
where $I_e(r,r')\in\{0,1\}$ indicates whether two reasoning steps semantically entail each other. The final CoT stability score is obtained by averaging the consistency between each sampled reasoning chain and the original one:
\begin{equation}
C_{\mathrm{COTA}}(x)
=
\frac{1}{M}\sum_{m=1}^{M}\mathrm{COTA}\!\left(R^{(m)},R\right).
\end{equation}
If the sampled reasoning chains remain semantically consistent with the original reasoning process, the explanation is considered more stable and the confidence is higher.

\textbf{T3}~\cite{li2024think} (Think twice before trusting) mainly targets multiple-choice settings. Given a question \(x\) and a set of candidate options \(\mathcal{A}=\{a_1,\dots,a_{N_A}\}\), T3 first asks the model to generate an explanation supporting each candidate option from the perspective that this candidate may be true, producing an explanation set. It then feeds all explanations into a prompt, and uses TopK verbalization so that the model outputs multiple answers and their probabilities in one response. In this way, the model can perform a comparative confidence evaluation over a more complete candidate option space, which helps reduce over-trust in a single wrong answer.

\textbf{Topo-UQ}~\cite{da2025understanding} formulates UE as a problem of comparing multiple sampled reasoning topologies. For the same question \(x\), the model first decomposes the reasoning process into a semantic directed graph \(\mathcal{G}=(\mathcal{V},\mathcal{E})\), where nodes represent intermediate conclusions or sub-answers, and edges represent dependency relations between reasoning steps. By sampling \(M\) times, we obtain a set of reasoning graphs \(\{\mathcal{G}^{(i)}\}_{i=1}^{M}\). Topo-UQ then measures the structural difference between two sampled reasoning graphs using a semantically enhanced graph edit distance:
\begin{equation}
\Delta_{ij}=\mathrm{GED}\!\left(\mathcal{G}^{(i)},\mathcal{G}^{(j)}\right),
\end{equation}
where the node substitution cost is defined based on cosine similarity. Based on all pairwise graph distances, the structural reasoning uncertainty is defined as
\begin{equation}
U_{\mathrm{Topo\text{-}UQ}}(x)=\operatorname{Var}\!\left(\{\Delta_{ij}\mid 1\le i<j\le M\}\right).
\end{equation}
If the sampled reasoning graphs exhibit more diverse topologies, then the pairwise distances become more dispersed, indicating higher uncertainty.

\textbf{IUQ}~\cite{mei2025reasoning} (Introspective UQ) asks the model to explicitly check and reflect on whether its own reasoning chain is reliable and whether it contains weaknesses, and then to reassess the confidence of the answer. This method designs three introspection prompts with different levels of conservativeness. IUQ-Low asks the model in a relatively neutral way to reread the previous reasoning chain and its original confidence. IUQ-Medium further asks the model to examine the original reasoning more critically and actively look for possible weaknesses, logical jumps, or insufficient evidence. IUQ-High is the most conservative, like IUQ-Medium, it asks the model to explicitly find flaws in the reasoning, but it does not provide the original confidence from the first stage, so that the model reviews the original reasoning more independently.

Zhang et al.~\cite{zhang2025all} propose a graph-based framework for UE that merges multiple sampled reasoning chains into a unified directed graph and then scores candidate answers according to different structural criteria. Let \(\mathcal{G}=(\mathcal{V},\mathcal{E})\) denote the graph constructed from multiple sampled reasoning chains, where \(Q\in\mathcal{V}\) is the question node, answer nodes correspond to candidate answers, and the remaining nodes represent intermediate reasoning steps. Within-chain edges encode sequential logical relations inside a single reasoning chain, while cross-chain edges connect semantically equivalent steps across different chains. Based on this graph, the paper introduces three variants, namely \textbf{CenConf}, \textbf{PathConv}, and \textbf{PathWeight}. For \textbf{CenConf}~\cite{zhang2025all}, the method uses the Katz centrality of answer nodes to measure their global structural influence in the graph. For a node \(v\), its Katz centrality is defined by
\begin{equation}
Katz(v)=\eta \sum_{u} A_{vu}\,Katz(u)+\gamma,
\end{equation}
where \(A\) is the adjacency matrix of \(\mathcal{G}\), \(\eta\) is a decay factor, and \(\gamma\) is a bias term. The confidence of candidate answer \(a_i\) is then obtained by normalizing the centrality values of all answer nodes:
\begin{equation}
C_{\mathrm{CenConf}}(x, a_i)=\frac{Katz(a_i)}{\sum_{a_j\in \mathcal{A}} Katz(a_j)}.
\end{equation} For \textbf{PathConv}~\cite{zhang2025all}, the underlying intuition is that an answer supported by more reasoning paths is more reliable. Its confidence is defined as
\begin{equation}
C_{\mathrm{PathConv}}(x, a_i)=
\frac{|\mathrm{Paths}(Q\rightarrow a_i)|}
{\sum_{a_j\in \mathcal{A}} |\mathrm{Paths}(Q\rightarrow a_j)|},
\end{equation}
where
\begin{equation}
\mathrm{Paths}(Q\rightarrow a_i)=\{\varrho \mid \varrho=(Q \rightarrow \cdots \rightarrow a_i)\in \mathcal{G}\}.
\end{equation} Building on this, \textbf{PathWeight}~\cite{zhang2025all} further considers the importance of merged nodes along each path. Specifically, nodes connected by cross-chain edges are merged into a single node, and the resulting node weight reflects how many original reasoning-step nodes are merged into it. Let \(\nu(u)\) denote the weight of node \(u\). If \(\varrho\) denotes a path from the question node \(Q\) to the candidate answer node \(a_i\), its path score is defined as
\begin{equation}
\mathrm{Score}(\varrho)=\prod_{u\in \varrho}\nu(u).
\end{equation}
The confidence of candidate answer \(a_i\) is then given by
\begin{equation}
C_{\mathrm{PathWeight}}(x,a_i)=
\frac{\sum_{\varrho\in \mathrm{Paths}(Q\rightarrow a_i)} \mathrm{Score}(\varrho)}
{\sum_{a_j\in \mathcal{A}} \sum_{\varrho'\in \mathrm{Paths}(Q\rightarrow a_j)} \mathrm{Score}(\varrho')}.
\end{equation}

Overall, explanation-based methods are useful because they move uncertainty estimation from final-answer comparison to reasoning-process analysis. This can reveal cases where the final answer appears plausible but the supporting reasoning is unstable or incomplete. However, these methods also inherit the fragility of generated explanations, since different reasoning chains may start from different assumptions or perspectives while still being valid.

\subsection{Multi-agent}
\label{Sec:agent}

Multi-agent methods estimate uncertainty by introducing multiple agents with different roles, perspectives, or capabilities and then using their interactions as an additional source of evidence. Unlike single-model methods, which rely on one model's own generations, these approaches exploit agreement, disagreement, revision, and debate across agents to assess answer reliability. The general intuition is that an answer that remains stable under multi-agent interaction is more likely to be trustworthy.

\textbf{Collab}~\cite{yang2024confidence} (CollabCalibration) first assigns different skills to a group of expert agents, and each expert agent provides a candidate answer together with its confidence score for the input $x$. The answers produced by these expert agents constitute the candidate answer space for the following evaluation stage, regardless of whether the task is open-ended or closed-ended. These candidate answers are then re-evaluated by several general agents. After deliberation, the answer with the highest frequency among the general agents is taken as the final prediction, and the mean confidence score associated with that answer is used as the final confidence. The core idea is that an answer endorsed by multiple perspectives and preserved after deliberation is more likely to be reliable, and should therefore receive a higher confidence score.

Although \textbf{ArgLLMs}~\cite{freedman2025argumentative} is not explicitly designed as a multi-agent method, it is based on the idea of introducing supporters and opponents to construct an argumentation tree around the predicted answer $y$, and then estimates its final confidence by propagating argumentative strengths through the tree. Specifically, the root node corresponds to the main answer $y$, its child nodes represent arguments that support or attack this answer, and deeper nodes represent further support or rebuttal for these arguments. For each node $v$ in the tree, the model first assigns a verbalized confidence $C_{\mathrm{VC}}(x,v)\in[0,1]$, which represents the initial plausibility of that argument itself. If $v$ is a leaf node, then its final strength is directly given by
\begin{equation}
\rho(v)=C_{\mathrm{VC}}(x,v).
\end{equation}
For a non-leaf node, let $R^{-}(v)$ and $R^{+}(v)$ denote the sets of child nodes that attack and support $v$, respectively. The method first aggregates the effects from the attacking side and the supporting side:
\begin{equation}
\begin{split}
att(v)=1-\prod_{u\in R^{-}(v)}\bigl(1-\rho(u)\bigr), \\
sup(v)=1-\prod_{u\in R^{+}(v)}\bigl(1-\rho(u)\bigr).
\end{split}
\end{equation}
Then it defines
\begin{equation}
\delta(v)=sup(v)-att(v).
\end{equation}
The final strength of node $v$ is obtained by the following piecewise linear update:
\begin{equation}
\rho(v)=
\begin{cases}
C_{\mathrm{VC}}(x,v)+\delta(v)\bigl(1-C_{\mathrm{VC}}(x,v)\bigr), & \delta(v)>0,\\[4pt]
C_{\mathrm{VC}}(x,v)+\delta(v)\,C_{\mathrm{VC}}(x,v), & \delta(v)\le 0.
\end{cases}
\end{equation}
Here, $\delta(v)>0$ means that the supporting effect is stronger than the attacking effect, so the node strength moves toward $1$; otherwise, the attacking side dominates, and the node strength moves toward $0$. Finally, the strength of the root node corresponding to the predicted answer $y$ is used as the confidence:
\begin{equation}
C_{\mathrm{ArgLLMs}}(x,y)=\rho(v_{\mathrm{root}}).
\end{equation}
Therefore, the essence of ArgLLMs is to first generate supporting and opposing arguments around an answer, then assign each argument an initial score, and finally obtain the confidence of the main answer through strength propagation over the tree structure.

\textbf{DAE}~\cite{feng2024rethinking} (DiverseAgentEntropy) first generates a set of conceptually diverse rewrites for the input question, denoted as $\mathcal{X}(x)=\{x_1,\dots,x_M\}$, and assigns each rewritten question $x_i$ to one agent. Each agent then produces an initial answer $y_i$ based on its assigned input. During interaction, each agent selects another agent whose current answer is different and with whom it has not interacted before, conducts a pairwise discussion, and updates its own answer accordingly. The method uses the number of answer revisions as a proxy for uncertainty. If an agent frequently changes its answer, its prediction is considered less stable. Let $q_i$ denote the number of times the $i$-th agent revises its answer during interaction, and let $w_i$ denote its reliability weight, defined as
\begin{equation}
w_i=\frac{T-q_i+1}{\sum_{j=1}^{M}(T-q_j+1)},
\end{equation}
where $T$ denotes the interaction rounds. Let $\mathcal{Z}=\{z_1,\dots,z_K\}$ denote the set of distinct final answers obtained after clustering. The induced predictive distribution is then defined as
\begin{equation}
\tilde{p}_k
=
\sum_{i=1}^{M}w_i\,\mathbf{1}\{y_i=z_k\}.
\end{equation}
Based on this distribution, the final uncertainty is measured by entropy:
\begin{equation}
U_{\mathrm{DAE}}(x)= -\sum_{k=1}^{K}\tilde{p}_k\log \tilde{p}_k.
\end{equation}

Overall, multi-agent methods enlarge the evidence space by introducing disagreement, debate, and revision among agents. This makes them appealing for difficult questions where a single model call may be unreliable. However, their effectiveness depends on whether the interaction genuinely surfaces complementary evidence. If agents share similar biases or if deliberation merely amplifies a majority opinion, multi-agent interaction may increase cost without improving uncertainty estimation.

\subsection{Hybrid}
\label{Sec:hybrid}

Hybrid methods are motivated by the observation that no single uncertainty signal is sufficient to capture model reliability robustly across settings. Different signals encode different aspects of uncertainty: consistency reflects output stability, verbalization reflects self-assessment, and candidate comparison or explanation reflects relative preference or reasoning support. Hybrid methods therefore combine multiple complementary signals in order to produce uncertainty estimates that are more robust than those obtained from any single source alone.

\textbf{BSDetector}~\cite{chen2024quantifying} first computes an external consistency score between the predicted answer $y$ and the sampled responses, denoted by $C_{\mathrm{SC}}(x,y)$. Besides this consistency signal, BSDetector also introduces the model's self-reflection certainty for answer $y$, denoted by $C_{Ref}(x,y)$. Specifically, the model is further asked about the correctness of $y$, and its categorical responses $\{\mathrm{Correct},\mathrm{Incorrect},\mathrm{Not\ sure}\}$ are mapped to numerical scores $\{1.0,0.0,0.5\}$. The final confidence is then defined as
\begin{equation}
C_{\mathrm{BSDetector}}(x,y)=\alpha\,C_{\mathrm{SC}}(x,y)+(1-\alpha)\,C_{Ref}(x,y),
\end{equation}
where $\alpha=0.7$ is a weighting parameter.

\textbf{UF}~\cite{zhang2024calibrating} (UF Calibration) mainly targets closed-ended QA and decomposes confidence into question-level uncertainty and answer fidelity. For an input $x$ and a candidate option $a_i$, the final confidence is defined as
\begin{equation}
C_{\mathrm{UF}}(x,a_i)=\bigl(1-U_Q(x)\bigr)\cdot F(a_i),
\end{equation}
where $U_Q(x)$ measures question-level uncertainty from the entropy of sampled responses, and $F(a_i)$ measures the fidelity of option $a_i$. To compute fidelity, UF constructs multiple faithful chains by iteratively testing whether the model returns to the original option after replacing the current option with "all other options are wrong". Let $m_i$ denote the position of option $a_i$ in a faithful chain $h_j$. The local fidelity is then defined as
\begin{equation}
\mathrm{Fidelity}_{h_j}(a_i)=\frac{m_i}{\sum_{i'=1}^{N_A}m_{i'}},
\end{equation}
where $h_j$ denotes the $j$-th faithful chain and $N_A$ is the number of candidate options in the chain. The overall fidelity is obtained by aggregating over all faithful chains:
\begin{equation}
F(a_i)=\sum_{j=1}^{N_H}p_{\mathrm{sampled}}(h_j)\cdot \mathrm{Fidelity}_{h_j}(a_i),
\end{equation}
where $N_H$ is the number of faithful chains and $p_{\mathrm{sampled}}(h_j)$ denotes the frequency of the starting option of chain $h_j$ among the sampled responses. In this way, UF separates "how uncertain the model is about the question" from "how strongly the model sticks to a candidate option", and then combines them into the final confidence.

\textbf{SteerConf}~\cite{zhou2025steerconf} 
estimates confidence by querying the model with multiple confidence-steering prompts, based on the idea that reliable confidence should remain stable under different prompting styles. Given an input $x$, it obtains answer--confidence pairs $\{(y_l,c_l)\}_{l=-L}^{L}$, where $y_l$ is the answer generated at steering level $l$ and $c_l\in[0,1]$ is the corresponding verbalized confidence. SteerConf then extracts three signals. The first is the mean confidence
\begin{equation}
\mu_c(x)=\frac{1}{2L+1}\sum_{l=-L}^{L}c_{l}.
\end{equation}
The second is the confidence consistency, defined through the standard deviation
\begin{equation}
\sigma_c(x)=\sqrt{\frac{1}{2L+1}\sum_{l=-L}^{L}\bigl(c_{l}(x)-\mu_c(x)\bigr)^2},
\end{equation}
and
\begin{equation}
\kappa_{\mathrm{conf}}(x)=\frac{1}{1+\sigma_c(x)/\mu_c(x)}.
\end{equation}
The third is the answer consistency $\kappa_{\mathrm{ans}}(x)$, which is defined as the frequency of the most common answer among all steering prompts. The final confidence is then defined as
\begin{equation}
C_{\mathrm{SteerConf}}(x)=\mu_c(x)\cdot \kappa_{\mathrm{conf}}(x)\cdot \kappa_{\mathrm{ans}}(x).
\end{equation}
SteerConf further selects the final answer by mapping the calibrated confidence to a discrete steering index and using the answer generated at that index.

\textbf{DiNCo}~\cite{wang2025calibrating} (Distractor-Normalized Coherence) treats verbalized confidence as a relative rather than absolute quantity. Even if an answer itself has a high confidence score, it may still be unreliable if its mutually exclusive alternatives also receive high scores. DiNCo first generates $M$ candidate answers with verbalized confidence scores, selects the highest-confidence answer as the prediction $y$, and treats the remaining candidates as distractors $D(x,y)$. DiNCo then normalizes the verbalized confidence of $y$ by the confidence assigned to its mutually exclusive distractors:
\begin{equation}
C_{\mathrm{NVC}}(x, y)=\frac{C_{\mathrm{VC}}(x, y)}{\beta({D}(x, y))},
\end{equation}
and the normalization factor is defined as
\begin{multline}
\beta({D}(x, y))=\max\!\Bigl(1,\ C_{\mathrm{VC}}(x, y)+\\
\sum_{y'\in {D}(x,y)} C_{\mathrm{VC}}(x, y')\cdot w_{\mathrm{unique}}(y')\cdot w_{\mathrm{contra}}(y')\Bigr).
\end{multline}
Here, $w_{\mathrm{unique}}(y')$ measures how replaceable distractor $y'$ is with respect to other distractors, while $w_{\mathrm{contra}}(y')$ measures how strongly distractor $y'$ conflicts with the predicted answer $y$. Both are computed by an NLI model:
\begin{equation}
w_{\mathrm{unique}}(y')=\frac{1}{\sum_{y''\in D(x, y)} p_{\mathrm{NLI}}(\mathrm{ent}\mid y', y'')},
\end{equation}
and
\begin{equation}
w_{\mathrm{contra}}(y')=\frac{p_{\mathrm{NLI}}(\mathrm{con}\mid y, y')+p_{\mathrm{NLI}}(\mathrm{con}\mid y', y)}{2}.
\end{equation}
After obtaining the normalized verbalized confidence, DiNCo combines it with the consistency score $C_{\mathrm{SC}}(x, y)$ and defines the final confidence as
\begin{equation}
C_{\mathrm{DiNCo}}(x, y)=\frac{1}{2}C_{\mathrm{SC}}(x, y)+\frac{1}{2}C_{\mathrm{NVC}}(x, y).
\end{equation}

Overall, hybrid methods combine complementary signals. This helps compensate for the weakness of any single signal. The limitation is that hybrid methods introduce additional design choices, including how to normalize heterogeneous signals and how to weight them. Without careful design, the combination may become task-specific or difficult to interpret.

\subsection{Others}
\label{Sec:other}

The methods discussed above mainly belong to the training-free black-box UE paradigm, where uncertainty is inferred directly from the target model's observable outputs. Beyond this setting, black-box UE can also be improved by introducing additional supervision or cross-model information. Under the supervised-learning paradigm, \textbf{APRICOT}~\cite{ulmer2024calibrating} trains an auxiliary calibrator to predict the target model's confidence using only the input question and generated text. It clusters question representations and uses the empirical correctness rate within each cluster as supervision, thereby learning a mapping from input patterns to correctness probabilities without accessing internal model information. Li et al.~\cite{li2024graph} instead construct supervision signals from the consistency structure among sampled answers by building a similarity graph and using clustering results as node features for a GNN.

Black-box UE can also exploit cross-model information. \textbf{MUSE}~\cite{kruse2025simple} estimates epistemic uncertainty from disagreement among multiple models' predictive distributions and further selects a lower-disagreement model subset for aggregation. Xue et al.~\cite{xue2025verify} argue that self-sampling consistency may be close to its upper bound and introduce cross-model consistency by comparing the target model's answer with that of a verifier model. Their results show that even a weaker verifier can provide useful additional information.

\section{Experiment}
\subsection{Setup}
To systematically evaluate the applicability and robustness of black-box UE methods across different task formats and model scales, we adopt a unified experimental setup covering datasets, model selection, prompting strategy, inference settings, and evaluation protocols, as described below. This unified design helps ensure a fair comparison across methods by minimizing the effect of differences unrelated to the UE mechanism itself.

\textbf{Datasets and task types.}
We select four widely used benchmarks, namely TriviaQA~\cite{joshi2017triviaqa}, HotpotQA~\cite{yang2018hotpotqa}, CoQA~\cite{reddy2019coqa}, and TruthfulQA~\cite{lin2022truthfulqa}, to cover several representative scenarios, including factual QA, multi-hop reasoning, conversational QA, and hallucination-prone evaluation. The first three datasets are treated as open-ended QA, while TruthfulQA is used in its multiple-choice format and is therefore treated as a closed-ended QA. For all datasets, we use the test split whenever it is available; otherwise, we use the validation split. For methods that require parameter tuning, such as uncertainty score monotonic transformation or hyperparameter selection, we randomly reserve 50 examples from the corresponding test or validation split for lightweight tuning, and then randomly sample 500 examples from the remaining data for evaluation. All sampling procedures are conducted with a fixed random seed to ensure reproducibility.

\textbf{Model settings.}
We evaluate both open-source and closed-source black-box models, including Qwen3-4B-Instruct-2507 (Qwen3-4B), Qwen3-30B-A3B-Instruct-2507 (Qwen3-30B)~\cite{qwen3technicalreport}, DeepSeek-V3.2~\cite{liu2025deepseek}, and GPT-5-mini~\cite{singh2025openai}. This model set is chosen to cover different parameter scales (small and large) as well as different reasoning capabilities, so as to examine the stability of UE methods under cross-model transfer.

\textbf{Auxiliary models.}  
We use \texttt{deberta-v3-large-\allowbreak mnli-\allowbreak fever-\allowbreak anli-\allowbreak ling-\allowbreak wanli} as the NLI model, mainly for computing entailment or contradiction relations and semantic consistency between answers. For methods that require semantic vector representations, we use \texttt{all-MiniLM-L6-v2} as the text embedding model to compute cosine similarity between answers.

\textbf{Prompts and output format.}
To keep answer accuracy as comparable as possible across methods and to attribute differences mainly to the UE mechanism, we adopt Chain-of-Thought (CoT) generation as the default setting for all methods. For verbalization-based methods, we use the prompt templates from the VPD~\cite{wang2025don} paper and require the model to explicitly output both the reasoning process and the final answer. For sampling-based methods, we use the same CoT format but remove the requirement of confidence output. For other methods, the prompts are generally kept consistent with those used in the original papers.

\textbf{Sampling and inference hyperparameters.}  
For settings that require repeated sampling, we set the temperature to $1.0$ and fix the number of samples to $5$. For settings without repeated sampling, we set the temperature to $0.1$ to approximate near-deterministic generation. The maximum generation length is set to $8192$ tokens, while all other decoding parameters follow the default configuration of each model.

\textbf{Correctness adjudication.}
For the closed-ended QA (TruthfulQA), we normalize the model output using string and regular-expression rules, extract the final predicted option, and directly compare it with the gold answer. For the other open-ended QA (TriviaQA, HotpotQA, and CoQA), we adopt an LLM-as-a-judge protocol. Specifically, we use GPT-5.1 as the judge model, provide it with the question, the reference answer, and the model-generated answer, and ask it to output a binary correctness label (correct/incorrect).

\textbf{Evaluation metrics.}
We evaluate confidence quality from two perspectives: discrimination and calibration. Specifically, we use AUROC~\cite{bradley1997use}, Expected Calibration Error (ECE) ~\cite{naeini2015obtaining}, and Brier Score~\cite{glenn1950verification}. AUROC treats answer correctness as a binary label and uses the predicted confidence \(C\) as the score, measuring how well the method ranks correct answers above incorrect ones. ECE groups samples into confidence bins, computes the difference between average confidence and empirical accuracy within each bin, and then takes the sum across bins; in this paper, the number of bins is fixed to 10. The Brier Score is defined as
\begin{equation}
\frac{1}{N}\sum_i(C(x^{(i)},y^{(i)})-\mathbf{1}\{y^{(i)}=\hat{y}^{(i)}\})^2,
\end{equation}
where \(N\) denotes the total number of evaluated samples. This metric reflects both calibration and sharpness, while penalizing high-confidence errors more heavily.

\subsection{Implementation Details}

We evaluate representative methods from each category of black-box UE. Since different methods vary substantially in their inference procedures and confidence computation, and some original papers provide multiple implementations or variants, we describe below the key implementation details and parameter settings adopted for each representative method in our experiment.

\textbf{Verbalization-based methods.}  
We evaluate four verbalization-based methods, namely CoT~\cite{xiong2306can}, TopK~\cite{tian2023just}, VPD~\cite{wang2025don}, and Ling~\cite{tian2023just}. For TopK, we set \(K=2\). For VPD, we ignore the "None of the above" option during evaluation and select the candidate with the highest confidence among the remaining options as the primary answer. For Ling, we follow UF~\cite{zhang2024calibrating} for the set of uncertainty expressions and the mapping from linguistic expressions to confidence scores. All of these verbalization-based methods can also be combined with sampling. In the sampling-aggregated setting, we average confidence scores at the semantic-cluster level and then select the answer from the cluster with the highest average confidence as the final answer for evaluation.

\textbf{Sampling-based methods.}  
We evaluate SE~\cite{farquhar2024detecting}, SelfCheck~\cite{manakul2023selfcheckgpt}, Deg~\cite{lin2023generating}, EigV~\cite{lin2023generating}, Ecc~\cite{lin2023generating}, KLE~\cite{nikitin2024kernel}, SEU~\cite{grewal2024improving}, SINdex~\cite{abdaljalil2025sindex}, SNNE~\cite{nguyen2025beyond}, SPUQ~\cite{gao2024spuq}, and InvE~\cite{song2025inv}. For SelfCheck, Deg, EigV, Ecc, and SNNE, alternative similarity measures such as Jaccard, BERTScore, ROUGE-L, and cosine similarity are also possible; in this work, we adopt the more commonly used NLI-based version introduced above. For the perturbation-based methods SPUQ and InvE, we construct 4 rewritten versions of the original question, which together with the original question form a set of 5 questions. For SPUQ, all perturbed inputs are generated using a consistent high temperature, and confidence is then estimated following the original cross-sample aggregation strategy.

\textbf{Explanation-based methods.}  
We evaluate T3~\cite{li2024think}, COTA~\cite{tanneru2024quantifying} and PathWeight~\cite{zhang2025all}. For COTA and PathWeight, directly generated reasoning steps are often too long and not convenient for later structured processing. Therefore, we use the CoT extractor from RACE~\cite{wang2025joint}. Specifically, we take the CoT generated by the sampling-based method as input, and use the extractor to produce structured reasoning steps for later computation.

\textbf{Multi-agent methods.}  
We evaluate Collab~\cite{yang2024confidence} and ArgLLMs~\cite{freedman2025argumentative}. For Collab, we use five role-specific agents and five general agents. For the role-specific agents, we follow the original code and use the CoT and Knowledge role types. For ArgLLMs, we use the prompt format recommended in the paper, namely the "OPRO" prompt for the argument miner and the "analyst" prompt for the uncertainty estimator, and set the reasoning tree depth to 2.

\textbf{Hybrid methods.}  
We evaluate BSDetector~\cite{chen2024quantifying}, UF~\cite{zhang2024calibrating}, SteerConf~\cite{zhou2025steerconf}, and DiNCo~\cite{wang2025calibrating}. For the sampling-related parts in BSDetector and DiNCo, we directly reuse the sampled outputs obtained in the sampling-based methods, so that the comparison remains consistent across different method categories. For DiNCo, in order to obtain multiple distinct candidate answers more directly, we simplify the original procedure by adopting a TopK-like prompt format and setting \(K=5\) in our implementation.
\section{Evaluation Results}

This section presents the main evaluation results of black-box UE methods. We organize the section around two evaluation settings: open-ended QA and closed-ended QA. For open-ended QA, we first discuss verbalization-based, sampling-based, and other black-box UE methods separately, and then provide an overall comparison across all evaluated methods. We further analyze performance differences across datasets, model families, and sampling budgets. For closed-ended QA, the candidate answers are predefined, so we evaluate the subset of black-box UE methods that can estimate uncertainty over option-based predictions. This organization allows us to examine the overall effectiveness of different UE methods and to analyze key factors that influence their performance, including task format, dataset characteristics, model family.

\begin{table}[!hbtp]
\centering
\caption{ Results of verbalization-based methods on open-ended QA. We compare four methods under two settings, namely single-pass and sampling-aggregated. Avg. denotes the macro-average over all dataset--model pairs shown in the table. For visualization, darker row-wise shading indicates better performance among the compared methods, with higher values preferred for Acc/AUROC and lower values preferred for ECE/Brier.}

\label{tab:verbal-open}
\resizebox{0.49\textwidth}{!}{
\begin{tabular}{@{}>{\centering\arraybackslash}p{0.8cm}
                >{\centering\arraybackslash}p{1.0cm}
                >{\centering\arraybackslash}p{1.8cm}
                *{8}{>{\centering\arraybackslash}p{1.0cm}}@{}}
                
\toprule
\multirow{2}{*}{\textbf{Metric}} &
\multirow{2}{*}{\textbf{Dataset}} &
\multirow{2}{*}{\textbf{Model}} &
\multicolumn{4}{c}{\textbf{Single-pass}} &
\multicolumn{4}{c}{\textbf{Sampling-aggregated}} \\
\cmidrule(r){4-7}\cmidrule(l){8-11}
& & &
\textbf{CoT} & \textbf{TopK} & \textbf{VPD} & \textbf{Ling} &
\textbf{CoT} & \textbf{TopK} & \textbf{VPD} & \textbf{Ling} \\
\midrule

\multirow{16}{*}{\textbf{Acc}} &
\multirow{4}{=}{\parbox[c]{1.2cm}{\centering\textbf{TriviaQA}}} &
Qwen3-4B
& \heatrowhigh{48.45}{48.45}{53.00} & \heatrowhigh{50.62}{48.45}{53.00} & \heatrowhigh{52.14}{48.45}{53.00} & \heatrowhigh{50.20}{48.45}{53.00}
& \heatrowhigh{48.68}{48.45}{53.00} & \heatrowhigh{52.40}{48.45}{53.00} & \heatrowhigh{53.00}{48.45}{53.00} & \heatrowhigh{49.80}{48.45}{53.00} \\
& & Qwen3-30B
& \heatrowhigh{72.03}{72.03}{74.80} & \heatrowhigh{73.49}{72.03}{74.80} & \heatrowhigh{72.14}{72.03}{74.80} & \heatrowhigh{73.43}{72.03}{74.80}
& \heatrowhigh{73.60}{72.03}{74.80} & \heatrowhigh{73.55}{72.03}{74.80} & \heatrowhigh{73.80}{72.03}{74.80} & \heatrowhigh{74.80}{72.03}{74.80} \\
& & DeepSeek-V3.2
& \heatrowhigh{89.00}{85.60}{89.60} & \heatrowhigh{87.60}{85.60}{89.60} & \heatrowhigh{88.20}{85.60}{89.60} & \heatrowhigh{85.60}{85.60}{89.60}
& \heatrowhigh{89.60}{85.60}{89.60} & \heatrowhigh{88.20}{85.60}{89.60} & \heatrowhigh{88.60}{85.60}{89.60} & \heatrowhigh{87.20}{85.60}{89.60} \\
& & GPT-5-mini
& \heatrowhigh{88.80}{86.80}{89.00} & \heatrowhigh{87.60}{86.80}{89.00} & \heatrowhigh{88.60}{86.80}{89.00} & \heatrowhigh{86.80}{86.80}{89.00}
& \heatrowhigh{88.20}{86.80}{89.00} & \heatrowhigh{88.40}{86.80}{89.00} & \heatrowhigh{88.20}{86.80}{89.00} & \heatrowhigh{89.00}{86.80}{89.00} \\
\cmidrule{2-11}
& \multirow{4}{=}{\parbox[c]{1.2cm}{\centering\textbf{HotpotQA}}} &
Qwen3-4B
& \heatrowhigh{26.47}{24.40}{27.67} & \heatrowhigh{26.07}{24.40}{27.67} & \heatrowhigh{25.00}{24.40}{27.67} & \heatrowhigh{26.21}{24.40}{27.67}
& \heatrowhigh{27.67}{24.40}{27.67} & \heatrowhigh{26.00}{24.40}{27.67} & \heatrowhigh{24.40}{24.40}{27.67} & \heatrowhigh{27.40}{24.40}{27.67} \\
& & Qwen3-30B
& \heatrowhigh{35.08}{32.99}{37.88} & \heatrowhigh{36.82}{32.99}{37.88} & \heatrowhigh{32.99}{32.99}{37.88} & \heatrowhigh{37.53}{32.99}{37.88}
& \heatrowhigh{37.00}{32.99}{37.88} & \heatrowhigh{37.20}{32.99}{37.88} & \heatrowhigh{36.55}{32.99}{37.88} & \heatrowhigh{37.88}{32.99}{37.88} \\
& & DeepSeek-V3.2
& \heatrowhigh{56.11}{53.11}{59.20} & \heatrowhigh{53.52}{53.11}{59.20} & \heatrowhigh{54.34}{53.11}{59.20} & \heatrowhigh{53.11}{53.11}{59.20}
& \heatrowhigh{59.20}{53.11}{59.20} & \heatrowhigh{56.40}{53.11}{59.20} & \heatrowhigh{57.40}{53.11}{59.20} & \heatrowhigh{55.40}{53.11}{59.20} \\
& & GPT-5-mini
& \heatrowhigh{62.40}{61.40}{65.00} & \heatrowhigh{61.40}{61.40}{65.00} & \heatrowhigh{62.05}{61.40}{65.00} & \heatrowhigh{64.00}{61.40}{65.00}
& \heatrowhigh{63.00}{61.40}{65.00} & \heatrowhigh{64.40}{61.40}{65.00} & \heatrowhigh{65.00}{61.40}{65.00} & \heatrowhigh{63.60}{61.40}{65.00} \\
\cmidrule{2-11}
& \multirow{4}{=}{\parbox[c]{1.2cm}{\centering\textbf{CoQA}}} &
Qwen3-4B
& \heatrowhigh{82.20}{82.00}{84.60} & \heatrowhigh{83.60}{82.00}{84.60} & \heatrowhigh{82.00}{82.00}{84.60} & \heatrowhigh{83.60}{82.00}{84.60}
& \heatrowhigh{83.40}{82.00}{84.60} & \heatrowhigh{84.60}{82.00}{84.60} & \heatrowhigh{83.80}{82.00}{84.60} & \heatrowhigh{84.40}{82.00}{84.60} \\
& & Qwen3-30B
& \heatrowhigh{88.20}{86.20}{88.80} & \heatrowhigh{86.80}{86.20}{88.80} & \heatrowhigh{88.38}{86.20}{88.80} & \heatrowhigh{88.00}{86.20}{88.80}
& \heatrowhigh{88.80}{86.20}{88.80} & \heatrowhigh{86.20}{86.20}{88.80} & \heatrowhigh{88.20}{86.20}{88.80} & \heatrowhigh{87.20}{86.20}{88.80} \\
& & DeepSeek-V3.2
& \heatrowhigh{92.38}{83.17}{92.40} & \heatrowhigh{90.40}{83.17}{92.40} & \heatrowhigh{90.00}{83.17}{92.40} & \heatrowhigh{83.17}{83.17}{92.40}
& \heatrowhigh{92.20}{83.17}{92.40} & \heatrowhigh{91.00}{83.17}{92.40} & \heatrowhigh{92.40}{83.17}{92.40} & \heatrowhigh{87.40}{83.17}{92.40} \\
& & GPT-5-mini
& \heatrowhigh{95.60}{93.00}{95.60} & \heatrowhigh{93.00}{93.00}{95.60} & \heatrowhigh{94.40}{93.00}{95.60} & \heatrowhigh{94.20}{93.00}{95.60}
& \heatrowhigh{94.60}{93.00}{95.60} & \heatrowhigh{93.20}{93.00}{95.60} & \heatrowhigh{94.40}{93.00}{95.60} & \heatrowhigh{93.60}{93.00}{95.60} \\
\cmidrule{2-11}
& \multicolumn{2}{c}{\textbf{Avg.}} &
\heatrowhigh{69.73}{68.82}{70.50} & \heatrowhigh{69.24}{68.82}{70.50} & \heatrowhigh{69.19}{68.82}{70.50} & \heatrowhigh{68.82}{68.82}{70.50} &
\heatrowhigh{70.50}{68.82}{70.50} & \heatrowhigh{70.13}{68.82}{70.50} & \heatrowhigh{70.48}{68.82}{70.50} & \heatrowhigh{69.81}{68.82}{70.50} \\
\midrule

\multirow{16}{*}{\textbf{AUROC}} &
\multirow{4}{=}{\parbox[c]{1.2cm}{\centering\textbf{TriviaQA}}} &
Qwen3-4B
& \heatrowhigh{73.70}{69.30}{84.26} & \heatrowhigh{74.10}{69.30}{84.26} & \heatrowhigh{74.74}{69.30}{84.26} & \heatrowhigh{69.30}{69.30}{84.26}
& \heatrowhigh{84.25}{69.30}{84.26} & \heatrowhigh{84.26}{69.30}{84.26} & \heatrowhigh{82.97}{69.30}{84.26} & \heatrowhigh{84.20}{69.30}{84.26} \\
& & Qwen3-30B
& \heatrowhigh{69.17}{68.27}{84.65} & \heatrowhigh{68.82}{68.27}{84.65} & \heatrowhigh{70.09}{68.27}{84.65} & \heatrowhigh{68.27}{68.27}{84.65}
& \heatrowhigh{84.65}{68.27}{84.65} & \heatrowhigh{84.42}{68.27}{84.65} & \heatrowhigh{82.72}{68.27}{84.65} & \heatrowhigh{80.91}{68.27}{84.65} \\
& & DeepSeek-V3.2
& \heatrowhigh{73.50}{68.15}{83.94} & \heatrowhigh{77.89}{68.15}{83.94} & \heatrowhigh{80.20}{68.15}{83.94} & \heatrowhigh{68.15}{68.15}{83.94}
& \heatrowhigh{80.16}{68.15}{83.94} & \heatrowhigh{78.57}{68.15}{83.94} & \heatrowhigh{83.94}{68.15}{83.94} & \heatrowhigh{73.86}{68.15}{83.94} \\
& & GPT-5-mini
& \heatrowhigh{78.67}{77.25}{86.30} & \heatrowhigh{79.78}{77.25}{86.30} & \heatrowhigh{83.33}{77.25}{86.30} & \heatrowhigh{77.25}{77.25}{86.30}
& \heatrowhigh{78.87}{77.25}{86.30} & \heatrowhigh{84.49}{77.25}{86.30} & \heatrowhigh{86.30}{77.25}{86.30} & \heatrowhigh{82.54}{77.25}{86.30} \\
\cmidrule{2-11}
& \multirow{4}{=}{\parbox[c]{1.2cm}{\centering\textbf{HotpotQA}}} &
Qwen3-4B
& \heatrowhigh{73.23}{72.43}{84.57} & \heatrowhigh{73.65}{72.43}{84.57} & \heatrowhigh{79.71}{72.43}{84.57} & \heatrowhigh{72.43}{72.43}{84.57}
& \heatrowhigh{81.33}{72.43}{84.57} & \heatrowhigh{83.39}{72.43}{84.57} & \heatrowhigh{84.02}{72.43}{84.57} & \heatrowhigh{84.57}{72.43}{84.57} \\
& & Qwen3-30B
& \heatrowhigh{70.52}{70.52}{86.09} & \heatrowhigh{70.69}{70.52}{86.09} & \heatrowhigh{76.29}{70.52}{86.09} & \heatrowhigh{71.66}{70.52}{86.09}
& \heatrowhigh{84.32}{70.52}{86.09} & \heatrowhigh{86.09}{70.52}{86.09} & \heatrowhigh{83.76}{70.52}{86.09} & \heatrowhigh{82.02}{70.52}{86.09} \\
& & DeepSeek-V3.2
& \heatrowhigh{77.64}{75.53}{86.13} & \heatrowhigh{79.91}{75.53}{86.13} & \heatrowhigh{78.01}{75.53}{86.13} & \heatrowhigh{75.53}{75.53}{86.13}
& \heatrowhigh{84.7}{75.53}{86.13} & \heatrowhigh{86.13}{75.53}{86.13} & \heatrowhigh{85.01}{75.53}{86.13} & \heatrowhigh{81.06}{75.53}{86.13} \\
& & GPT-5-mini
& \heatrowhigh{83.06}{79.47}{85.59} & \heatrowhigh{83.18}{79.47}{85.59} & \heatrowhigh{81.57}{79.47}{85.59} & \heatrowhigh{79.47}{79.47}{85.59}
& \heatrowhigh{85.59}{79.47}{85.59} & \heatrowhigh{82.58}{79.47}{85.59} & \heatrowhigh{83.80}{79.47}{85.59} & \heatrowhigh{81.26}{79.47}{85.59} \\
\cmidrule{2-11}
& \multirow{4}{=}{\parbox[c]{1.2cm}{\centering\textbf{CoQA}}} &
Qwen3-4B
& \heatrowhigh{71.37}{71.37}{80.94} & \heatrowhigh{75.20}{71.37}{80.94} & \heatrowhigh{75.10}{71.37}{80.94} & \heatrowhigh{73.00}{71.37}{80.94}
& \heatrowhigh{77.86}{71.37}{80.94} & \heatrowhigh{75.61}{71.37}{80.94} & \heatrowhigh{80.94}{71.37}{80.94} & \heatrowhigh{71.89}{71.37}{80.94} \\
& & Qwen3-30B
& \heatrowhigh{59.94}{59.94}{82.14} & \heatrowhigh{67.04}{59.94}{82.14} & \heatrowhigh{78.50}{59.94}{82.14} & \heatrowhigh{66.56}{59.94}{82.14}
& \heatrowhigh{73.09}{59.94}{82.14} & \heatrowhigh{74.41}{59.94}{82.14} & \heatrowhigh{82.14}{59.94}{82.14} & \heatrowhigh{75.18}{59.94}{82.14} \\
& & DeepSeek-V3.2
& \heatrowhigh{66.80}{58.41}{82.15} & \heatrowhigh{69.20}{58.41}{82.15} & \heatrowhigh{67.91}{58.41}{82.15} & \heatrowhigh{58.41}{58.41}{82.15}
& \heatrowhigh{81.72}{58.41}{82.15} & \heatrowhigh{81.54}{58.41}{82.15} & \heatrowhigh{82.15}{58.41}{82.15} & \heatrowhigh{76.32}{58.41}{82.15} \\
& & GPT-5-mini
& \heatrowhigh{68.17}{58.07}{82.29} & \heatrowhigh{76.07}{58.07}{82.29} & \heatrowhigh{80.25}{58.07}{82.29} & \heatrowhigh{58.07}{58.07}{82.29}
& \heatrowhigh{72.74}{58.07}{82.29} & \heatrowhigh{82.29}{58.07}{82.29} & \heatrowhigh{72.65}{58.07}{82.29} & \heatrowhigh{74.38}{58.07}{82.29} \\
\cmidrule{2-11}
& \multicolumn{2}{c}{\textbf{Avg.}} &
\heatrowhigh{72.15}{69.84}{82.53} & \heatrowhigh{74.63}{69.84}{82.53} & \heatrowhigh{77.14}{69.84}{82.53} & \heatrowhigh{69.84}{69.84}{82.53} &
\heatrowhigh{80.77}{69.84}{82.53} & \heatrowhigh{81.98}{69.84}{82.53} & \heatrowhigh{82.53}{69.84}{82.53} & \heatrowhigh{79.02}{69.84}{82.53} \\
\midrule

\multirow{16}{*}{\textbf{ECE}} &
\multirow{4}{=}{\parbox[c]{1.2cm}{\centering\textbf{TriviaQA}}} &
Qwen3-4B
& \heatrowlow{48.01}{8.83}{48.01} & \heatrowlow{40.07}{8.83}{48.01} & \heatrowlow{27.19}{8.83}{48.01} & \heatrowlow{45.27}{8.83}{48.01}
& \heatrowlow{22.19}{8.83}{48.01} & \heatrowlow{17.24}{8.83}{48.01} & \heatrowlow{8.83}{8.83}{48.01} & \heatrowlow{26.13}{8.83}{48.01} \\
& & Qwen3-30B
& \heatrowlow{24.77}{3.87}{24.77} & \heatrowlow{19.23}{3.87}{24.77} & \heatrowlow{15.18}{3.87}{24.77} & \heatrowlow{23.51}{3.87}{24.77}
& \heatrowlow{8.69}{3.87}{24.77} & \heatrowlow{6.95}{3.87}{24.77} & \heatrowlow{3.87}{3.87}{24.77} & \heatrowlow{10.29}{3.87}{24.77} \\
& & DeepSeek-V3.2
& \heatrowlow{6.67}{3.28}{9.45} & \heatrowlow{4.99}{3.28}{9.45} & \heatrowlow{3.28}{3.28}{9.45} & \heatrowlow{9.45}{3.28}{9.45}
& \heatrowlow{3.53}{3.28}{9.45} & \heatrowlow{3.91}{3.28}{9.45} & \heatrowlow{4.29}{3.28}{9.45} & \heatrowlow{6.32}{3.28}{9.45} \\
& & GPT-5-mini
& \heatrowlow{5.06}{2.94}{9.72} & \heatrowlow{2.94}{2.94}{9.72} & \heatrowlow{5.11}{2.94}{9.72} & \heatrowlow{9.72}{2.94}{9.72}
& \heatrowlow{6.56}{2.94}{9.72} & \heatrowlow{2.95}{2.94}{9.72} & \heatrowlow{5.86}{2.94}{9.72} & \heatrowlow{5.35}{2.94}{9.72} \\
\cmidrule{2-11}
& \multirow{4}{=}{\parbox[c]{1.2cm}{\centering\textbf{HotpotQA}}} &
Qwen3-4B
& \heatrowlow{64.33}{7.86}{64.41} & \heatrowlow{58.99}{7.86}{64.41} & \heatrowlow{19.77}{7.86}{64.41} & \heatrowlow{64.41}{7.86}{64.41}
& \heatrowlow{31.93}{7.86}{64.41} & \heatrowlow{30.01}{7.86}{64.41} & \heatrowlow{7.86}{7.86}{64.41} & \heatrowlow{36.66}{7.86}{64.41} \\
& & Qwen3-30B
& \heatrowlow{58.18}{13.11}{58.18} & \heatrowlow{52.77}{13.11}{58.18} & \heatrowlow{37.08}{13.11}{58.18} & \heatrowlow{53.61}{13.11}{58.18}
& \heatrowlow{26.69}{13.11}{58.18} & \heatrowlow{26.60}{13.11}{58.18} & \heatrowlow{13.11}{13.11}{58.18} & \heatrowlow{29.83}{13.11}{58.18} \\
& & DeepSeek-V3.2
& \heatrowlow{31.75}{5.66}{32.48} & \heatrowlow{30.58}{5.66}{32.48} & \heatrowlow{17.72}{5.66}{32.48} & \heatrowlow{32.48}{5.66}{32.48}
& \heatrowlow{8.57}{5.66}{32.48} & \heatrowlow{11.34}{5.66}{32.48} & \heatrowlow{5.66}{5.66}{32.48} & \heatrowlow{14.07}{5.66}{32.48} \\
& & GPT-5-mini
& \heatrowlow{11.79}{2.90}{23.28} & \heatrowlow{19.67}{2.90}{23.28} & \heatrowlow{11.19}{2.90}{23.28} & \heatrowlow{23.28}{2.90}{23.28}
& \heatrowlow{2.90}{2.90}{23.28} & \heatrowlow{6.00}{2.90}{23.28} & \heatrowlow{3.69}{2.90}{23.28} & \heatrowlow{10.85}{2.90}{23.28} \\
\cmidrule{2-11}
& \multirow{4}{=}{\parbox[c]{1.2cm}{\centering\textbf{CoQA}}} &
Qwen3-4B
& \heatrowlow{15.54}{6.40}{15.54} & \heatrowlow{10.75}{6.40}{15.54} & \heatrowlow{11.60}{6.40}{15.54} & \heatrowlow{13.98}{6.40}{15.54}
& \heatrowlow{6.40}{6.40}{15.54} & \heatrowlow{7.81}{6.40}{15.54} & \heatrowlow{8.49}{6.40}{15.54} & \heatrowlow{11.27}{6.40}{15.54} \\
& & Qwen3-30B
& \heatrowlow{7.86}{4.42}{8.89} & \heatrowlow{6.86}{4.42}{8.89} & \heatrowlow{5.20}{4.42}{8.89} & \heatrowlow{8.62}{4.42}{8.89}
& \heatrowlow{6.60}{4.42}{8.89} & \heatrowlow{4.42}{4.42}{8.89} & \heatrowlow{8.89}{4.42}{8.89} & \heatrowlow{6.97}{4.42}{8.89} \\
& & DeepSeek-V3.2
& \heatrowlow{6.49}{3.21}{14.91} & \heatrowlow{3.21}{3.21}{14.91} & \heatrowlow{6.07}{3.21}{14.91} & \heatrowlow{14.91}{3.21}{14.91}
& \heatrowlow{10.59}{3.21}{14.91} & \heatrowlow{8.76}{3.21}{14.91} & \heatrowlow{14.21}{3.21}{14.91} & \heatrowlow{7.55}{3.21}{14.91} \\
& & GPT-5-mini
& \heatrowlow{1.04}{1.04}{18.84} & \heatrowlow{2.44}{1.04}{18.84} & \heatrowlow{10.33}{1.04}{18.84} & \heatrowlow{5.36}{1.04}{18.84}
& \heatrowlow{10.82}{1.04}{18.84} & \heatrowlow{16.86}{1.04}{18.84} & \heatrowlow{18.84}{1.04}{18.84} & \heatrowlow{9.88}{1.04}{18.84} \\
\cmidrule{2-11}
& \multicolumn{2}{c}{\textbf{Avg.}} &
\heatrowlow{23.46}{8.63}{25.38} & \heatrowlow{21.04}{8.63}{25.38} & \heatrowlow{14.14}{8.63}{25.38} & \heatrowlow{25.38}{8.63}{25.38} &
\heatrowlow{12.12}{8.63}{25.38} & \heatrowlow{11.90}{8.63}{25.38} & \heatrowlow{8.63}{8.63}{25.38} & \heatrowlow{14.60}{8.63}{25.38} \\
\midrule

\multirow{16}{*}{\textbf{Brier}} &
\multirow{4}{=}{\parbox[c]{1.2cm}{\centering\textbf{TriviaQA}}} &
Qwen3-4B
& \heatrowlow{47.13}{17.41}{47.13} & \heatrowlow{38.33}{17.41}{47.13} & \heatrowlow{29.26}{17.41}{47.13} & \heatrowlow{43.37}{17.41}{47.13}
& \heatrowlow{22.17}{17.41}{47.13} & \heatrowlow{19.30}{17.41}{47.13} & \heatrowlow{17.41}{17.41}{47.13} & \heatrowlow{24.02}{17.41}{47.13} \\
& & Qwen3-30B
& \heatrowlow{25.47}{12.77}{25.47} & \heatrowlow{21.75}{12.77}{25.47} & \heatrowlow{20.27}{12.77}{25.47} & \heatrowlow{23.57}{12.77}{25.47}
& \heatrowlow{13.62}{12.77}{25.47} & \heatrowlow{12.77}{12.77}{25.47} & \heatrowlow{13.58}{12.77}{25.47} & \heatrowlow{13.85}{12.77}{25.47} \\
& & DeepSeek-V3.2
& \heatrowlow{9.48}{7.85}{12.01} & \heatrowlow{9.43}{7.85}{12.01} & \heatrowlow{9.14}{7.85}{12.01} & \heatrowlow{12.01}{7.85}{12.01}
& \heatrowlow{7.85}{7.85}{12.01} & \heatrowlow{8.89}{7.85}{12.01} & \heatrowlow{8.39}{7.85}{12.01} & \heatrowlow{9.68}{7.85}{12.01} \\
& & GPT-5-mini
& \heatrowlow{8.13}{7.83}{10.36} & \heatrowlow{9.03}{7.83}{10.36} & \heatrowlow{7.83}{7.83}{10.36} & \heatrowlow{10.36}{7.83}{10.36}
& \heatrowlow{9.17}{7.83}{10.36} & \heatrowlow{7.88}{7.83}{10.36} & \heatrowlow{7.90}{7.83}{10.36} & \heatrowlow{8.00}{7.83}{10.36} \\
\cmidrule{2-11}
& \multirow{4}{=}{\parbox[c]{1.2cm}{\centering\textbf{HotpotQA}}} &
Qwen3-4B
& \heatrowlow{61.03}{13.72}{61.03} & \heatrowlow{52.78}{13.72}{61.03} & \heatrowlow{23.04}{13.72}{61.03} & \heatrowlow{60.03}{13.72}{61.03}
& \heatrowlow{26.58}{13.72}{61.03} & \heatrowlow{23.74}{13.72}{61.03} & \heatrowlow{13.72}{13.72}{61.03} & \heatrowlow{28.12}{13.72}{61.03} \\
& & Qwen3-30B
& \heatrowlow{55.98}{17.57}{55.98} & \heatrowlow{49.52}{17.57}{55.98} & \heatrowlow{34.96}{17.57}{55.98} & \heatrowlow{50.96}{17.57}{55.98}
& \heatrowlow{23.57}{17.57}{55.98} & \heatrowlow{22.44}{17.57}{55.98} & \heatrowlow{17.57}{17.57}{55.98} & \heatrowlow{25.91}{17.57}{55.98} \\
& & DeepSeek-V3.2
& \heatrowlow{31.35}{15.63}{31.59} & \heatrowlow{28.97}{15.63}{31.59} & \heatrowlow{22.22}{15.63}{31.59} & \heatrowlow{31.59}{15.63}{31.59}
& \heatrowlow{16.02}{15.63}{31.59} & \heatrowlow{16.32}{15.63}{31.59} & \heatrowlow{15.63}{15.63}{31.59} & \heatrowlow{19.55}{15.63}{31.59} \\
& & GPT-5-mini
& \heatrowlow{17.27}{14.60}{23.71} & \heatrowlow{21.88}{14.60}{23.71} & \heatrowlow{18.60}{14.60}{23.71} & \heatrowlow{23.71}{14.60}{23.71}
& \heatrowlow{14.60}{14.60}{23.71} & \heatrowlow{16.43}{14.60}{23.71} & \heatrowlow{15.41}{14.60}{23.71} & \heatrowlow{18.00}{14.60}{23.71} \\
\cmidrule{2-11}
& \multirow{4}{=}{\parbox[c]{1.2cm}{\centering\textbf{CoQA}}} &
Qwen3-4B
& \heatrowlow{16.31}{11.76}{16.31} & \heatrowlow{13.67}{11.76}{16.31} & \heatrowlow{13.08}{11.76}{16.31} & \heatrowlow{14.49}{11.76}{16.31}
& \heatrowlow{11.87}{11.76}{16.31} & \heatrowlow{12.02}{11.76}{16.31} & \heatrowlow{11.76}{11.76}{16.31} & \heatrowlow{13.13}{11.76}{16.31} \\
& & Qwen3-30B
& \heatrowlow{10.89}{9.09}{11.40} & \heatrowlow{11.40}{9.09}{11.40} & \heatrowlow{9.09}{9.09}{11.40} & \heatrowlow{10.73}{9.09}{11.40}
& \heatrowlow{9.47}{9.09}{11.40} & \heatrowlow{10.39}{9.09}{11.40} & \heatrowlow{9.78}{9.09}{11.40} & \heatrowlow{10.17}{9.09}{11.40} \\
& & DeepSeek-V3.2
& \heatrowlow{7.02}{7.02}{15.76} & \heatrowlow{8.28}{7.02}{15.76} & \heatrowlow{9.71}{7.02}{15.76} & \heatrowlow{15.76}{7.02}{15.76}
& \heatrowlow{8.91}{7.02}{15.76} & \heatrowlow{9.5}{7.02}{15.76} & \heatrowlow{9.92}{7.02}{15.76} & \heatrowlow{10.47}{7.02}{15.76} \\
& & GPT-5-mini
& \heatrowlow{4.13}{4.13}{11.09} & \heatrowlow{5.87}{4.13}{11.09} & \heatrowlow{6.18}{4.13}{11.09} & \heatrowlow{5.68}{4.13}{11.09}
& \heatrowlow{8.56}{4.13}{11.09} & \heatrowlow{10.46}{4.13}{11.09} & \heatrowlow{11.09}{4.13}{11.09} & \heatrowlow{8.65}{4.13}{11.09} \\
\cmidrule{2-11}
& \multicolumn{2}{c}{\textbf{Avg.}} &
\heatrowlow{24.52}{12.68}{25.19} & \heatrowlow{22.58}{12.68}{25.19} & \heatrowlow{16.95}{12.68}{25.19} & \heatrowlow{25.19}{12.68}{25.19} &
\heatrowlow{14.37}{12.68}{25.19} & \heatrowlow{14.18}{12.68}{25.19} & \heatrowlow{12.68}{12.68}{25.19} & \heatrowlow{15.80}{12.68}{25.19} \\
\bottomrule
\end{tabular}}
\end{table}

\subsection{Results on Open-ended QA}
\subsubsection{Verbalization-based}

Table~\ref{tab:verbal-open} reports the results of verbalization-based methods on open-ended QA. A first clear pattern is that \textbf{sampling aggregation consistently improves over single-pass verbalization across most settings}. This indicates that repeated elicitation provides a more stable uncertainty signal than relying on a single self-reported confidence score. It also narrows the performance gap among CoT~\cite{xiong2306can}, TopK~\cite{tian2023just}, and VPD~\cite{wang2025don} to some extent.

Among the four methods, Ling~\cite{tian2023just} performs relatively weakly overall. This suggests that natural-language expressions of uncertainty are difficult to map reliably into quantitative confidence scores. Although the model may use hedging expressions or epistemic markers to convey uncertainty, these signals do not yield well-calibrated numeric confidence as reliably as explicitly verbalized confidence scores do.

\begin{figure*}[!htbp]
    \centering
    \includegraphics[width=0.95\linewidth]{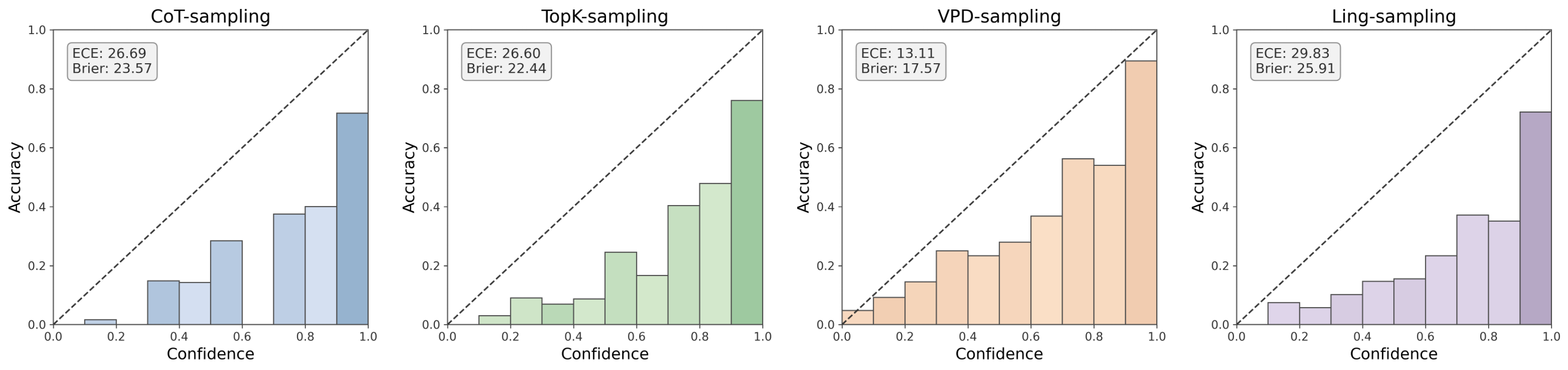}
    \caption{Reliability diagrams of Qwen3-30B on HotpotQA. In these diagrams, the darker the color, the higher the density.}
    \label{fig:verb-sample-ece}
\end{figure*}

By contrast, \textbf{VPD~\cite{wang2025don} shows consistently strong overall performance.} This may be because the method encourages the model to reason over multiple candidate answers within a single generation, thereby inducing implicit comparison across alternatives rather than confidence assignment to only one response. In addition, the \emph{"None of the above"} option provides an explicit mechanism for expressing residual uncertainty outside the candidate set. This is especially helpful on more difficult datasets such as HotpotQA, where it may help mitigate overconfidence, a pattern that is also reflected in Figure~\ref{fig:verb-sample-ece}.

In terms of answer accuracy, CoT~\cite{xiong2306can} achieves the best average performance. However, the accuracy gap among these verbalization-based methods is generally small across most settings, and no method shows a consistently overwhelming advantage in answer generation itself. This suggests that \textbf{the main differences lie not in answer generation itself, but rather in how effectively these methods expose usable uncertainty signals.}

\begin{figure}[!htbp]
    \centering
    \includegraphics[width=0.95\linewidth]{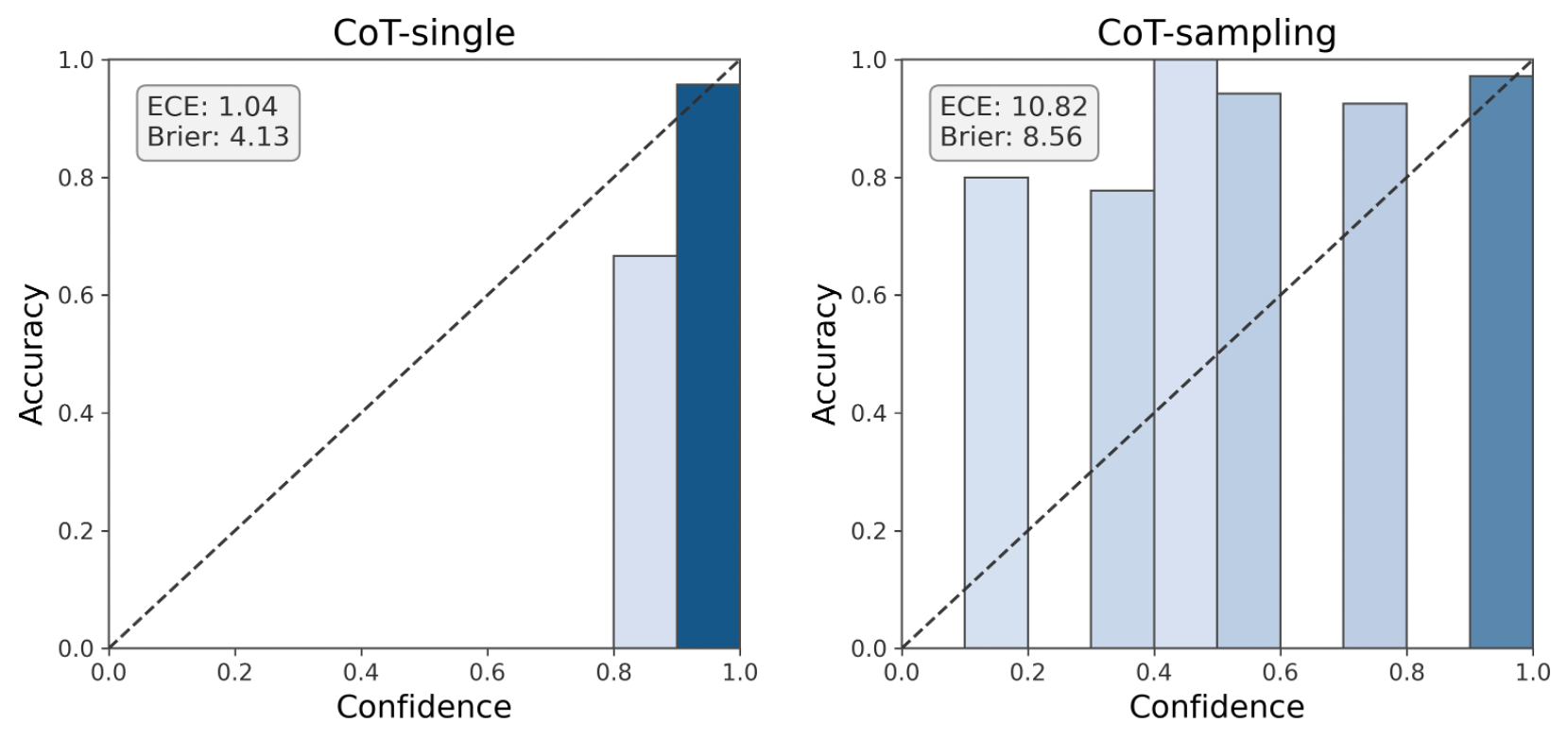}
    \caption{Reliability diagrams of GPT-5-mini on CoQA. In these diagrams, the darker the color, the higher the density.}
    \label{fig:verb-single-sample-ece}
\end{figure}

One notable exception arises on CoQA, where some sampling-aggregated results of DeepSeek-V3.2 and GPT-5-mini are worse than their single-pass counterparts on calibration metrics, as shown in Figure~\ref{fig:verb-single-sample-ece}. A plausible explanation is that CoQA already yields relatively high answer accuracy in some settings, so the single-pass confidence distribution is not far from the empirical correctness rate. Under repeated sampling, however, even a small number of inconsistent generations can noticeably reduce the aggregated confidence, which may in turn worsen calibration. In other words, in high-accuracy settings, sampling may improve robustness while simultaneously introducing additional downward bias.

\subsubsection{Sampling-based}

\begin{table*}[t]
\centering
\caption{Results of sampling-based methods on open-ended QA. We compare representative sampling-based methods under a unified evaluation setting. Avg. denotes the macro-average over all dataset--model pairs shown in the table. The same row-wise shading rule as in Table~2 is used for visualization.}
\label{tab:sample-open}
\resizebox{0.99\textwidth}{!}{
\begin{tabular}{@{}>{\centering\arraybackslash}p{0.9cm}
                >{\centering\arraybackslash}p{1.2cm}
                >{\centering\arraybackslash}p{1.8cm}
                *{11}{>{\centering\arraybackslash}p{0.85cm}}@{\hspace{0.1cm}}}
\toprule
\textbf{Metric} &
\textbf{Dataset} &
\textbf{Model}  &
\textbf{SE} & \makebox[\linewidth][c]{\textbf{\scriptsize SelfCheck}} & \textbf{SEU} & \textbf{EigV} & \textbf{Ecc} & \textbf{Deg} & \textbf{KLE} & \textbf{SINdex} & \textbf{SNNE} & \textbf{SPUQ} & \textbf{InvE} \\
\midrule

\multirow{16}{*}{\textbf{AUROC}} &
\multirow{4}{=}{\parbox[c]{1.2cm}{\centering\textbf{TriviaQA}}} &
Qwen3-4B
& \heatrowhigh{81.02}{78.69}{83.79} & \heatrowhigh{83.79}{78.69}{83.79} & \heatrowhigh{78.69}{78.69}{83.79} 
& \heatrowhigh{80.79}{78.69}{83.79} & \heatrowhigh{80.94}{78.69}{83.79} & \heatrowhigh{83.70}{78.69}{83.79} & \heatrowhigh{83.43}{78.69}{83.79} 
& \heatrowhigh{80.12}{78.69}{83.79} & \heatrowhigh{83.32}{78.69}{83.79} & \heatrowhigh{83.62}{78.69}{83.79} &  \heatrowhigh{79.70}{78.69}{83.79} \\
& & Qwen3-30B
& \heatrowhigh{82.43}{81.27}{84.50} & \heatrowhigh{84.41}{81.27}{84.50} & \heatrowhigh{81.65}{81.27}{84.50} 
& \heatrowhigh{82.70}{81.27}{84.50} & \heatrowhigh{81.27}{81.27}{84.50} & \heatrowhigh{83.72}{81.27}{84.50} & \heatrowhigh{83.74}{81.27}{84.50} 
& \heatrowhigh{82.16}{81.27}{84.50} & \heatrowhigh{82.99}{81.27}{84.50} & \heatrowhigh{84.50}{81.27}{84.50} &  \heatrowhigh{84.21}{81.27}{84.50} \\
& & DeepSeek-V3.2
& \heatrowhigh{72.80}{71.81}{78.22} & \heatrowhigh{72.22}{71.81}{78.22} & \heatrowhigh{74.15}{71.81}{78.22} 
& \heatrowhigh{75.06}{71.81}{78.22} & \heatrowhigh{72.12}{71.81}{78.22} & \heatrowhigh{71.93}{71.81}{78.22} & \heatrowhigh{71.81}{71.81}{78.22} 
& \heatrowhigh{74.20}{71.81}{78.22} & \heatrowhigh{71.83}{71.81}{78.22} & \heatrowhigh{74.71}{71.81}{78.22} &  \heatrowhigh{78.22}{71.81}{78.22} \\
& & GPT-5-mini
& \heatrowhigh{70.83}{70.83}{78.55} & \heatrowhigh{72.41}{70.83}{78.55} & \heatrowhigh{71.75}{70.83}{78.55} 
& \heatrowhigh{76.39}{70.83}{78.55} & \heatrowhigh{73.15}{70.83}{78.55} & \heatrowhigh{73.95}{70.83}{78.55} & \heatrowhigh{74.02}{70.83}{78.55} 
& \heatrowhigh{71.68}{70.83}{78.55} & \heatrowhigh{73.41}{70.83}{78.55} & \heatrowhigh{78.55}{70.83}{78.55} &  \heatrowhigh{77.38}{70.83}{78.55} \\
\cmidrule{2-14}
& \multirow{4}{=}{\parbox[c]{1.2cm}{\centering\textbf{HotpotQA}}} &
Qwen3-4B
& \heatrowhigh{76.13}{71.06}{79.75} & \heatrowhigh{79.75}{71.06}{79.75} & \heatrowhigh{75.85}{71.06}{79.75} 
& \heatrowhigh{76.39}{71.06}{79.75} & \heatrowhigh{77.41}{71.06}{79.75} & \heatrowhigh{78.02}{71.06}{79.75} & \heatrowhigh{78.14}{71.06}{79.75} 
& \heatrowhigh{76.09}{71.06}{79.75} & \heatrowhigh{78.43}{71.06}{79.75} & \heatrowhigh{76.15}{71.06}{79.75} &  \heatrowhigh{71.06}{71.06}{79.75} \\
& & Qwen3-30B
& \heatrowhigh{79.94}{78.33}{81.34} & \heatrowhigh{79.93}{78.33}{81.34} & \heatrowhigh{78.33}{78.33}{81.34} 
& \heatrowhigh{79.91}{78.33}{81.34} & \heatrowhigh{80.20}{78.33}{81.34} & \heatrowhigh{81.15}{78.33}{81.34} & \heatrowhigh{80.70}{78.33}{81.34} 
& \heatrowhigh{79.26}{78.33}{81.34} & \heatrowhigh{81.18}{78.33}{81.34} & \heatrowhigh{80.19}{78.33}{81.34} &  \heatrowhigh{78.85}{78.33}{81.34} \\
& & DeepSeek-V3.2
& \heatrowhigh{81.61}{76.85}{82.67} & \heatrowhigh{82.61}{76.85}{82.67} & \heatrowhigh{78.82}{76.85}{82.67} 
& \heatrowhigh{81.85}{76.85}{82.67} & \heatrowhigh{80.34}{76.85}{82.67} & \heatrowhigh{82.67}{76.85}{82.67} & \heatrowhigh{82.36}{76.85}{82.67} 
& \heatrowhigh{80.43}{76.85}{82.67} & \heatrowhigh{82.66}{76.85}{82.67} & \heatrowhigh{81.17}{76.85}{82.67} &  \heatrowhigh{76.85}{76.85}{82.67} \\
& & GPT-5-mini
& \heatrowhigh{74.94}{74.94}{82.22} & \heatrowhigh{79.05}{74.94}{82.22} & \heatrowhigh{76.39}{74.94}{82.22} 
& \heatrowhigh{77.82}{74.94}{82.22} & \heatrowhigh{75.81}{74.94}{82.22} & \heatrowhigh{78.04}{74.94}{82.22} & \heatrowhigh{77.68}{74.94}{82.22} 
& \heatrowhigh{76.85}{74.94}{82.22} & \heatrowhigh{77.99}{74.94}{82.22} & \heatrowhigh{80.72}{74.94}{82.22} &  \heatrowhigh{78.23}{74.94}{82.22} \\
\cmidrule{2-14}
& \multirow{4}{=}{\parbox[c]{1.2cm}{\centering\textbf{CoQA}}} &
Qwen3-4B
& \heatrowhigh{73.63}{66.36}{77.81} & \heatrowhigh{76.70}{66.36}{77.81} & \heatrowhigh{76.43}{66.36}{77.81} 
& \heatrowhigh{76.62}{66.36}{77.81} & \heatrowhigh{76.28}{66.36}{77.81} & \heatrowhigh{77.81}{66.36}{77.81} & \heatrowhigh{77.60}{66.36}{77.81} 
& \heatrowhigh{76.50}{66.36}{77.81} & \heatrowhigh{77.27}{66.36}{77.81} & \heatrowhigh{70.46}{66.36}{77.81} &  \heatrowhigh{66.36}{66.36}{77.81} \\
& & Qwen3-30B
& \heatrowhigh{71.30}{71.30}{75.03} & \heatrowhigh{72.38}{71.30}{75.03} & \heatrowhigh{73.15}{71.30}{75.03} 
& \heatrowhigh{72.65}{71.30}{75.03} & \heatrowhigh{75.03}{71.30}{75.03} & \heatrowhigh{74.07}{71.30}{75.03} & \heatrowhigh{74.12}{71.30}{75.03} 
& \heatrowhigh{73.26}{71.30}{75.03} & \heatrowhigh{74.09}{71.30}{75.03} & \heatrowhigh{72.80}{71.30}{75.03} &  \heatrowhigh{71.63}{71.30}{75.03} \\
& & DeepSeek-V3.2
& \heatrowhigh{79.16}{72.72}{82.48} & \heatrowhigh{82.48}{72.72}{82.48} & \heatrowhigh{76.98}{72.72}{82.48} 
& \heatrowhigh{81.27}{72.72}{82.48} & \heatrowhigh{80.25}{72.72}{82.48} & \heatrowhigh{81.37}{72.72}{82.48} & \heatrowhigh{81.99}{72.72}{82.48} 
& \heatrowhigh{79.61}{72.72}{82.48} & \heatrowhigh{82.15}{72.72}{82.48} & \heatrowhigh{76.50}{72.72}{82.48} &  \heatrowhigh{72.72}{72.72}{82.48} \\
& & GPT-5-mini
& \heatrowhigh{69.25}{69.25}{76.25} & \heatrowhigh{73.59}{69.25}{76.25} & \heatrowhigh{76.25}{69.25}{76.25} 
& \heatrowhigh{69.62}{69.25}{76.25} & \heatrowhigh{73.00}{69.25}{76.25} & \heatrowhigh{71.72}{69.25}{76.25} & \heatrowhigh{71.39}{69.25}{76.25} 
& \heatrowhigh{74.66}{69.25}{76.25} & \heatrowhigh{71.29}{69.25}{76.25} & \heatrowhigh{71.55}{69.25}{76.25} &  \heatrowhigh{71.65}{69.25}{76.25} \\
\cmidrule{2-14}
& \multicolumn{2}{c}{\textbf{Avg.}} &
\heatrowhigh{76.09}{75.57}{78.28} & \heatrowhigh{78.28}{75.57}{78.28} & \heatrowhigh{76.54}{75.57}{78.28} 
& \heatrowhigh{77.63}{75.57}{78.28} & \heatrowhigh{77.15}{75.57}{78.28} & \heatrowhigh{78.18}{75.57}{78.28} & \heatrowhigh{78.08}{75.57}{78.28} 
& \heatrowhigh{77.07}{75.57}{78.28} & \heatrowhigh{78.05}{75.57}{78.28} & \heatrowhigh{77.58}{75.57}{78.28} &  \heatrowhigh{75.57}{75.57}{78.28} \\
\midrule

\multirow{16}{*}{\textbf{ECE}} &
\multirow{4}{=}{\parbox[c]{1.2cm}{\centering\textbf{TriviaQA}}} &
Qwen3-4B
& \heatrowlow{20.91}{8.88}{29.04} & \heatrowlow{22.36}{8.88}{29.04} & \heatrowlow{29.04}{8.88}{29.04}
& \heatrowlow{26.84}{8.88}{29.04} & \heatrowlow{15.52}{8.88}{29.04} & \heatrowlow{15.09}{8.88}{29.04} & \heatrowlow{13.47}{8.88}{29.04} 
& \heatrowlow{24.99}{8.88}{29.04} & \heatrowlow{27.74}{8.88}{29.04} & \heatrowlow{20.60}{8.88}{29.04} &  \heatrowlow{8.88}{8.88}{29.04} \\
& & Qwen3-30B
& \heatrowlow{8.54}{8.49}{12.89} & \heatrowlow{11.07}{8.49}{12.89} & \heatrowlow{12.33}{8.49}{12.89} 
& \heatrowlow{12.89}{8.49}{12.89} & \heatrowlow{8.49}{8.49}{12.89} & \heatrowlow{8.57}{8.49}{12.89} & \heatrowlow{8.85}{8.49}{12.89} 
& \heatrowlow{11.47}{8.49}{12.89} & \heatrowlow{12.52}{8.49}{12.89} & \heatrowlow{9.98}{8.49}{12.89} &  \heatrowlow{8.76}{8.49}{12.89} \\
& & DeepSeek-V3.2
& \heatrowlow{7.23}{5.28}{11.60} & \heatrowlow{7.29}{5.28}{11.60} & \heatrowlow{10.83}{5.28}{11.60} 
& \heatrowlow{7.66}{5.28}{11.60} & \heatrowlow{9.44}{5.28}{11.60} & \heatrowlow{6.11}{5.28}{11.60} & \heatrowlow{6.58}{5.28}{11.60} 
& \heatrowlow{11.60}{5.28}{11.60} & \heatrowlow{5.31}{5.28}{11.60} & \heatrowlow{5.57}{5.28}{11.60} &  \heatrowlow{8.29}{5.28}{11.60} \\
& & GPT-5-mini
& \heatrowlow{13.87}{4.02}{16.48} & \heatrowlow{7.98}{4.02}{16.48} & \heatrowlow{11.44}{4.02}{16.48} 
& \heatrowlow{8.68}{4.02}{16.48} & \heatrowlow{16.48}{4.02}{16.48} & \heatrowlow{9.53}{4.02}{16.48} & \heatrowlow{8.98}{4.02}{16.48} 
& \heatrowlow{11.37}{4.02}{16.48} & \heatrowlow{8.31}{4.02}{16.48} & \heatrowlow{5.56}{4.02}{16.48} &  \heatrowlow{8.87}{4.02}{16.48} \\
\cmidrule{2-14}
& \multirow{4}{=}{\parbox[c]{1.2cm}{\centering\textbf{HotpotQA}}} &
Qwen3-4B
& \heatrowlow{32.57}{22.52}{55.29} & \heatrowlow{35.35}{22.52}{55.29} & \heatrowlow{55.29}{22.52}{55.29} 
& \heatrowlow{40.41}{22.52}{55.29} & \heatrowlow{26.36}{22.52}{55.29} & \heatrowlow{28.75}{22.52}{55.29} & \heatrowlow{24.62}{22.52}{55.29} 
& \heatrowlow{46.40}{22.52}{55.29} & \heatrowlow{41.42}{22.52}{55.29} & \heatrowlow{34.10}{22.52}{55.29} &  \heatrowlow{22.52}{22.52}{55.29} \\
& & Qwen3-30B
& \heatrowlow{20.47}{13.32}{44.63} & \heatrowlow{27.02}{13.32}{44.63} & \heatrowlow{44.63}{13.32}{44.63} 
& \heatrowlow{29.15}{13.32}{44.63} & \heatrowlow{15.40}{13.32}{44.63} & \heatrowlow{19.30}{13.32}{44.63} & \heatrowlow{17.56}{13.32}{44.63} 
& \heatrowlow{35.56}{13.32}{44.63} & \heatrowlow{20.44}{13.32}{44.63} & \heatrowlow{30.70}{13.32}{44.63} &  \heatrowlow{13.32}{13.32}{44.63} \\
& & DeepSeek-V3.2
& \heatrowlow{14.17}{8.15}{23.10} & \heatrowlow{16.93}{8.15}{23.10} & \heatrowlow{22.59}{8.15}{23.10} 
& \heatrowlow{21.16}{8.15}{23.10} & \heatrowlow{10.03}{8.15}{23.10} & \heatrowlow{11.30}{8.15}{23.10} & \heatrowlow{11.19}{8.15}{23.10} 
& \heatrowlow{23.10}{8.15}{23.10} & \heatrowlow{12.92}{8.15}{23.10} & \heatrowlow{17.46}{8.15}{23.10} &  \heatrowlow{8.15}{8.15}{23.10} \\
& & GPT-5-mini
& \heatrowlow{15.19}{9.82}{23.58} & \heatrowlow{22.00}{9.82}{23.58} & \heatrowlow{21.85}{9.82}{23.58} 
& \heatrowlow{21.88}{9.82}{23.58} & \heatrowlow{15.02}{9.82}{23.58} & \heatrowlow{12.14}{9.82}{23.58} & \heatrowlow{12.89}{9.82}{23.58}
& \heatrowlow{23.58}{9.82}{23.58} & \heatrowlow{13.25}{9.82}{23.58} & \heatrowlow{17.26}{9.82}{23.58} &  \heatrowlow{9.82}{9.82}{23.58} \\
\cmidrule{2-14}
& \multirow{4}{=}{\parbox[c]{1.2cm}{\centering\textbf{CoQA}}} &
Qwen3-4B
& \heatrowlow{11.14}{7.55}{13.97} & \heatrowlow{11.94}{7.55}{13.97} & \heatrowlow{9.65}{7.55}{13.97} 
& \heatrowlow{9.26}{7.55}{13.97} & \heatrowlow{13.97}{7.55}{13.97} & \heatrowlow{8.37}{7.55}{13.97} & \heatrowlow{9.29}{7.55}{13.97} 
& \heatrowlow{8.26}{7.55}{13.97} & \heatrowlow{7.55}{7.55}{13.97} & \heatrowlow{7.96}{7.55}{13.97} &  \heatrowlow{10.66}{7.55}{13.97} \\
& & Qwen3-30B
& \heatrowlow{10.60}{6.37}{15.16} & \heatrowlow{10.31}{6.37}{15.16} & \heatrowlow{8.07}{6.37}{15.16} 
& \heatrowlow{8.29}{6.37}{15.16} & \heatrowlow{15.16}{6.37}{15.16} & \heatrowlow{9.63}{6.37}{15.16} & \heatrowlow{9.73}{6.37}{15.16} 
& \heatrowlow{12.74}{6.37}{15.16} & \heatrowlow{9.29}{6.37}{15.16} & \heatrowlow{6.73}{6.37}{15.16} &  \heatrowlow{12.72}{6.37}{15.16} \\
& & DeepSeek-V3.2
& \heatrowlow{17.12}{4.25}{27.41} & \heatrowlow{4.25}{4.25}{27.41} & \heatrowlow{11.92}{4.25}{27.41} 
& \heatrowlow{12.49}{4.25}{27.41} & \heatrowlow{26.10}{4.25}{27.41} & \heatrowlow{17.90}{4.25}{27.41} & \heatrowlow{20.98}{4.25}{27.41} 
& \heatrowlow{21.51}{4.25}{27.41} & \heatrowlow{18.09}{4.25}{27.41} & \heatrowlow{10.43}{4.25}{27.41} &  \heatrowlow{27.41}{4.25}{27.41}  \\
& & GPT-5-mini
& \heatrowlow{14.80}{4.79}{22.98} & \heatrowlow{4.79}{4.79}{22.98} & \heatrowlow{12.05}{4.79}{22.98} 
& \heatrowlow{11.61}{4.79}{22.98} & \heatrowlow{22.98}{4.79}{22.98} & \heatrowlow{15.43}{4.79}{22.98} & \heatrowlow{15.56}{4.79}{22.98} 
& \heatrowlow{13.41}{4.79}{22.98} & \heatrowlow{15.00}{4.79}{22.98} & \heatrowlow{7.06}{4.79}{22.98} &  \heatrowlow{5.03}{4.79}{22.98} \\
\cmidrule{2-14}
& \multicolumn{2}{c}{\textbf{Avg.}} &
\heatrowlow{15.55}{12.04}{20.81} & \heatrowlow{15.11}{12.04}{20.81} & \heatrowlow{20.81}{12.04}{20.81} 
& \heatrowlow{17.53}{12.04}{20.81} & \heatrowlow{16.25}{12.04}{20.81} & \heatrowlow{13.51}{12.04}{20.81} & \heatrowlow{13.31}{12.04}{20.81} 
& \heatrowlow{20.33}{12.04}{20.81} & \heatrowlow{15.97}{12.04}{20.81} & \heatrowlow{14.45}{12.04}{20.81} &  \heatrowlow{12.04}{12.04}{20.81} \\
\midrule

\multirow{16}{*}{\textbf{Brier}} &
\multirow{4}{=}{\parbox[c]{1.2cm}{\centering\textbf{TriviaQA}}} &
Qwen3-4B
& \heatrowlow{21.42}{18.55}{28.00} & \heatrowlow{22.39}{18.55}{28.00} & \heatrowlow{28.00}{18.55}{28.00} 
& \heatrowlow{24.75}{18.55}{28.00} & \heatrowlow{19.64}{18.55}{28.00} & \heatrowlow{19.21}{18.55}{28.00} & \heatrowlow{19.00}{18.55}{28.00} 
& \heatrowlow{24.80}{18.55}{28.00} & \heatrowlow{25.48}{18.55}{28.00} & \heatrowlow{21.44}{18.55}{28.00} &  \heatrowlow{18.55}{18.55}{28.00} \\
& & Qwen3-30B
& \heatrowlow{13.32}{13.14}{15.33} & \heatrowlow{13.28}{13.14}{15.33} & \heatrowlow{15.33}{13.14}{15.33} 
& \heatrowlow{14.87}{13.14}{15.33} & \heatrowlow{13.44}{13.14}{15.33} & \heatrowlow{13.14}{13.14}{15.33} & \heatrowlow{13.37}{13.14}{15.33} 
& \heatrowlow{14.91}{13.14}{15.33} & \heatrowlow{15.03}{13.14}{15.33} & \heatrowlow{13.86}{13.14}{15.33} &  \heatrowlow{14.27}{13.14}{15.33} \\
& & DeepSeek-V3.2
& \heatrowlow{9.76}{8.33}{11.59} & \heatrowlow{8.33}{8.33}{11.59} & \heatrowlow{11.24}{8.33}{11.59} 
& \heatrowlow{9.48}{8.33}{11.59} & \heatrowlow{10.23}{8.33}{11.59} & \heatrowlow{9.43}{8.33}{11.59} & \heatrowlow{9.61}{8.33}{11.59} 
& \heatrowlow{11.50}{8.33}{11.59} & \heatrowlow{9.15}{8.33}{11.59} & \heatrowlow{8.85}{8.33}{11.59} &  \heatrowlow{11.59}{8.33}{11.59} \\
& & GPT-5-mini
& \heatrowlow{13.21}{8.51}{15.23} & \heatrowlow{9.36}{8.51}{15.23} & \heatrowlow{11.67}{8.51}{15.23} 
& \heatrowlow{9.54}{8.51}{15.23} & \heatrowlow{15.23}{8.51}{15.23} & \heatrowlow{10.90}{8.51}{15.23} & \heatrowlow{10.44}{8.51}{15.23} 
& \heatrowlow{11.61}{8.51}{15.23} & \heatrowlow{10.68}{8.51}{15.23} & \heatrowlow{8.54}{8.51}{15.23} &  \heatrowlow{10.77}{8.51}{15.23} \\
\cmidrule{2-14}
& \multirow{4}{=}{\parbox[c]{1.2cm}{\centering\textbf{HotpotQA}}} &
Qwen3-4B
& \heatrowlow{29.88}{22.18}{48.48} & \heatrowlow{31.65}{22.18}{48.48} & \heatrowlow{48.48}{22.18}{48.48} 
& \heatrowlow{35.23}{22.18}{48.48} & \heatrowlow{26.15}{22.18}{48.48} & \heatrowlow{27.09}{22.18}{48.48} & \heatrowlow{25.58}{22.18}{48.48} 
& \heatrowlow{42.09}{22.18}{48.48} & \heatrowlow{35.73}{22.18}{48.48} & \heatrowlow{29.66}{22.18}{48.48} &  \heatrowlow{22.18}{22.18}{48.48} \\
& & Qwen3-30B
& \heatrowlow{23.08}{19.10}{40.52} & \heatrowlow{26.84}{19.10}{40.52} & \heatrowlow{40.52}{19.10}{40.52} 
& \heatrowlow{27.88}{19.10}{40.52} & \heatrowlow{21.14}{19.10}{40.52} & \heatrowlow{22.11}{19.10}{40.52} & \heatrowlow{21.80}{19.10}{40.52} 
& \heatrowlow{35.10}{19.10}{40.52} & \heatrowlow{22.84}{19.10}{40.52} & \heatrowlow{28.23}{19.10}{40.52} &  \heatrowlow{19.10}{19.10}{40.52} \\
& & DeepSeek-V3.2
& \heatrowlow{18.30}{17.80}{24.55} & \heatrowlow{20.06}{17.80}{24.55} & \heatrowlow{24.29}{17.80}{24.55} 
& \heatrowlow{21.98}{17.80}{24.55} & \heatrowlow{17.80}{17.80}{24.55} & \heatrowlow{17.80}{17.80}{24.55} & \heatrowlow{18.05}{17.80}{24.55} 
& \heatrowlow{24.55}{17.80}{24.55} & \heatrowlow{18.31}{17.80}{24.55} & \heatrowlow{21.06}{17.80}{24.55} &  \heatrowlow{19.95}{17.80}{24.55} \\
& & GPT-5-mini
& \heatrowlow{21.48}{19.37}{26.04} & \heatrowlow{23.19}{19.37}{26.04} & \heatrowlow{23.93}{19.37}{26.04} 
& \heatrowlow{22.98}{19.37}{26.04} & \heatrowlow{21.02}{19.37}{26.04} & \heatrowlow{19.37}{19.37}{26.04} & \heatrowlow{19.52}{19.37}{26.04} 
& \heatrowlow{26.04}{19.37}{26.04} & \heatrowlow{19.44}{19.37}{26.04} & \heatrowlow{20.61}{19.37}{26.04} &  \heatrowlow{19.53}{19.37}{26.04} \\
\cmidrule{2-14}
& \multirow{4}{=}{\parbox[c]{1.2cm}{\centering\textbf{CoQA}}} &
Qwen3-4B
& \heatrowlow{12.75}{11.16}{14.73} & \heatrowlow{12.86}{11.16}{14.73} & \heatrowlow{11.95}{11.16}{14.73} 
& \heatrowlow{11.65}{11.16}{14.73} & \heatrowlow{14.09}{11.16}{14.73} & \heatrowlow{11.23}{11.16}{14.73} & \heatrowlow{12.05}{11.16}{14.73} 
& \heatrowlow{12.07}{11.16}{14.73} & \heatrowlow{11.16}{11.16}{14.73} & \heatrowlow{13.45}{11.16}{14.73} &  \heatrowlow{14.73}{11.16}{14.73} \\
& & Qwen3-30B
& \heatrowlow{11.17}{10.18}{14.56} & \heatrowlow{11.18}{10.18}{14.56} & \heatrowlow{11.53}{10.18}{14.56} 
& \heatrowlow{10.18}{10.18}{14.56} & \heatrowlow{13.80}{10.18}{14.56} & \heatrowlow{10.92}{10.18}{14.56} & \heatrowlow{11.00}{10.18}{14.56} 
& \heatrowlow{13.02}{10.18}{14.56} & \heatrowlow{10.71}{10.18}{14.56} & \heatrowlow{11.54}{10.18}{14.56} &  \heatrowlow{14.56}{10.18}{14.56} \\
& & DeepSeek-V3.2
& \heatrowlow{13.91}{7.00}{21.46} & \heatrowlow{7.00}{7.00}{21.46} & \heatrowlow{13.85}{7.00}{21.46} 
& \heatrowlow{11.46}{7.00}{21.46} & \heatrowlow{21.46}{7.00}{21.46} & \heatrowlow{14.22}{7.00}{21.46} & \heatrowlow{16.90}{7.00}{21.46} 
& \heatrowlow{17.65}{7.00}{21.46} & \heatrowlow{14.32}{7.00}{21.46} & \heatrowlow{10.47}{7.00}{21.46} &  \heatrowlow{19.55}{7.00}{21.46} \\
& & GPT-5-mini
& \heatrowlow{11.64}{5.80}{18.47} & \heatrowlow{5.80}{5.80}{18.47} & \heatrowlow{10.84}{5.80}{18.47} 
& \heatrowlow{10.31}{5.80}{18.47} & \heatrowlow{18.47}{5.80}{18.47} & \heatrowlow{12.71}{5.80}{18.47} & \heatrowlow{12.69}{5.80}{18.47}
& \heatrowlow{10.45}{5.80}{18.47} & \heatrowlow{12.37}{5.80}{18.47} & \heatrowlow{7.29}{5.80}{18.47} &  \heatrowlow{6.23}{5.80}{18.47} \\
\cmidrule{2-14}
& \multicolumn{2}{c}{\textbf{Avg.}} &
\heatrowlow{16.66}{15.68}{20.97} & \heatrowlow{16.00}{15.68}{20.97} & \heatrowlow{20.97}{15.68}{20.97} 
& \heatrowlow{17.52}{15.68}{20.97} & \heatrowlow{17.71}{15.68}{20.97} & \heatrowlow{15.68}{15.68}{20.97} & \heatrowlow{15.83}{15.68}{20.97} 
& \heatrowlow{20.28}{15.68}{20.97} & \heatrowlow{17.10}{15.68}{20.97} & \heatrowlow{16.25}{15.68}{20.97} &  \heatrowlow{15.92}{15.68}{20.97}\\
\bottomrule
\end{tabular}}
\end{table*}

Table~\ref{tab:sample-open} reports the results of sampling-based methods on the three open-ended QA datasets. Since all methods are evaluated using the same low-temperature predicted answers, their answer accuracy is identical across methods, with an average accuracy of 69.59. Therefore, this comparison mainly isolates the quality of the uncertainty signal itself, rather than the effect of differences in answer generation.

It is worth noting that AUROC and ECE/Brier are not equally comparable for sampling-based methods. AUROC mainly evaluates whether uncertainty scores can distinguish correct predictions from incorrect ones. It depends only on the relative ranking of scores and is insensitive to their scale. In contrast, ECE and Brier score require the scores themselves to be numerically calibrated. However, many sampling-based methods do not directly output calibrated confidence scores in the range $[0,1]$. Therefore, when computing ECE and Brier score, these scores usually need post-hoc mapping. This makes the absolute ECE and Brier values less directly comparable across sampling-based methods than AUROC. For this reason, we mainly use AUROC to compare the discrimination ability of these methods, while treating ECE and Brier as auxiliary references for calibration trends.

\textbf{Overall, methods based on NLI relation matrices achieve better AUROC performance than methods based on continuous semantic-space representations.} Specifically, SelfCheck~\cite{manakul2023selfcheckgpt}, Deg~\cite{lin2023generating}, KLE~\cite{nikitin2024kernel}, and SNNE~\cite{nguyen2025beyond} show strong and stable AUROC performance across most datasets and models. On average, these methods generally outperform methods such as SEU~\cite{grewal2024improving}, and SINdex~\cite{abdaljalil2025sindex}, which rely more heavily on embedding similarity or continuous semantic-space structure. A possible reason is that correctness in open-ended QA often depends on fine-grained factual consistency rather than broad semantic similarity. Two responses may be close in embedding space but still conflict on key entities, dates, quantities, or relations. Conversely, responses with different surface forms may express the same correct answer. Therefore, compared with simple vector similarity, the discrete semantic relations provided by NLI, including entailment, neutral, and contradiction, are more closely aligned with the central concern of UE, namely whether different responses support each other, are semantically equivalent, or contradict each other.

Although NLI-based methods share pairwise semantic relation information, they aggregate these relations in different ways. SelfCheck~\cite{manakul2023selfcheckgpt} mainly compares the predicted answer with sampled responses, directly measuring whether the current prediction is supported by other samples. Deg~\cite{lin2023generating} treats the NLI relation matrix as the adjacency matrix of a semantic graph and measures the concentration of the response set through node connectivity. KLE~\cite{nikitin2024kernel} constructs a kernel from the graph Laplacian and uses spectral entropy to characterize semantic diversity among responses. SNNE~\cite{nguyen2025beyond} uses pairwise entailment scores to construct a soft neighborhood entropy and does not rely on hard clustering, which allows it to produce smoother uncertainty signals when responses partially overlap in meaning. These results suggest that the key to sampling-based UE is not only generating multiple responses, but also modeling their semantic relations and aggregating these relations effectively into uncertainty scores.

SPUQ~\cite{gao2024spuq} shows competitive but not leading AUROC performance. This suggests that prompt perturbation is useful, since the method probes prediction stability under input changes rather than relying only on disagreement under a fixed input. At the same time, the results indicate that prompt perturbation alone is not sufficient to make SPUQ dominate among sampling-based methods. InvE~\cite{song2025inv} shows a different pattern, as its AUROC is not among the strongest, but its ECE and Brier scores are comparatively better. A plausible interpretation is that InvE focuses more on the structural consistency between input perturbations and output perturbations, which tends to yield smoother uncertainty estimates. Such scores may be less aggressive for ranking, but more likely to preserve a stable monotonic relation with empirical correctness. This may lead to better calibration-oriented metrics such as ECE and Brier.

\subsubsection{Other Approaches}

\begin{table}[htbp]
\centering
\caption{Results of other black-box UE methods on open-ended QA. We compare representative explanation-based, multi-agent, and hybrid methods. Avg. denotes the macro-average over all dataset--model pairs shown in the table. The same row-wise shading rule as in Table~2 is used for visualization.}
\label{tab:agent_hybrid_metrics}
\resizebox{0.49\textwidth}{!}{
\begin{tabular}{@{}>{\centering\arraybackslash}p{0.8cm}
                >{\centering\arraybackslash}p{1.0cm}
                >{\centering\arraybackslash}p{1.8cm}
                *{7}{>{\centering\arraybackslash}p{1.0cm}}@{\hspace{0.05cm}}}

\toprule
\textbf{Metric} &
\textbf{Dataset} &
\textbf{Model}  &
\textbf{COTA} & \makebox[\linewidth][c]{\textbf{\scriptsize PathWeight}} & \textbf{Collab} & \makebox[\linewidth][c]{\textbf{\scriptsize ArgLLMs}} & \makebox[\linewidth][c]{\textbf{\scriptsize BSDetector}} & \textbf{DiNCo} & \makebox[\linewidth][c]{\textbf{\scriptsize SteerConf}} \\
\midrule

\multirow{13}{*}{\textbf{Acc}} &
\multirow{4}{=}{\parbox[c]{1.2cm}{\centering\textbf{TriviaQA}}} &
Qwen3-4B
& \heatrowhigh{48.23}{45.25}{50.50} & \heatrowhigh{49.70}{45.25}{50.50} & \heatrowhigh{50.50}{45.25}{50.50} & \heatrowhigh{48.23}{45.25}{50.50} & \heatrowhigh{48.23}{45.25}{50.50} & \heatrowhigh{49.49}{45.25}{50.50} & \heatrowhigh{45.25}{45.25}{50.50} \\
& & Qwen3-30B
& \heatrowhigh{73.04}{70.80}{75.00} & \heatrowhigh{73.20}{70.80}{75.00} & \heatrowhigh{72.34}{70.80}{75.00} & \heatrowhigh{73.19}{70.80}{75.00} & \heatrowhigh{73.04}{70.80}{75.00} & \heatrowhigh{75.00}{70.80}{75.00} & \heatrowhigh{70.80}{70.80}{75.00} \\
& & DeepSeek-V3.2
& \heatrowhigh{88.60}{87.00}{88.60} & \heatrowhigh{88.00}{87.00}{88.60} & \heatrowhigh{87.00}{87.00}{88.60} & \heatrowhigh{88.60}{87.00}{88.60} & \heatrowhigh{88.60}{87.00}{88.60} & \heatrowhigh{88.40}{87.00}{88.60} & \heatrowhigh{87.60}{87.00}{88.60} \\
& & GPT-5-mini
& \heatrowhigh{88.40}{87.80}{88.80} & \heatrowhigh{88.00}{87.80}{88.80} & \heatrowhigh{87.80}{87.80}{88.80} & \heatrowhigh{88.40}{87.80}{88.80} & \heatrowhigh{88.40}{87.80}{88.80} & \heatrowhigh{88.60}{87.80}{88.80} & \heatrowhigh{88.80}{87.80}{88.80} \\
\cmidrule{2-10}
& \multirow{4}{=}{\parbox[c]{1.2cm}{\centering\textbf{HotpotQA}}} &
Qwen3-4B
& \heatrowhigh{25.88}{23.69}{25.88} & \heatrowhigh{24.60}{23.69}{25.88} & \heatrowhigh{23.69}{23.69}{25.88} & \heatrowhigh{25.88}{23.69}{25.88} & \heatrowhigh{25.88}{23.69}{25.88} & \heatrowhigh{24.18}{23.69}{25.88} & \heatrowhigh{25.47}{23.69}{25.88} \\
& & Qwen3-30B
& \heatrowhigh{37.27}{35.20}{37.80} & \heatrowhigh{35.40}{35.20}{37.80} & \heatrowhigh{37.80}{35.20}{37.80} & \heatrowhigh{37.42}{35.20}{37.80} & \heatrowhigh{37.27}{35.20}{37.80} & \heatrowhigh{36.49}{35.20}{37.80} & \heatrowhigh{35.20}{35.20}{37.80} \\
& & DeepSeek-V3.2
& \heatrowhigh{56.94}{54.20}{58.80} & \heatrowhigh{57.63}{54.20}{58.80} & \heatrowhigh{58.80}{54.20}{58.80} & \heatrowhigh{57.03}{54.20}{58.80} & \heatrowhigh{57.03}{54.20}{58.80} & \heatrowhigh{57.31}{54.20}{58.80} & \heatrowhigh{54.20}{54.20}{58.80} \\
& & GPT-5-mini
& \heatrowhigh{60.60}{59.20}{62.40} & \heatrowhigh{62.40}{59.20}{62.40} & \heatrowhigh{62.00}{59.20}{62.40} & \heatrowhigh{60.60}{59.20}{62.40} & \heatrowhigh{60.60}{59.20}{62.40} & \heatrowhigh{59.20}{59.20}{62.40} & \heatrowhigh{62.00}{59.20}{62.40} \\
\cmidrule{2-10}
& \multirow{4}{=}{\parbox[c]{1.2cm}{\centering\textbf{CoQA}}} &
Qwen3-4B
& \heatrowhigh{84.20}{82.20}{84.20} & \heatrowhigh{83.00}{82.20}{84.20} & \heatrowhigh{82.20}{82.20}{84.20} & \heatrowhigh{84.20}{82.20}{84.20} & \heatrowhigh{84.20}{82.20}{84.20} & \heatrowhigh{83.80}{82.20}{84.20} & \heatrowhigh{84.00}{82.20}{84.20} \\
& & Qwen3-30B
& \heatrowhigh{87.40}{85.00}{88.40} & \heatrowhigh{87.80}{85.00}{88.40} & \heatrowhigh{85.00}{85.00}{88.40} & \heatrowhigh{87.35}{85.00}{88.40} & \heatrowhigh{87.40}{85.00}{88.40} & \heatrowhigh{88.40}{85.00}{88.40} & \heatrowhigh{87.20}{85.00}{88.40} \\
& & DeepSeek-V3.2
& \heatrowhigh{90.40}{88.00}{93.20} & \heatrowhigh{93.20}{88.00}{93.20} & \heatrowhigh{88.00}{88.00}{93.20} & \heatrowhigh{90.36}{88.00}{93.20} & \heatrowhigh{90.40}{88.00}{93.20} & \heatrowhigh{89.20}{88.00}{93.20} & \heatrowhigh{90.80}{88.00}{93.20} \\
& & GPT-5-mini
& \heatrowhigh{94.00}{89.40}{96.40} & \heatrowhigh{94.40}{89.40}{96.40} & \heatrowhigh{89.40}{89.40}{96.40} & \heatrowhigh{94.00}{89.40}{96.40} & \heatrowhigh{94.00}{89.40}{96.40} & \heatrowhigh{93.40}{89.40}{96.40} & \heatrowhigh{96.40}{89.40}{96.40} \\
\cmidrule{2-10}
& \multicolumn{2}{c}{\textbf{Avg.}}
& \heatrowhigh{69.58}{68.71}{71.63} & \heatrowhigh{69.69}{68.71}{71.63} & \heatrowhigh{68.71}{68.71}{71.63} & \heatrowhigh{69.61}{68.71}{71.63} & \heatrowhigh{69.58}{68.71}{71.63} & \heatrowhigh{69.45}{68.71}{71.63} & \heatrowhigh{71.63}{68.71}{71.63}\\
\midrule

\multirow{13}{*}{\textbf{AUROC}} &
\multirow{4}{=}{\parbox[c]{1.2cm}{\centering\textbf{TriviaQA}}} &
Qwen3-4B
& \heatrowhigh{77.37}{77.37}{88.67} & \heatrowhigh{78.11}{77.37}{88.67} & \heatrowhigh{81.78}{77.37}{88.67} & \heatrowhigh{77.64}{77.37}{88.67} & \heatrowhigh{83.59}{77.37}{88.67} & \heatrowhigh{81.84}{77.37}{88.67} & \heatrowhigh{88.67}{77.37}{88.67}\\
& & Qwen3-30B
& \heatrowhigh{72.09}{72.09}{88.18} & \heatrowhigh{82.69}{72.09}{88.18} & \heatrowhigh{81.88}{72.09}{88.18} & \heatrowhigh{78.47}{72.09}{88.18} & \heatrowhigh{83.49}{72.09}{88.18} & \heatrowhigh{78.85}{72.09}{88.18} & \heatrowhigh{88.18}{72.09}{88.18} \\
& & DeepSeek-V3.2
& \heatrowhigh{69.37}{69.37}{80.63} & \heatrowhigh{78.37}{69.37}{80.63} & \heatrowhigh{79.81}{69.37}{80.63} & \heatrowhigh{69.93}{69.37}{80.63} & \heatrowhigh{72.57}{69.37}{80.63} & \heatrowhigh{80.63}{69.37}{80.63} & \heatrowhigh{78.13}{69.37}{80.63} \\
& & GPT-5-mini
& \heatrowhigh{62.44}{62.44}{80.78} & \heatrowhigh{72.56}{62.44}{80.78} & \heatrowhigh{80.78}{62.44}{80.78} & \heatrowhigh{72.96}{62.44}{80.78} & \heatrowhigh{70.80}{62.44}{80.78} & \heatrowhigh{77.99}{62.44}{80.78} & \heatrowhigh{77.30}{62.44}{80.78} \\
\cmidrule{2-10}
& \multirow{4}{=}{\parbox[c]{1.2cm}{\centering\textbf{HotpotQA}}} &
Qwen3-4B
& \heatrowhigh{66.68}{66.68}{86.29} & \heatrowhigh{78.53}{66.68}{86.29} & \heatrowhigh{79.17}{66.68}{86.29} & \heatrowhigh{75.35}{66.68}{86.29} & \heatrowhigh{79.15}{66.68}{86.29} & \heatrowhigh{83.12}{66.68}{86.29} & \heatrowhigh{86.29}{66.68}{86.29} \\
& & Qwen3-30B
& \heatrowhigh{73.41}{73.41}{87.08} & \heatrowhigh{81.31}{73.41}{87.08} & \heatrowhigh{82.18}{73.41}{87.08} & \heatrowhigh{75.97}{73.41}{87.08} & \heatrowhigh{78.46}{73.41}{87.08} & \heatrowhigh{83.50}{73.41}{87.08} & \heatrowhigh{87.08}{73.41}{87.08} \\
& & DeepSeek-V3.2
& \heatrowhigh{77.07}{72.73}{86.34} & \heatrowhigh{79.36}{72.73}{86.34} & \heatrowhigh{81.81}{72.73}{86.34} & \heatrowhigh{72.73}{72.73}{86.34} & \heatrowhigh{81.02}{72.73}{86.34} & \heatrowhigh{83.73}{72.73}{86.34} & \heatrowhigh{86.34}{72.73}{86.34} \\
& & GPT-5-mini
& \heatrowhigh{70.10}{70.10}{83.57} & \heatrowhigh{74.43}{70.10}{83.57} & \heatrowhigh{83.16}{70.10}{83.57} & \heatrowhigh{80.84}{70.10}{83.57} & \heatrowhigh{82.58}{70.10}{83.57} & \heatrowhigh{83.40}{70.10}{83.57} & \heatrowhigh{83.57}{70.10}{83.57} \\
\cmidrule{2-10}
& \multirow{4}{=}{\parbox[c]{1.2cm}{\centering\textbf{CoQA}}} &
Qwen3-4B
& \heatrowhigh{67.06}{57.95}{80.13} & \heatrowhigh{75.43}{57.95}{80.13} & \heatrowhigh{68.40}{57.95}{80.13} & \heatrowhigh{57.95}{57.95}{80.13} & \heatrowhigh{72.68}{57.95}{80.13} & \heatrowhigh{80.13}{57.95}{80.13} & \heatrowhigh{79.45}{57.95}{80.13} \\
& & Qwen3-30B
& \heatrowhigh{59.61}{59.61}{81.05} & \heatrowhigh{70.83}{59.61}{81.05} & \heatrowhigh{75.02}{59.61}{81.05} & \heatrowhigh{65.83}{59.61}{81.05} & \heatrowhigh{71.34}{59.61}{81.05} & \heatrowhigh{81.05}{59.61}{81.05} & \heatrowhigh{78.41}{59.61}{81.05} \\
& & DeepSeek-V3.2
& \heatrowhigh{69.63}{69.63}{86.02} & \heatrowhigh{75.47}{69.63}{86.02} & \heatrowhigh{79.23}{69.63}{86.02} & \heatrowhigh{70.41}{69.63}{86.02} & \heatrowhigh{84.25}{69.63}{86.02} & \heatrowhigh{86.02}{69.63}{86.02} & \heatrowhigh{80.12}{69.63}{86.02} \\
& & GPT-5-mini
& \heatrowhigh{72.32}{65.36}{79.47} & \heatrowhigh{65.36}{65.36}{79.47} & \heatrowhigh{71.05}{65.36}{79.47} & \heatrowhigh{67.16}{65.36}{79.47} & \heatrowhigh{71.38}{65.36}{79.47} & \heatrowhigh{79.47}{65.36}{79.47} & \heatrowhigh{77.56}{65.36}{79.47} \\
\cmidrule{2-10}
& \multicolumn{2}{c}{\textbf{Avg.}}
& \heatrowhigh{69.76}{69.76}{82.64} & \heatrowhigh{76.02}{69.76}{82.64} & \heatrowhigh{78.68}{69.76}{82.64} & \heatrowhigh{72.10}{69.76}{82.64} & \heatrowhigh{77.61}{69.76}{82.64} & \heatrowhigh{81.65}{69.76}{82.64} & \heatrowhigh{82.64}{69.76}{82.64} \\
\midrule

\multirow{13}{*}{\textbf{ECE}} &
\multirow{4}{=}{\parbox[c]{1.2cm}{\centering\textbf{TriviaQA}}} &
Qwen3-4B
& \heatrowlow{5.97}{5.97}{30.78} & \heatrowlow{30.08}{5.97}{30.78} & \heatrowlow{27.71}{5.97}{30.78} & \heatrowlow{30.78}{5.97}{30.78} & \heatrowlow{26.18}{5.97}{30.78} & \heatrowlow{14.00}{5.97}{30.78} & \heatrowlow{16.68}{5.97}{30.78} \\
& & Qwen3-30B
& \heatrowlow{30.49}{5.47}{30.49} & \heatrowlow{14.76}{5.47}{30.49} & \heatrowlow{6.28}{5.47}{30.49} & \heatrowlow{19.60}{5.47}{30.49} & \heatrowlow{11.91}{5.47}{30.49} & \heatrowlow{8.17}{5.47}{30.49} & \heatrowlow{5.47}{5.47}{30.49} \\
& & DeepSeek-V3.2
& \heatrowlow{33.66}{4.75}{33.66} & \heatrowlow{5.88}{4.75}{33.66} & \heatrowlow{5.90}{4.75}{33.66} & \heatrowlow{11.74}{4.75}{33.66} & \heatrowlow{7.26}{4.75}{33.66} & \heatrowlow{7.73}{4.75}{33.66} & \heatrowlow{4.75}{4.75}{33.66} \\
& & GPT-5-mini
& \heatrowlow{41.46}{8.24}{41.46} & \heatrowlow{8.52}{8.24}{41.46} & \heatrowlow{10.05}{8.24}{41.46} & \heatrowlow{10.46}{8.24}{41.46} & \heatrowlow{8.24}{8.24}{41.46} & \heatrowlow{10.95}{8.24}{41.46} & \heatrowlow{32.21}{8.24}{41.46} \\
\cmidrule{2-10}
& \multirow{4}{=}{\parbox[c]{1.2cm}{\centering\textbf{HotpotQA}}} &
Qwen3-4B
& \heatrowlow{15.58}{15.58}{47.17} & \heatrowlow{47.17}{15.58}{47.17} & \heatrowlow{39.94}{15.58}{47.17} & \heatrowlow{36.55}{15.58}{47.17} & \heatrowlow{42.16}{15.58}{47.17} & \heatrowlow{32.53}{15.58}{47.17} & \heatrowlow{20.16}{15.58}{47.17} \\
& & Qwen3-30B
& \heatrowlow{10.66}{10.66}{40.70} & \heatrowlow{38.24}{10.66}{40.70} & \heatrowlow{27.88}{10.66}{40.70} & \heatrowlow{40.70}{10.66}{40.70} & \heatrowlow{32.63}{10.66}{40.70} & \heatrowlow{18.91}{10.66}{40.70} & \heatrowlow{20.16}{10.66}{40.70} \\
& & DeepSeek-V3.2
& \heatrowlow{13.35}{4.34}{27.19} & \heatrowlow{18.44}{4.34}{27.19} & \heatrowlow{9.59}{4.34}{27.19} & \heatrowlow{27.19}{4.34}{27.19} & \heatrowlow{20.11}{4.34}{27.19} & \heatrowlow{10.03}{4.34}{27.19} & \heatrowlow{4.34}{4.34}{27.19} \\
& & GPT-5-mini
& \heatrowlow{13.77}{6.73}{23.12} & \heatrowlow{19.82}{6.73}{23.12} & \heatrowlow{7.13}{6.73}{23.12} & \heatrowlow{20.94}{6.73}{23.12} & \heatrowlow{23.12}{6.73}{23.12} & \heatrowlow{6.73}{6.73}{23.12} & \heatrowlow{19.99}{6.73}{23.12} \\
\cmidrule{2-10}
& \multirow{4}{=}{\parbox[c]{1.2cm}{\centering\textbf{CoQA}}} &
Qwen3-4B
& \heatrowlow{30.36}{6.77}{30.36} & \heatrowlow{8.90}{6.77}{30.36} & \heatrowlow{13.36}{6.77}{30.36} & \heatrowlow{15.78}{6.77}{30.36} & \heatrowlow{12.91}{6.77}{30.36} & \heatrowlow{7.80}{6.77}{30.36} & \heatrowlow{6.77}{6.77}{30.36} \\
& & Qwen3-30B
& \heatrowlow{36.34}{4.87}{36.34} & \heatrowlow{7.37}{4.87}{36.34} & \heatrowlow{8.39}{4.87}{36.34} & \heatrowlow{10.53}{4.87}{36.34} & \heatrowlow{11.13}{4.87}{36.34} & \heatrowlow{11.15}{4.87}{36.34} & \heatrowlow{4.87}{4.87}{36.34} \\
& & DeepSeek-V3.2
& \heatrowlow{53.77}{7.11}{53.77} & \heatrowlow{13.87}{7.11}{53.77} & \heatrowlow{16.31}{7.11}{53.77} & \heatrowlow{7.80}{7.11}{53.77} & \heatrowlow{7.11}{7.11}{53.77} & \heatrowlow{9.96}{7.11}{53.77} & \heatrowlow{13.07}{7.11}{53.77} \\
& & GPT-5-mini
& \heatrowlow{54.21}{5.25}{54.21} & \heatrowlow{12.57}{5.25}{54.21} & \heatrowlow{20.64}{5.25}{54.21} & \heatrowlow{5.85}{5.25}{54.21} & \heatrowlow{5.25}{5.25}{54.21} & \heatrowlow{14.27}{5.25}{54.21} & \heatrowlow{44.00}{5.25}{54.21} \\
\cmidrule{2-10}
& \multicolumn{2}{c}{\textbf{Avg.}}
& \heatrowlow{28.30}{12.69}{28.30} & \heatrowlow{18.80}{12.69}{28.30} & \heatrowlow{16.11}{12.69}{28.30} & \heatrowlow{19.83}{12.69}{28.30} & \heatrowlow{17.33}{12.69}{28.30} & \heatrowlow{12.68}{12.69}{28.30} & \heatrowlow{16.15}{12.69}{28.30} \\
\midrule

\multirow{13}{*}{\textbf{Brier}} &
\multirow{4}{=}{\parbox[c]{1.2cm}{\centering\textbf{TriviaQA}}} &
Qwen3-4B
& \heatrowlow{19.63}{16.77}{30.25} & \heatrowlow{28.39}{16.77}{30.25} & \heatrowlow{26.53}{16.77}{30.25} & \heatrowlow{30.25}{16.77}{30.25} & \heatrowlow{24.24}{16.77}{30.25} & \heatrowlow{19.66}{16.77}{30.25} & \heatrowlow{16.77}{16.77}{30.25} \\
& & Qwen3-30B
& \heatrowlow{27.92}{11.96}{27.92} & \heatrowlow{15.89}{11.96}{27.92} & \heatrowlow{14.16}{11.96}{27.92} & \heatrowlow{19.93}{11.96}{27.92} & \heatrowlow{13.85}{11.96}{27.92} & \heatrowlow{15.21}{11.96}{27.92} & \heatrowlow{11.96}{11.96}{27.92} \\
& & DeepSeek-V3.2
& \heatrowlow{22.89}{8.35}{22.89} & \heatrowlow{8.35}{8.35}{22.89} & \heatrowlow{10.15}{8.35}{22.89} & \heatrowlow{12.07}{8.35}{22.89} & \heatrowlow{8.58}{8.35}{22.89} & \heatrowlow{8.93}{8.35}{22.89} & \heatrowlow{9.00}{8.35}{22.89} \\
& & GPT-5-mini
& \heatrowlow{30.80}{9.36}{30.80} & \heatrowlow{9.75}{9.36}{30.80} & \heatrowlow{10.79}{9.36}{30.80} & \heatrowlow{10.91}{9.36}{30.80} & \heatrowlow{9.73}{9.36}{30.80} & \heatrowlow{10.58}{9.36}{30.80} & \heatrowlow{18.74}{9.36}{30.80} \\
\cmidrule{2-10}
& \multirow{4}{=}{\parbox[c]{1.2cm}{\centering\textbf{HotpotQA}}} &
Qwen3-4B
& \heatrowlow{21.87}{17.23}{38.49} & \heatrowlow{38.49}{17.23}{38.49} & \heatrowlow{32.54}{17.23}{38.49} & \heatrowlow{33.99}{17.23}{38.49} & \heatrowlow{36.10}{17.23}{38.49} & \heatrowlow{24.34}{17.23}{38.49} & \heatrowlow{17.23}{17.23}{38.49} \\
& & Qwen3-30B
& \heatrowlow{20.82}{18.99}{37.96} & \heatrowlow{32.01}{18.99}{37.96} & \heatrowlow{25.07}{18.99}{37.96} & \heatrowlow{37.96}{18.99}{37.96} & \heatrowlow{30.63}{18.99}{37.96} & \heatrowlow{19.25}{18.99}{37.96} & \heatrowlow{18.99}{18.99}{37.96} \\
& & DeepSeek-V3.2
& \heatrowlow{21.01}{14.89}{29.05} & \heatrowlow{22.46}{14.89}{29.05} & \heatrowlow{18.09}{14.89}{29.05} & \heatrowlow{29.05}{14.89}{29.05} & \heatrowlow{22.05}{14.89}{29.05} & \heatrowlow{17.38}{14.89}{29.05} & \heatrowlow{14.89}{14.89}{29.05} \\
& & GPT-5-mini
& \heatrowlow{23.33}{16.38}{23.46} & \heatrowlow{23.46}{16.38}{23.46} & \heatrowlow{16.83}{16.38}{23.46} & \heatrowlow{22.67}{16.38}{23.46} & \heatrowlow{22.86}{16.38}{23.46} & \heatrowlow{16.38}{16.38}{23.46} & \heatrowlow{20.33}{16.38}{23.46} \\
\cmidrule{2-10}
& \multirow{4}{=}{\parbox[c]{1.2cm}{\centering\textbf{CoQA}}} &
Qwen3-4B
& \heatrowlow{25.14}{11.10}{25.14} & \heatrowlow{12.07}{11.10}{25.14} & \heatrowlow{15.08}{11.10}{25.14} & \heatrowlow{16.65}{11.10}{25.14} & \heatrowlow{14.03}{11.10}{25.14} & \heatrowlow{11.94}{11.10}{25.14} & \heatrowlow{11.10}{11.10}{25.14} \\
& & Qwen3-30B
& \heatrowlow{29.07}{9.89}{29.07} & \heatrowlow{10.04}{9.89}{29.07} & \heatrowlow{13.08}{9.89}{29.07} & \heatrowlow{11.91}{9.89}{29.07} & \heatrowlow{11.54}{9.89}{29.07} & \heatrowlow{10.90}{9.89}{29.07} & \heatrowlow{9.89}{9.89}{29.07} \\
& & DeepSeek-V3.2
& \heatrowlow{40.26}{7.32}{40.26} & \heatrowlow{10.54}{7.32}{40.26} & \heatrowlow{13.15}{7.32}{40.26} & \heatrowlow{8.59}{7.32}{40.26} & \heatrowlow{7.32}{7.32}{40.26} & \heatrowlow{9.26}{7.32}{40.26} & \heatrowlow{10.42}{7.32}{40.26} \\
& & GPT-5-mini
& \heatrowlow{38.85}{5.66}{38.85} & \heatrowlow{9.59}{5.66}{38.85} & \heatrowlow{15.99}{5.66}{38.85} & \heatrowlow{5.89}{5.66}{38.85} & \heatrowlow{5.66}{5.66}{38.85} & \heatrowlow{9.80}{5.66}{38.85} & \heatrowlow{23.66}{5.66}{38.85} \\
\cmidrule{2-10}
& \multicolumn{2}{c}{\textbf{Avg.}}
& \heatrowlow{26.80}{14.47}{26.80} & \heatrowlow{18.44}{14.47}{26.80} & \heatrowlow{17.63}{14.47}{26.80} & \heatrowlow{19.99}{14.47}{26.80} & \heatrowlow{17.21}{14.47}{26.80} & \heatrowlow{14.47}{14.47}{26.80} & \heatrowlow{15.25}{14.47}{26.80} \\
\bottomrule
\end{tabular}}
\end{table}

The previous two subsections report verbalization-based and sampling-based methods separately because they constitute the two most fundamental and widely used families of black-box uncertainty estimation methods. In contrast, explanation-based, multi-agent, and hybrid methods contain fewer representative methods in our benchmark and usually rely on intermediate information from model outputs or multiple complementary signals. Therefore, we discuss these methods together in this subsection to provide a more focused analysis of the performance of more complex UE methods.

Table~\ref{tab:agent_hybrid_metrics} reports the results of explanation-based, multi-agent, and hybrid methods on the open-ended QA. Overall, \textbf{hybrid methods perform best}, with DiNCo~\cite{wang2025calibrating} and SteerConf~\cite{zhou2025steerconf} standing out. DiNCo~\cite{wang2025calibrating} benefits from combining cross-sample consistency with within-generation candidate comparison, thereby capturing both output stability across generations and relative preference within a single generation. SteerConf shows that verbalized confidence becomes more useful when evaluated under controlled prompt perturbations, as this helps distinguish stable confidence from prompt-sensitive surface confidence.

By contrast, \textbf{explanation-based and multi-agent methods are relatively weaker overall in the open-ended setting.} For COTA~\cite{tanneru2024quantifying}, one likely limitation is that its reasoning-chain consistency criterion is too strict. Even when two reasoning chains support the same conclusion, their local steps may differ enough to substantially reduce the measured overlap. As a result, the method may struggle to separate high-risk from low-risk samples using step-level alignment alone. PathWeight~\cite{zhang2025all} partially alleviates this issue by modeling reasoning support through graph structure rather than strict local step matching. By merging related reasoning paths and aggregating support for the same answer across chains, it can recover some robustness. Even so, it remains sensitive to local mismatch and variation in reasoning expression, which likely limits its overall performance.

A similar issue appears in the multi-agent setting. For Collab~\cite{yang2024confidence}, minority answers may be overly influenced by majority answers during deliberation, which can reinforce an incorrect consensus rather than correct it. For ArgLLMs~\cite{freedman2025argumentative}, the debate between supporting and opposing arguments does not always focus on the evidence that is most central to answer correctness; in some cases, it may instead emphasize local or secondary points. This can weaken rather than improve discriminative performance. In addition, the method may inherit overconfidence from the initial self-evaluation assigned to individual arguments, which then propagates through the argumentation tree. These results therefore indicate that \textbf{multi-agent interaction does not automatically yield better UE in open-ended QA}. If interaction fails to stay aligned with the core evidence underlying correctness, it may simply introduce additional noise.

\begin{figure*}[h]
    \centering
    \includegraphics[width=0.95\linewidth]{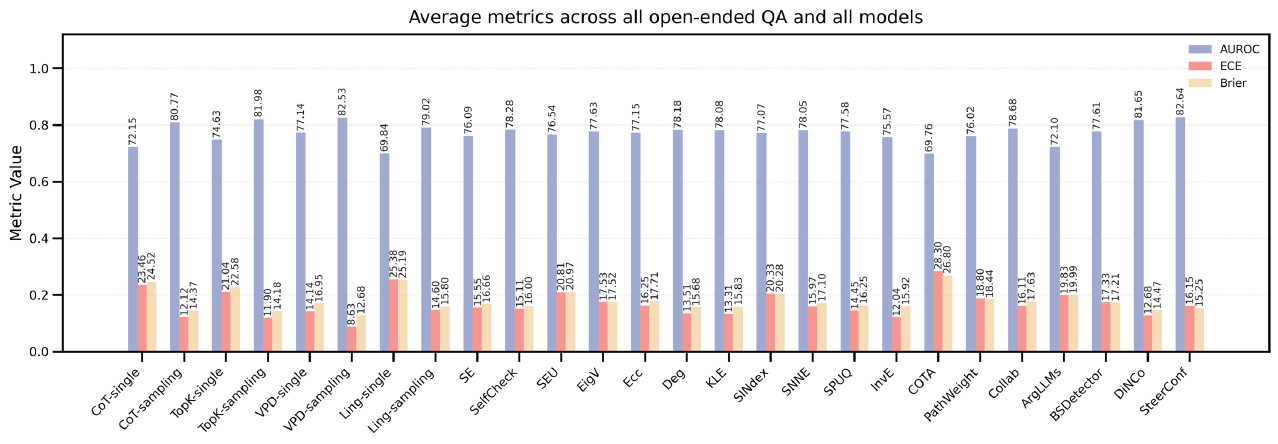}
    \caption{Comparison of UE methods on open-ended QA in terms of AUROC, ECE, and Brier Score.}
    \label{fig:open-set}
\end{figure*}
In terms of answer accuracy, the differences among these methods are relatively small, with SteerConf~\cite{zhou2025steerconf} performing slightly better on average. This may be related to its answer selection mechanism, where the calibrated confidence score is used to select the answer generated under the corresponding prompt style. As a result, the benefit of this strategy may extend beyond UE itself and slightly improve the overall quality of the final answer.

\subsubsection{Overall Comparison Across All Methods}

Figure~\ref{fig:open-set} summarizes the overall performance of all methods on the open-ended QA. \textbf{Sampling-aggregated verbalization methods and hybrid methods are both highly competitive in this setting.} In particular, SteerConf~\cite{zhou2025steerconf} achieves the strongest overall discriminative performance, while sampling-aggregated VPD~\cite{wang2025don} provides the best overall calibration. DiNCo~\cite{wang2025calibrating} also performs strongly across all metrics. Another noteworthy finding is that \textbf{VPD performs strongly even under single-pass generation}, suggesting that LLMs already contain useful uncertainty signals that can be elicited directly through appropriate prompting.

\subsubsection{Evaluation across Datasets}

\begin{figure}[!htbp]
    \centering
    \includegraphics[width=0.95\linewidth]{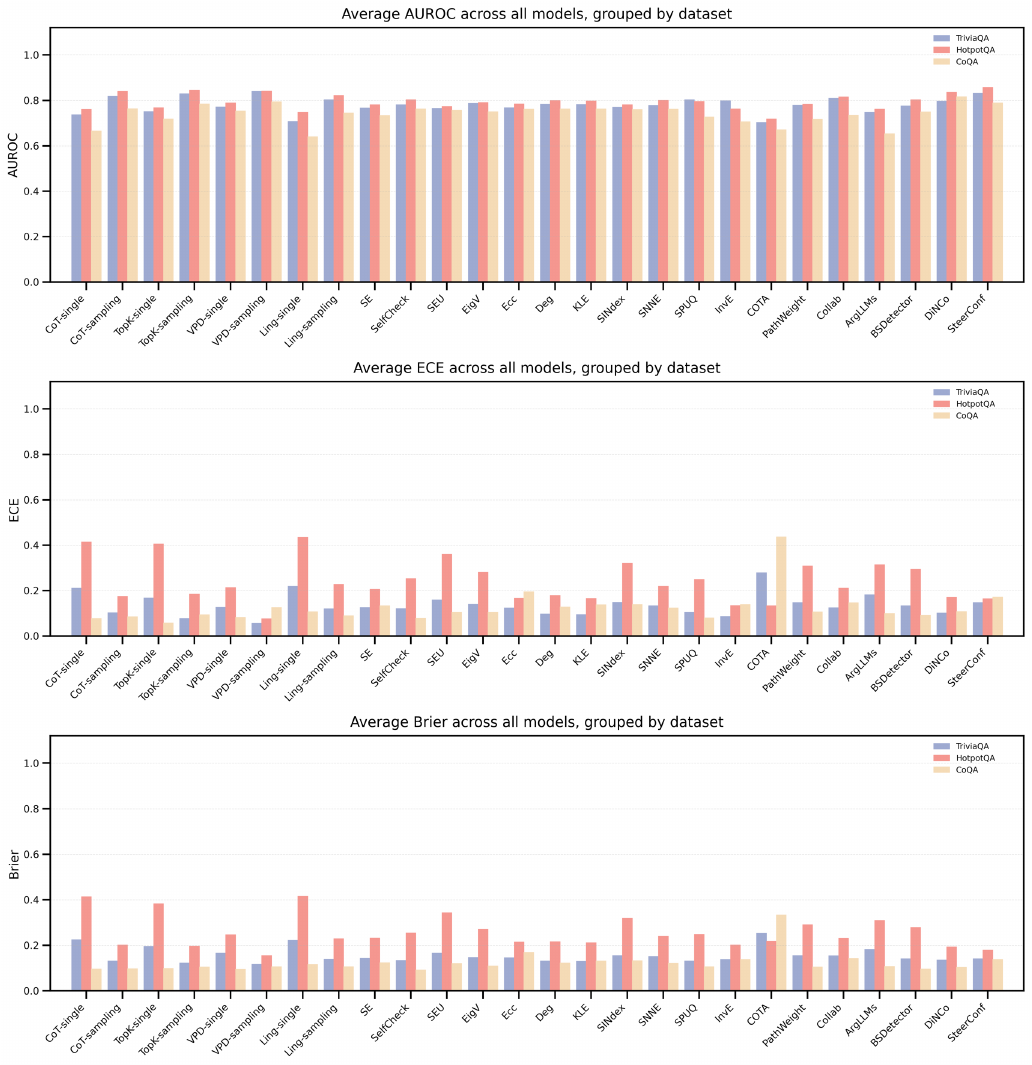}
    \caption{Comparison of UE methods across open-ended datasets, with results averaged over all evaluated models.}
    \label{fig:avg_dataset}
\end{figure}

Figure~\ref{fig:avg_dataset} compares the performance of different methods on TriviaQA, HotpotQA, and CoQA, with all results averaged over different models. In terms of AUROC, \textbf{TriviaQA and HotpotQA are generally higher than CoQA for most methods}. This may be because CoQA involves longer conversational contexts, which introduce additional uncertainty that cannot be fully captured by methods mainly based on question-answer pairs or response-level similarity. The calibration metrics show a different pattern. For Brier score and ECE, \textbf{HotpotQA is substantially worse than TriviaQA and CoQA for many methods}, mainly because this dataset is more challenging and models exhibit overconfidence, leading to a larger gap between predicted confidence and empirical correctness.

Looking further within each dataset, on TriviaQA, \textbf{sampling-aggregated VPD~\cite{wang2025don} performs best overall}, with strong performance in both discrimination and calibration. For HotpotQA, \textbf{SteerConf~\cite{zhou2025steerconf} achieves the best discriminative performance}, while \textbf{sampling-aggregated VPD performs best in calibration}. For CoQA, \textbf{DiNCo~\cite{wang2025calibrating} achieves the strongest discriminative performance}, while \textbf{sampling-aggregated TopK~\cite{tian2023just} performs best in calibration}.

\subsubsection{Evaluation across Models}

\begin{figure}[!htbp]
    \centering
    \includegraphics[width=0.94\linewidth]{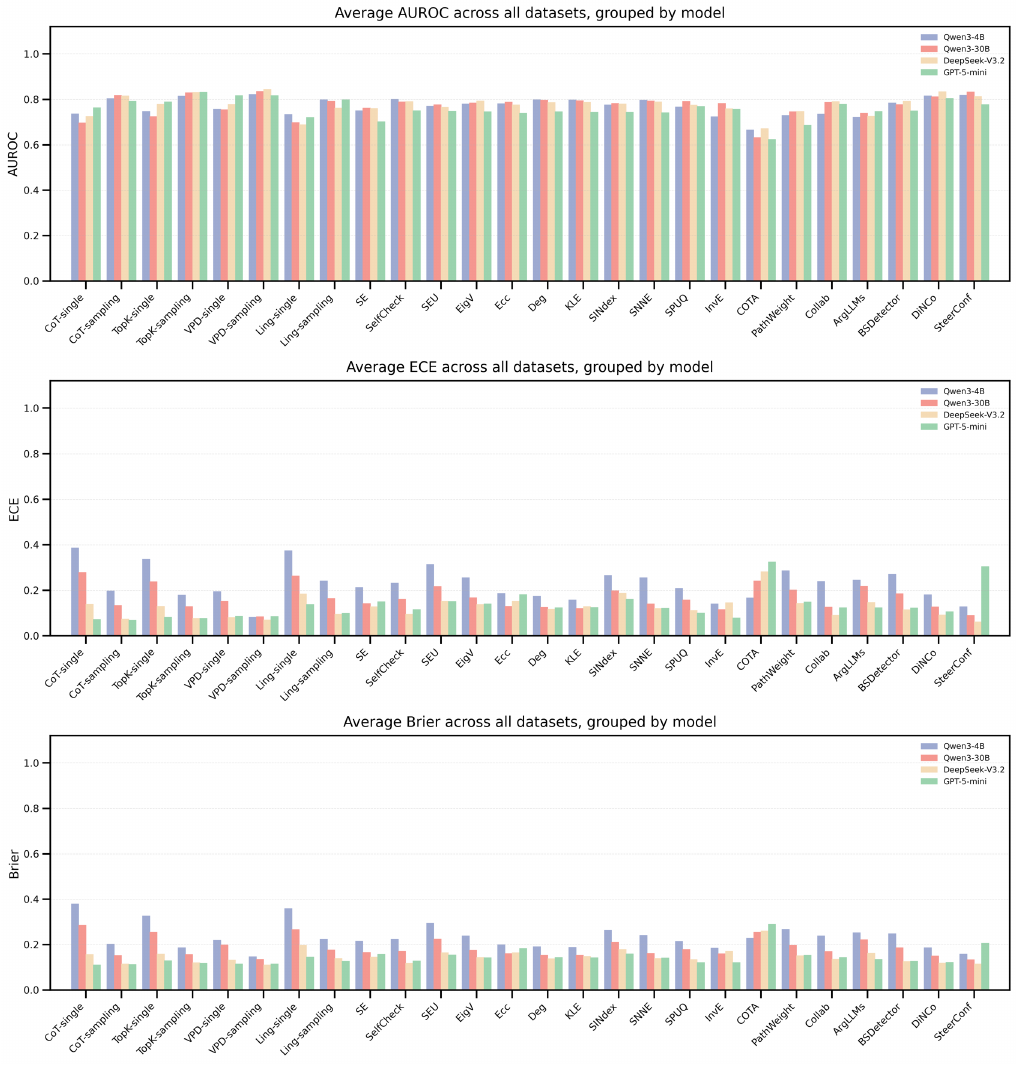}
    \caption{Comparison of UE methods across models, with results averaged over the open-ended datasets.}
    \label{fig:avg_model}
\end{figure}

Figure~\ref{fig:avg_model} compares the performance of different methods on Qwen3-4B, Qwen3-30B, DeepSeek-V3.2, and GPT-5-mini, with all results averaged over the open-ended datasets. In terms of AUROC, \textbf{Qwen3-30B and DeepSeek-V3.2 achieve relatively high scores for most methods, Qwen3-4B is slightly lower but still competitive,} whereas \textbf{GPT-5-mini shows weaker AUROC for some methods}. One possible explanation is that GPT-5-mini produces more diverse surface forms across samples, even when the underlying semantics remain similar, which may weaken methods relying on response similarity or consistency. For ECE and Brier score, \textbf{these metrics generally decrease as overall model performance improves}, although this may largely reflect higher base accuracy rather than intrinsically more reliable confidence expression. GPT-5-mini also shows generally good calibration, but under SteerConf~\cite{zhou2025steerconf}, its ECE and Brier score are relatively high, likely because it is overly sensitive to the "least confident" steering prompt and becomes systematically under-confident.

Looking further within each model, for the Qwen series, \textbf{SteerConf~\cite{zhou2025steerconf} achieves the best discriminative performance}, while \textbf{sampling-aggregated VPD~\cite{wang2025don} performs best on calibration metrics}. For DeepSeek-V3.2, \textbf{sampling-aggregated VPD achieves the strongest discriminative performance}, while \textbf{SteerConf~\cite{zhou2025steerconf} performs best in calibration}. For GPT-5-mini, \textbf{sampling-aggregated TopK~\cite{tian2023just} achieves the best discriminative performance}, while \textbf{single-pass CoT~\cite{xiong2306can} performs best in calibration}.

\subsubsection{Evaluation under Different Sample Sizes}

\begin{figure*}[!htbp]
    \centering
    \includegraphics[width=0.95\linewidth]{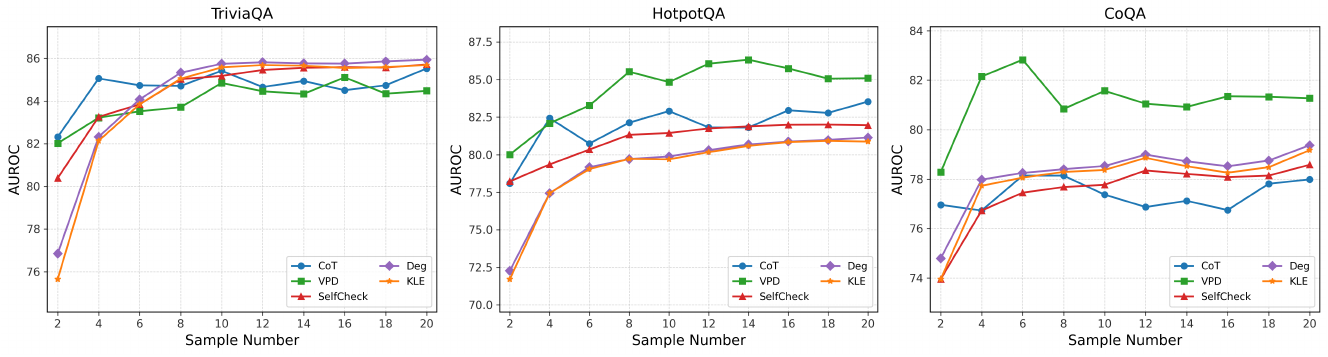}
    \caption{AUROC trends with increasing sample size across datasets on Qwen3-4B-Instruct.}
    \label{fig:sample_num}
\end{figure*}

Figure~\ref{fig:sample_num} shows the effect of varying the number of samples for several representative methods. \textbf{For sampling-based methods, AUROC generally increases as the sample size grows,} indicating that additional samples provide stronger evidence for uncertainty estimation and improve discrimination. By contrast, \textbf{verbalization-based methods show more noticeable fluctuations as the sample size increases and do not exhibit the same steady gains.}

\subsection{Results on Closed-ended QA}

\begin{table*}[htbp]
\centering
\caption{Results of black-box UE methods on the closed-ended QA. We compare verbalization-based methods under single-pass and sampling-aggregated settings, together with other black-box UE methods applicable to the closed-ended QA. Avg. denotes the macro-average over all models shown in the table. The same row-wise shading rule as in Table~2 is used for visualization.}

\label{tab:closed_all}
\resizebox{0.99\textwidth}{!}{
\begin{tabular}{@{}>{\centering\arraybackslash}p{0.8cm}
                >{\centering\arraybackslash}p{1.9cm}
                *{16}{>{\centering\arraybackslash}p{0.8cm}}@{\hspace{0.15cm}}}
\toprule
\multirow{2}{*}{\textbf{Metric}} &
\multirow{2}{*}{\textbf{Model}} &
\multicolumn{4}{c}{\textbf{Single-pass}} &
\multicolumn{4}{c}{\textbf{Sampling-aggregated}} &
\multirow{2}{*}{\textbf{SE}} &
\multirow{2}{*}{\textbf{COTA}} &
\multirow{2}{*}{\makebox[\linewidth][c]{\textbf{\scriptsize PathWeight}}} &
\multirow{2}{*}{\textbf{T3}} &
\multirow{2}{*}{\textbf{Collab}} &
\multirow{2}{*}{\makebox[\linewidth][c]{\textbf{\scriptsize ArgLLMs}}} &
\multirow{2}{*}{\textbf{UF}} &
\multirow{2}{*}{\makebox[\linewidth][c]{\textbf{\scriptsize SteerConf}}} \\
\cmidrule(lr){3-6}
\cmidrule(lr){7-10}
& &
\textbf{CoT} & \textbf{TopK} & \textbf{VPD} & \textbf{Ling} &
\textbf{CoT} & \textbf{TopK} & \textbf{VPD} & \textbf{Ling} &
& & & & & & & \\
\midrule

\multirow{5}{*}{\textbf{Acc}} &
Qwen3-4B
& \heatrowhigh{69.14}{68.80}{70.60} & \heatrowhigh{68.93}{68.80}{70.60} & \heatrowhigh{70.14}{68.80}{70.60} & \heatrowhigh{68.94}{68.80}{70.60}
& \heatrowhigh{70.40}{68.80}{70.60} & \heatrowhigh{69.80}{68.80}{70.60} & \heatrowhigh{70.60}{68.80}{70.60} & \heatrowhigh{68.80}{68.80}{70.60}
& \heatrowhigh{68.94}{68.60}{72.00} & \heatrowhigh{68.94}{68.60}{72.00} & \heatrowhigh{68.60}{68.60}{72.00} & \heatrowhigh{70.60}{68.60}{72.00} & \heatrowhigh{71.14}{68.60}{72.00} & \heatrowhigh{68.94}{68.60}{72.00} & \heatrowhigh{69.80}{68.60}{72.00} & \heatrowhigh{72.00}{68.60}{72.00} \\

& Qwen3-30B
& \heatrowhigh{75.20}{71.89}{77.60} & \heatrowhigh{74.00}{71.89}{77.60} & \heatrowhigh{75.60}{71.89}{77.60} & \heatrowhigh{73.20}{71.89}{77.60}
& \heatrowhigh{77.60}{71.89}{77.60} & \heatrowhigh{75.60}{71.89}{77.60} & \heatrowhigh{76.40}{71.89}{77.60} & \heatrowhigh{71.89}{71.89}{77.60}
& \heatrowhigh{74.95}{74.95}{82.80} & \heatrowhigh{74.95}{74.95}{82.80} & \heatrowhigh{75.60}{74.95}{82.80} & \heatrowhigh{75.75}{74.95}{82.80} & \heatrowhigh{76.60}{74.95}{82.80} & \heatrowhigh{74.95}{74.95}{82.80} & \heatrowhigh{76.40}{74.95}{82.80} & \heatrowhigh{82.80}{74.95}{82.80} \\

& DeepSeek-V3.2
& \heatrowhigh{81.40}{74.60}{81.40} & \heatrowhigh{79.55}{74.60}{81.40} & \heatrowhigh{79.60}{74.60}{81.40} & \heatrowhigh{74.60}{74.60}{81.40}
& \heatrowhigh{79.40}{74.60}{81.40} & \heatrowhigh{78.40}{74.60}{81.40} & \heatrowhigh{81.40}{74.60}{81.40} & \heatrowhigh{77.60}{74.60}{81.40}
& \heatrowhigh{78.00}{66.00}{82.80} & \heatrowhigh{78.00}{66.00}{82.80} & \heatrowhigh{77.20}{66.00}{82.80} & \heatrowhigh{66.00}{66.00}{82.80} & \heatrowhigh{80.96}{66.00}{82.80} & \heatrowhigh{78.00}{66.00}{82.80} & \heatrowhigh{79.80}{66.00}{82.80} & \heatrowhigh{82.80}{66.00}{82.80} \\

& GPT-5-mini
& \heatrowhigh{80.80}{76.40}{81.56} & \heatrowhigh{79.60}{76.40}{81.56} & \heatrowhigh{81.00}{76.40}{81.56} & \heatrowhigh{77.15}{76.40}{81.56}
& \heatrowhigh{81.20}{76.40}{81.56} & \heatrowhigh{80.40}{76.40}{81.56} & \heatrowhigh{81.56}{76.40}{81.56} & \heatrowhigh{76.40}{76.40}{81.56}
& \heatrowhigh{81.00}{80.20}{81.69} & \heatrowhigh{81.00}{80.20}{81.69} & \heatrowhigh{80.20}{80.20}{81.69} & \heatrowhigh{81.69}{80.20}{81.69} & \heatrowhigh{81.00}{80.20}{81.69} & \heatrowhigh{81.00}{80.20}{81.69} & \heatrowhigh{80.40}{80.20}{81.69} & \heatrowhigh{81.40}{80.20}{81.69} \\

\cmidrule{2-18}
& \textbf{Avg.}
& \heatrowhigh{76.64}{73.47}{77.49} & \heatrowhigh{75.52}{73.47}{77.49} & \heatrowhigh{76.59}{73.47}{77.49} & \heatrowhigh{73.47}{73.47}{77.49}
& \heatrowhigh{77.15}{73.47}{77.49} & \heatrowhigh{76.05}{73.47}{77.49} & \heatrowhigh{77.49}{73.47}{77.49} & \heatrowhigh{73.67}{73.47}{77.49}
& \heatrowhigh{75.72}{73.51}{79.60} & \heatrowhigh{75.72}{73.51}{79.60} & \heatrowhigh{75.40}{73.51}{79.60} & \heatrowhigh{73.51}{73.51}{79.60} & \heatrowhigh{77.43}{73.51}{79.60} & \heatrowhigh{75.72}{73.51}{79.60} & \heatrowhigh{76.60}{73.51}{79.60} & \heatrowhigh{79.60}{73.51}{79.60} \\
\midrule

\multirow{5}{*}{\textbf{AUROC}} &
Qwen3-4B
& \heatrowhigh{76.07}{73.37}{82.46} & \heatrowhigh{75.59}{73.37}{82.46} & \heatrowhigh{73.37}{73.37}{82.46} & \heatrowhigh{78.61}{73.37}{82.46}
& \heatrowhigh{78.28}{73.37}{82.46} & \heatrowhigh{82.46}{73.37}{82.46} & \heatrowhigh{81.00}{73.37}{82.46} & \heatrowhigh{78.62}{73.37}{82.46}
& \heatrowhigh{69.60}{55.09}{85.47} & \heatrowhigh{55.09}{55.09}{85.47} & \heatrowhigh{59.72}{55.09}{85.47} & \heatrowhigh{85.47}{55.09}{85.47} & \heatrowhigh{65.52}{55.09}{85.47} & \heatrowhigh{77.64}{55.09}{85.47} & \heatrowhigh{83.30}{55.09}{85.47} & \heatrowhigh{73.44}{55.09}{85.47} \\

& Qwen3-30B
& \heatrowhigh{79.08}{72.87}{86.83} & \heatrowhigh{83.32}{72.87}{86.83} & \heatrowhigh{77.14}{72.87}{86.83} & \heatrowhigh{72.87}{72.87}{86.83}
& \heatrowhigh{85.36}{72.87}{86.83} & \heatrowhigh{86.83}{72.87}{86.83} & \heatrowhigh{85.57}{72.87}{86.83} & \heatrowhigh{78.94}{72.87}{86.83}
& \heatrowhigh{70.95}{47.99}{86.40} & \heatrowhigh{47.99}{47.99}{86.40} & \heatrowhigh{63.71}{47.99}{86.40} & \heatrowhigh{85.28}{47.99}{86.40} & \heatrowhigh{75.88}{47.99}{86.40} & \heatrowhigh{75.85}{47.99}{86.40} & \heatrowhigh{86.40}{47.99}{86.40} & \heatrowhigh{79.77}{47.99}{86.40} \\

& DeepSeek-V3.2
& \heatrowhigh{72.47}{72.47}{86.86} & \heatrowhigh{84.75}{72.47}{86.86} & \heatrowhigh{85.57}{72.47}{86.86} & \heatrowhigh{73.50}{72.47}{86.86}
& \heatrowhigh{79.70}{72.47}{86.86} & \heatrowhigh{86.40}{72.47}{86.86} & \heatrowhigh{86.86}{72.47}{86.86} & \heatrowhigh{73.65}{72.47}{86.86}
& \heatrowhigh{70.30}{52.88}{87.88} & \heatrowhigh{52.88}{52.88}{87.88} & \heatrowhigh{65.45}{52.88}{87.88} & \heatrowhigh{86.51}{52.88}{87.88} & \heatrowhigh{75.42}{52.88}{87.88} & \heatrowhigh{77.67}{52.88}{87.88} & \heatrowhigh{87.88}{52.88}{87.88} & \heatrowhigh{79.93}{52.88}{87.88} \\

& GPT-5-mini
& \heatrowhigh{75.51}{73.26}{84.30} & \heatrowhigh{76.50}{73.26}{84.30} & \heatrowhigh{81.74}{73.26}{84.30} & \heatrowhigh{73.26}{73.26}{84.30}
& \heatrowhigh{80.16}{73.26}{84.30} & \heatrowhigh{83.52}{73.26}{84.30} & \heatrowhigh{84.30}{73.26}{84.30} & \heatrowhigh{81.38}{73.26}{84.30}
& \heatrowhigh{65.92}{44.99}{85.98} & \heatrowhigh{44.99}{44.99}{85.98} & \heatrowhigh{62.69}{44.99}{85.98} & \heatrowhigh{85.98}{44.99}{85.98} & \heatrowhigh{76.85}{44.99}{85.98} & \heatrowhigh{77.68}{44.99}{85.98} & \heatrowhigh{84.82}{44.99}{85.98} & \heatrowhigh{72.85}{44.99}{85.98} \\

\cmidrule{2-18}
& \textbf{Avg.}
& \heatrowhigh{75.78}{50.24}{85.81} & \heatrowhigh{80.04}{50.24}{85.81} & \heatrowhigh{79.45}{50.24}{85.81} & \heatrowhigh{74.56}{50.24}{85.81}
& \heatrowhigh{80.88}{50.24}{85.81} & \heatrowhigh{84.80}{50.24}{85.81} & \heatrowhigh{84.43}{50.24}{85.81} & \heatrowhigh{78.15}{50.24}{85.81}
& \heatrowhigh{69.20}{50.24}{85.81} & \heatrowhigh{50.24}{50.24}{85.81} & \heatrowhigh{62.89}{50.24}{85.81} & \heatrowhigh{85.81}{50.24}{85.81} & \heatrowhigh{73.42}{50.24}{85.81} & \heatrowhigh{77.21}{50.24}{85.81} & \heatrowhigh{85.60}{50.24}{85.81} & \heatrowhigh{76.50}{50.24}{85.81} \\
\midrule

\multirow{5}{*}{\textbf{ECE}} &
Qwen3-4B
& \heatrowlow{26.74}{7.41}{26.74} & \heatrowlow{24.85}{7.41}{26.74} & \heatrowlow{19.37}{7.41}{26.74} & \heatrowlow{25.98}{7.41}{26.74}
& \heatrowlow{18.51}{7.41}{26.74} & \heatrowlow{16.72}{7.41}{26.74} & \heatrowlow{7.41}{7.41}{26.74} & \heatrowlow{22.54}{7.41}{26.74}
& \heatrowlow{20.41}{7.76}{28.64} & \heatrowlow{14.88}{7.76}{28.64} & \heatrowlow{28.64}{7.76}{28.64} & \heatrowlow{20.53}{7.76}{28.64} & \heatrowlow{14.46}{7.76}{28.64} & \heatrowlow{15.06}{7.76}{28.64} & \heatrowlow{12.05}{7.76}{28.64} & \heatrowlow{7.76}{7.76}{28.64} \\

& Qwen3-30B
& \heatrowlow{20.83}{3.24}{20.83} & \heatrowlow{16.68}{3.24}{20.83} & \heatrowlow{3.24}{3.24}{20.83} & \heatrowlow{19.83}{3.24}{20.83}
& \heatrowlow{11.63}{3.24}{20.83} & \heatrowlow{13.53}{3.24}{20.83} & \heatrowlow{7.55}{3.24}{20.83} & \heatrowlow{18.57}{3.24}{20.83}
& \heatrowlow{17.35}{4.25}{20.45} & \heatrowlow{18.94}{4.25}{20.45} & \heatrowlow{20.45}{4.25}{20.45} & \heatrowlow{12.74}{4.25}{20.45} & \heatrowlow{7.85}{4.25}{20.45} & \heatrowlow{16.35}{4.25}{20.45} & \heatrowlow{12.13}{4.25}{20.45} & \heatrowlow{4.25}{4.25}{20.45} \\

& DeepSeek-V3.2
& \heatrowlow{10.68}{3.58}{13.12} & \heatrowlow{13.12}{3.58}{13.12} & \heatrowlow{5.46}{3.58}{13.12} & \heatrowlow{17.07}{3.58}{13.12}
& \heatrowlow{6.83}{3.58}{13.12} & \heatrowlow{7.00}{3.58}{13.12} & \heatrowlow{3.58}{3.58}{13.12} & \heatrowlow{9.96}{3.58}{13.12}
& \heatrowlow{12.65}{2.30}{19.13} & \heatrowlow{12.03}{2.30}{19.13} & \heatrowlow{19.13}{2.30}{19.13} & \heatrowlow{13.45}{2.30}{19.13} & \heatrowlow{4.66}{2.30}{19.13} & \heatrowlow{12.26}{2.30}{19.13} & \heatrowlow{9.89}{2.30}{19.13} & \heatrowlow{2.30}{2.30}{19.13} \\

& GPT-5-mini
& \heatrowlow{10.82}{5.17}{16.60} & \heatrowlow{7.57}{5.17}{16.60} & \heatrowlow{7.68}{5.17}{16.60} & \heatrowlow{16.60}{5.17}{16.60}
& \heatrowlow{7.09}{5.17}{16.60} & \heatrowlow{5.17}{5.17}{16.60} & \heatrowlow{5.45}{5.17}{16.60} & \heatrowlow{13.36}{5.17}{16.60}
& \heatrowlow{15.94}{6.99}{25.44} & \heatrowlow{20.13}{6.99}{25.44} & \heatrowlow{18.62}{6.99}{25.44} & \heatrowlow{6.99}{6.99}{25.44} & \heatrowlow{12.11}{6.99}{25.44} & \heatrowlow{11.48}{6.99}{25.44} & \heatrowlow{13.89}{6.99}{25.44} & \heatrowlow{25.44}{6.99}{25.44} \\

\cmidrule{2-18}
& \textbf{Avg.}
& \heatrowlow{17.27}{6.00}{21.71} & \heatrowlow{15.55}{6.00}{21.71} & \heatrowlow{8.94}{6.00}{21.71} & \heatrowlow{19.87}{6.00}{21.71}
& \heatrowlow{11.02}{6.00}{21.71} & \heatrowlow{10.60}{6.00}{21.71} & \heatrowlow{6.00}{6.00}{21.71} & \heatrowlow{16.11}{6.00}{21.71}
& \heatrowlow{16.64}{6.00}{21.71} & \heatrowlow{16.50}{6.00}{21.71} & \heatrowlow{21.71}{6.00}{21.71} & \heatrowlow{13.43}{6.00}{21.71} & \heatrowlow{9.61}{6.00}{21.71} & \heatrowlow{13.94}{6.00}{21.71} & \heatrowlow{11.99}{6.00}{21.71} & \heatrowlow{9.94}{6.00}{21.71} \\
\midrule

\multirow{5}{*}{\textbf{Brier}} &
Qwen3-4B
& \heatrowlow{27.10}{15.92}{27.10} & \heatrowlow{25.72}{15.92}{27.10} & \heatrowlow{22.38}{15.92}{27.10} & \heatrowlow{25.95}{15.92}{27.10}
& \heatrowlow{20.57}{15.92}{27.10} & \heatrowlow{19.68}{15.92}{27.10} & \heatrowlow{15.92}{15.92}{27.10} & \heatrowlow{24.01}{15.92}{27.10}
& \heatrowlow{22.12}{15.79}{28.18} & \heatrowlow{24.89}{15.79}{28.18} & \heatrowlow{28.18}{15.79}{28.18} & \heatrowlow{21.01}{15.79}{28.18} & \heatrowlow{21.12}{15.79}{28.18} & \heatrowlow{20.07}{15.79}{28.18} & \heatrowlow{15.79}{15.79}{28.18} & \heatrowlow{18.20}{15.79}{28.18} \\

& Qwen3-30B
& \heatrowlow{21.79}{13.40}{21.79} & \heatrowlow{19.48}{13.40}{21.79} & \heatrowlow{15.33}{13.40}{21.79} & \heatrowlow{21.43}{13.40}{21.79}
& \heatrowlow{14.58}{13.40}{21.79} & \heatrowlow{16.81}{13.40}{21.79} & \heatrowlow{13.40}{13.40}{21.79} & \heatrowlow{20.91}{13.40}{21.79}
& \heatrowlow{18.82}{12.64}{24.44} & \heatrowlow{24.44}{12.64}{24.44} & \heatrowlow{20.77}{12.64}{24.44} & \heatrowlow{16.18}{12.64}{24.44} & \heatrowlow{15.67}{12.64}{24.44} & \heatrowlow{18.77}{12.64}{24.44} & \heatrowlow{13.93}{12.64}{24.44} & \heatrowlow{12.64}{12.64}{24.44} \\

& DeepSeek-V3.2
& \heatrowlow{14.87}{10.61}{19.58} & \heatrowlow{16.61}{10.61}{19.58} & \heatrowlow{11.53}{10.61}{19.58} & \heatrowlow{19.58}{10.61}{19.58}
& \heatrowlow{13.27}{10.61}{19.58} & \heatrowlow{13.64}{10.61}{19.58} & \heatrowlow{10.61}{10.61}{19.58} & \heatrowlow{16.18}{10.61}{19.58}
& \heatrowlow{16.08}{11.26}{20.05} & \heatrowlow{20.05}{11.26}{20.05} & \heatrowlow{19.33}{11.26}{20.05} & \heatrowlow{17.78}{11.26}{20.05} & \heatrowlow{13.25}{11.26}{20.05} & \heatrowlow{15.57}{11.26}{20.05} & \heatrowlow{11.26}{11.26}{20.05} & \heatrowlow{11.70}{11.26}{20.05} \\

& GPT-5-mini
& \heatrowlow{15.04}{11.59}{18.28} & \heatrowlow{14.93}{11.59}{18.28} & \heatrowlow{13.21}{11.59}{18.28} & \heatrowlow{18.28}{11.59}{18.28}
& \heatrowlow{13.00}{11.59}{18.28} & \heatrowlow{12.46}{11.59}{18.28} & \heatrowlow{11.59}{11.59}{18.28} & \heatrowlow{16.27}{11.59}{18.28}
& \heatrowlow{16.72}{12.05}{22.62} & \heatrowlow{22.62}{12.05}{22.62} & \heatrowlow{18.36}{12.05}{22.62} & \heatrowlow{12.05}{12.05}{22.62} & \heatrowlow{13.61}{12.05}{22.62} & \heatrowlow{14.92}{12.05}{22.62} & \heatrowlow{14.41}{12.05}{22.62} & \heatrowlow{19.69}{12.05}{22.62} \\

\cmidrule{2-18}
& \textbf{Avg.}
& \heatrowlow{19.70}{12.88}{23.00} & \heatrowlow{19.18}{12.88}{23.00} & \heatrowlow{15.61}{12.88}{23.00} & \heatrowlow{21.31}{12.88}{23.00}
& \heatrowlow{15.35}{12.88}{23.00} & \heatrowlow{15.65}{12.88}{23.00} & \heatrowlow{12.88}{12.88}{23.00} & \heatrowlow{19.34}{12.88}{23.00}
& \heatrowlow{18.48}{12.88}{23.00} & \heatrowlow{23.00}{12.88}{23.00} & \heatrowlow{21.66}{12.88}{23.00} & \heatrowlow{16.76}{12.88}{23.00} & \heatrowlow{15.91}{12.88}{23.00} & \heatrowlow{17.33}{12.88}{23.00} & \heatrowlow{13.84}{12.88}{23.00} & \heatrowlow{15.56}{12.88}{23.00} \\
\bottomrule
\end{tabular}}
\end{table*}

\begin{figure}[!htbp]
    \centering
    \includegraphics[width=0.95\linewidth]{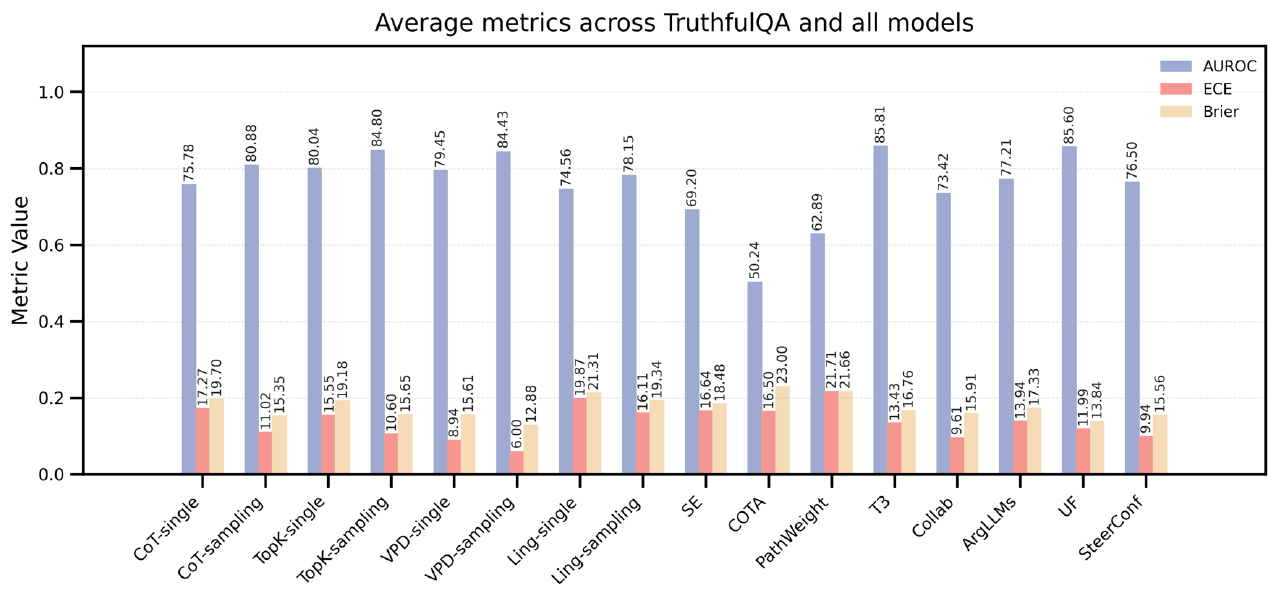}
    \caption{Comparison of UE methods on the closed-ended QA in terms of AUROC, ECE, and Brier Score.}
    \label{fig:closed_all}
\end{figure}

Table~\ref{tab:closed_all} reports the results of different methods on closed-ended QA. For verbalization-based methods, \textbf{sampling aggregation outperforms single-pass generation on most metrics}, indicating that repeated sampling can still provide additional uncertainty signals even when the answer space is explicitly given. Similar to the open-set setting, \textbf{TopK~\cite{tian2023just} and VPD~\cite{wang2025don} perform strongly overall, while Ling~\cite{tian2023just} remains relatively weak}. A closer comparison shows that TopK slightly outperforms VPD on AUROC, suggesting that both methods can effectively exploit candidate answers for discrimination. By contrast, \textbf{VPD has a clearer advantage in calibration,} likely because it explicitly requires the model to output a normalized distribution over candidate answers, making its confidence values more naturally aligned with empirical correctness.

Overall, \textbf{methods specifically designed for multiple-choice settings perform best.} T3~\cite{li2024think} generates an explanation for each candidate option and compares these explanations jointly, thereby exploiting the structural advantage of having a known candidate set. UF~\cite{zhang2024calibrating} decomposes confidence into question-level uncertainty and answer fidelity, providing a more explicit treatment of different uncertainty sources. Compared with more general black-box UE methods, these methods are better matched to the closed-ended task format, which explains their clear advantage on this benchmark. It is worth noting that T3 shows relatively low accuracy on DeepSeek-V3.2, possibly because presenting explanations for all candidate options simultaneously introduces additional interference and makes final answer selection less stable.

Figure~\ref{fig:closed_all} provides an aggregated comparison of all methods in the closed-ended setting. The results show that \textbf{methods specifically designed for this setting, such as T3~\cite{li2024think} and UF~\cite{zhang2024calibrating}, achieve the strongest discriminative performance,} suggesting that explicit candidate modeling is highly beneficial when the answer space is known. Meanwhile, among more general black-box methods, sampling-based verbalization methods remain competitive, \textbf{with VPD~\cite{wang2025don} achieving the best overall calibration.}
\section{Conclusion and Future Directions}

This paper provides a systematic review and empirical analysis of black-box UE methods for LLMs. We organize existing approaches into five categories and benchmark 24 representative methods across open-ended QA and closed-ended QA settings. In open-ended settings, verbalization-based methods already show strong competitiveness. In particular, well-designed candidate-comparison methods such as VPD~\cite{wang2025don} can achieve performance comparable to some sampling-based methods even under single-pass generation. Meanwhile, hybrid methods such as SteerConf~\cite{zhou2025steerconf} and DiNCo~\cite{wang2025calibrating}, which combine multiple uncertainty signals, also perform well in several settings. 
In closed-ended QA, the predefined candidate space provides useful additional structure, which is effectively leveraged by methods such as T3~\cite{li2024think} and UF~\cite{zhang2024calibrating}.
Beyond its practical value under restricted model access, black-box UE offers an output-level perspective that can inform uncertainty estimation research more broadly. By examining only observable model behaviors, black-box methods reveal which external signals—such as response consistency, verbalized self-assessment, candidate comparison, and hybrid uncertainty cues—are indicative of model reliability. These empirical patterns can guide white-box UE by clarifying what internal representations, logits, or activation-based signals should capture, explain, or complement. They also provide insights for UE in MLLMs, where language-model components often remain responsible for generating final answers, explanations, and reasoning traces despite the presence of visual or multimodal inputs. In this sense, the taxonomy and empirical findings presented in this work not only establish a practical benchmark for black-box UE, but also offer a reference point for future studies on uncertainty estimation in both LLMs and MLLMs. To support broader and more systematic progress, we make our project publicly available at \href{https://github.com/wangjy0116/Black-Box-UE-Hub}{Black-Box-UE-Hub}.

Despite recent progress in black-box UE, several important challenges remain. First, future research should move beyond single-turn QA toward more realistic settings, including multi-turn interaction, agents, and tool use~\cite{duan2025uprop, zhang2026confidence,oh2026uncertainty}. In these scenarios, uncertainty is no longer a static property of a final answer, but evolves dynamically with dialogue history, intermediate reasoning states, action choices, tool feedback, and accumulated errors from earlier steps. Therefore, methods designed for single-turn answer prediction, such as TopK~\cite{tian2023just} or VPD~\cite{wang2025don}, may no longer be fully applicable, because uncertainty may arise not only from answer selection, but also from process-level factors such as state tracking, plan revision, user ambiguity, and tool execution. 
Second, future work requires benchmarks explicitly designed for UE rather than adapted from general-purpose QA tasks. Existing evaluations mostly rely on standard QA datasets, where uncertainty labels, calibration targets, and evaluation protocols are not central design objectives. This may obscure whether performance differences arise from the intrinsic limitations of UE methods or from mismatches between benchmark design and the UE objective. Although recent efforts such as UBench~\cite{wang2025ubench} have begun to explore UE-oriented evaluation, existing benchmarks still do not fully cover the diverse settings faced by black-box UE methods, especially scenarios where models jointly generate answers and estimate confidence. Future benchmarks should therefore place UE at the center of task design, covering broader task formats and application scenarios with standardized protocols for calibration, selective prediction, and robustness evaluation. Such benchmarks would support a more systematic assessment of different UE methods. 
In addition, black-box UE needs to account for the heterogeneous sources of uncertainty underlying LLM outputs. Recent studies have moved beyond the traditional aleatoric/epistemic distinction by examining factors such as input ambiguity, knowledge gaps, and decoding randomness~\cite{hou2023decomposing,taparia2026anatomy}. However, many of these analyses assume access to information or intervention mechanisms that may be unavailable in strict black-box API settings. This calls for source-aware black-box UE methods that can infer different uncertainty causes from observable behaviors alone, rather than producing only a single confidence score. For example, uncertainty may stem from ambiguous prompts, conflicting contexts, insufficient model knowledge, or unstable decoding behaviors, each of which requires different interpretations and mitigation strategies. Distinguishing these sources would make black-box UE more diagnostic, interpretable, and useful for downstream decision-making.

\ifCLASSOPTIONcompsoc
{\small
		\bibliographystyle{unsrt}
		\bibliography{reference}
}

\ifCLASSOPTIONcaptionsoff
  \newpage
\fi




\end{document}